# Neural Network Parameter-optimization of Gaussian Pre-marginalized Directed Acyclic Graphs

**Mehrzad Saremi** MEHRZAD.SAREMI@GMAIL.COM
*Graduate from the Department of Artificial Intelligence*
*Amirkabir University of Technology*
*Tehran, Iran*

## Abstract

Finding the parameters of a latent variable causal model is central to causal inference and causal identification. In this article, we show that existing graphical structures that are used in causal inference are not stable under marginalization of Gaussian Bayesian networks, and present a graphical structure that faithfully represents margins of Gaussian Bayesian networks. We present the first duality between parameter optimization of a latent variable model and training a feed-forward neural network in the parameter space of the assumed family of distributions. Based on this observation, we develop an algorithm for parameter optimization of these graphical structures using the observational distribution. Then, we provide conditions for causal effect identifiability in the Gaussian setting. We propose a meta-algorithm that checks whether a causal effect is identifiable or not. Moreover, we lay a grounding for generalizing the duality between a neural network and a causal model from the Gaussian to other distributions.

## 1 Introduction

Bayesian networks are widely used in probabilistic reasoning and causal inference. They factorize a probability distribution over an acyclic graph. Among Bayesian networks, causal Bayesian networks encode causal relationships between random variables and, simultaneously encoding causal and probabilistic hypotheses, are useful in interventional inference. Functional Bayesian networks, as a sub-type of causal Bayesian networks, encode structural equation models (SEMs), which are useful in both interventional and counterfactual analysis (Pearl et al., 2000). A challenge in the domain of causal inference is modeling causal relationships with the presence of latent mutual ancestral variables assumed to play the role of "confounders" (common causes) of visible variables. These models are margins of (causal) Bayesian networks.

These margins are constructed over directed (hyper-)graphs; the directed edges encode the direct causation with respect to the hyper-graph, and the hyper-edges encode the latent common causes. Conventionally, a type of graph with the combination of directed and *bi-directed* edges used for modeling these margins (Bareinboim et al., 2020; Spirtes et al., 2000). However, as shown in various studies, bi-directed edges are too weak to encode the margins of Bayesian networks (Evans, 2016; Fritz, 2012; Forré and Mooij, 2017). Despite the earlier convention and as a surprisingly recent direction in causal inference, common causes are not necessarily assumed to be binary relations encoded by bi-directed edges, but multi-ary relations that require hyper-edges to be encoded (Evans, 2016, 2018). A prominent advantage of using hyper-edges is that such structures are stable under the marginalization of Bayesian networks (Evans, 2016).

The existing classes of the graphical structures that are used to encode the margins of Bayesian networks, or causal models with latent common causes, range from simple directed acyclic graph (DAG)-based Bayesian networks through marginalized (hyper-)graphs like acyclic directed mixed graphs (ADMGs) (Richardson and Spirtes, 2002) and marginal DAGs (mDAGs) (Evans, 2016) to graphical models with cycles and confounding latent variables (Forré and Mooij, 2017). Despite mDAGs' nice statistical properties (Shpitser et al., 2014), as we show in this article, they fail to capture margins of Bayesian networks when additional constraints are required. As the first contribution of this paper, we propose new graphical structures that relax mDAGs and are useful to encode the margins of Gaussian Bayesian networks. These structures are applicable to representing the margins of both causal and functional Bayesian networks and remain stable under marginalization. We call these structures the pre-marginalized DAG (pmDAG).

Neural networks are a recent phenomenon in the domain of machine learning. These networks are usually trained using a mechanism called back-propagation. Hinton (Rumelhart et al., 1986; Plaut and Hinton, 1987) popularized the term backpropagation as a means for neural networks to learn the conditional distribution of data. The combination of causal models and neural networks has recently appeared in the literature. For example, Kocaoglu et al. (2017) introduced causality to GANs for getting proper distributions of images based on interdependent image labels. They construct their generative models in a way that is compatible with a causal DAG. Pawlowski et al. (2020) construct upon the ideas in Kocaoglu et al. (2017) and add counterfactual analysis to the research line. Zečević et al. (2021) have used graph neural networks (GNNs) to train their causal Bayesian networks. They incorporated interventions by seeing them as lack of neighborhoods in the graph neural network layers. Among these, the closest to our work is that by Xia et al. (2021). The authors correspond each structural equation with a feed-forward neural network. By doing so, they construct an ADMG that is amenable to back-propagation-based gradient descent optimization. Then, they discuss the problem of causal effect identifiability within this new setting.

As the second contribution of this paper, we show that training a neural network is the dual problem of parameterizing a Gaussian pmDAG. To this end, we introduce medium structures that serve both as a pmDAG and a feed-forward neural network. In particular, we show that our approach is equivalent to the maximum likelihood estimation of the parameter space. Our work extends Xia et al. (2021) in two dimensions:

(i) We consider the whole causal graph to be dual to a neural network. This is also in contrast to, e.g. Pawlowski et al. (2020); Kocaoglu et al. (2017), where each function in the structural system is implemented using a neural network.

(ii) We optimize on the "parameter space" of either functions or Markov kernels, whereas the previous work train their model on the data domain. Unlike the existing works, we do not train our model on each individual observation but rather on the parameter space of the observed marginal distribution.

On these grounds, we note that this article is the first work that shows a general duality between neural networks and causal models and does not provide a mere combination of the two. This is especially true when one notices that this work is easily extensible to discrete distributions, which form a large class of distributions in terms of application. We do not cover this arguably more important extension in this article and restrict attention to Gaussian distributions. As an interesting fact, we do not use any non-linearity in our neural networks when we parameterize for the Gaussian case. This is without loss of generality. We propose a general factorization property, called the "synchronized factorization property," that is applicable in generalizing our solution to other distributions. Furthermore, this article covers both causal (indeterministic) and functional (deterministic) Bayesian networks. We develop a neural network-based algorithm for the parameter optimization of Gaussian pmDAGs but further simplify the naïve neural network algorithm and name the enhanced algorithm the "$SN^2$" algorithm.

After proposing the algorithm for training neural networks/parameterizing Gaussian pmDAGs, we proceed to the [causal effect] identifiability problem (Tian and Pearl, 2002; Tian and Shpitser, 2010; Bareinboim et al., 2020). Do-calculus is the standard mathematical (symbolic) tool for identifying and estimating causal effects (Pearl, 2012). Compatible with do-calculus, we propose a meta-algorithm that checks whether a causal effect is identifiable given a pmDAG and a set of iid observations. This meta-algorithm is almost-surely asymptotically correct.

At the end, we test the performance of our $SN^2$ algorithm on synthetically generated Gaussian data. The $SN^2$ has been implemented in CUDA and parallelizes the optimization procedure of the pmDAG. Given the different basis for this algorithm compared to the existing algorithms, we do not compare our algorithm to the existing ones. Besides this experiment, we test our $SN^2$ algorithm for the problem of causal effect identification on some well-known graphical structures and show the potential of our algorithm in this domain.

## 1.1 Outline

This paper will be organized as follows:

§2 We introduce the basic graphical and probabilistic definitions in this section. Existing and proposed graphical structures will be introduced. We proceed to introduce some general lemmata that we will use throughout this article. We associate this section to the properties of Gaussian graphical structures. We will use the results in this section throughout this article.

§3 In this section, we talk about the potential of the existing graphical structures to act as the backbone of our probabilistic models. We conclude that the existing graphical structures cause a probabilistic problem, namely the marginal instability problem. We conclude that our proposed graphical structure (pmDAG) is more fitting than the existing ones.

§4 This section is where we begin to define the problem formally. We define the maximum likelihood estimation of a graphical structure, based on which we propose an approach to parameterize the graphical structure given a sequence of observations. We relate the maximum likelihood estimation to the Kullback-Leibler divergence.

§5 This section relates two central concepts: the probabilistic graphical models and feed-forward neural networks. We further exploit this relationship to convert the problem of finding the optimal structural system of a graphical structural to the problem of training a feed-forward neural network. In this section, we introduce the "synchronized factorization property"—an interesting probabilistic property of neural networks that allows us to train the neural network.

§6 This is the point where we introduce the neural-network training algorithm. We use some lemmata to show that the neural-network duality of the problem requires no actual neural network implementation and that the $SN^2$ algorithm can be implemented using improved methods. Indeed, neural networks are a ladder that we climb at the beginning of this section and then put aside to develop an efficient algorithm.

§7 In this section, we explain the relationship between our work and the seminal work of Pearl (Pearl and Mackenzie, 2018; Pearl et al., 2000) on structural equation models. Therein, we

talk about the [causal effect] identifiability problem. We relate our results of parameter-optimization in §6 to the identifiability problem and show that the $\mathrm{SN}^2$ algorithm can be utilized to solve the latter problem. We propose a meta-algorithm that is capable of solving the identifiability problem asymptotically.

§8 We divide this section into two parts. In the first part, we measure the performance of our $\mathrm{SN}^2$ algorithm in the forward and backward phases. In the second part, we test our algorithm on observational data and divide the graphical structures into the identifiable and non-identifiable cases. We discuss the results as evidence that our meta-algorithm works.

§9 We conclude our work in this section. We talk about the advantages and disadvantages of our approach in comparison to existing methods. Furthermore, we consider the potential of this work for future research.

## 2 Preliminaries

This section deals with the basic graphical and probabilistic structures that we use throughout this article. In the following, we use a cursive typeface for random variables (e.g. $\mathcal{V}$) and a boldface cursive font for random vectors (set of random variables) (e.g. $\boldsymbol{\mathcal{V}}$). Moreover, we show a probabilistic event of a random variable using italic capital letters (e.g. $X$) and an outcome (actualization of a random variable) using italic small letters (e.g. $x$). For probabilistic events of random vectors, we use bold-italic capital letters (e.g. $\boldsymbol{X}$) and for the corresponding outcomes we use small bold-italic letters (e.g. $\boldsymbol{x}$). We use capital letters with italic typeface (e.g. $W$) for matrices of non-stochastic variables, and small bold-italic letters (e.g. $\boldsymbol{x}$) for vectors of non-stochastic variables. Non-stochastic variables themselves are usually represented using italic small letters (e.g. $x$) but sometimes using italic capital letters (e.g. $I$). This will be clear from the context. Finally, we use $\equiv$ to state that two generic objects are identical.[1]

We denote the probability distribution over a random vector $\boldsymbol{\mathcal{V}}$ using $\mathbb{P}_{\boldsymbol{\mathcal{V}}}$. It is defined over the measurable space of $\boldsymbol{\mathcal{V}}$, which we denote by $\langle \Omega_{\boldsymbol{\mathcal{X}}}, \sigma\Omega_{\boldsymbol{\mathcal{X}}} \rangle$. We write the margin of $\mathbb{P}_{\boldsymbol{\mathcal{V}}}$ over $\boldsymbol{\mathcal{U}} \subseteq \boldsymbol{\mathcal{V}}$ as $\mathbb{P}_{\boldsymbol{\mathcal{U}}}$ and define it as $\mathbb{P}_{\boldsymbol{\mathcal{U}}}(\boldsymbol{U}) := \mathbb{P}_{\boldsymbol{\mathcal{V}}}(\boldsymbol{U} \times \Omega_{\boldsymbol{\mathcal{V}} \setminus \boldsymbol{\mathcal{U}}})$, $\forall\ \boldsymbol{U} \in \sigma\Omega_{\boldsymbol{\mathcal{U}}}$. If $\mathfrak{P}_{\boldsymbol{\mathcal{V}}}$ is an arbitrary set of distributions over the generic random vector $\boldsymbol{\mathcal{V}}$, we define the margin of this set over $\boldsymbol{\mathcal{U}} \subseteq \boldsymbol{\mathcal{V}}$ as $\mathfrak{P}_{\boldsymbol{\mathcal{U}}} := \{\mathbb{P}_{\boldsymbol{\mathcal{U}}} | \mathbb{P}_{\boldsymbol{\mathcal{V}}} \in \mathfrak{P}\}$.

Throughout this article, we may use various notations. To see an overview of our notation, we encourage the reader to refer to Appendix A.

### 2.1 Graphical Definitions

For graphical structures, we use subclasses of *directed acyclic graphs* (DAGs) with latent variables. A DAG is made of a pair $\langle \boldsymbol{\mathcal{A}}, \hookrightarrow \rangle$ composed of a finite random vector $\boldsymbol{\mathcal{A}}$ of (visible or latent) variables that form the nodes of the graph and a homogeneous relation on $\boldsymbol{\mathcal{A}}$ that specifies the presence or absence of a directed edge. If $\mathcal{B} \hookrightarrow \mathcal{C}$ ($\mathcal{B}, \mathcal{C} \in \boldsymbol{\mathcal{A}}$), we say that $\mathcal{B}$ points to $\mathcal{C}$. The relation $\hookrightarrow$ is such that the edges never form a cycle, i.e. no random variable points to itself directly or indirectly; formally, $\mathcal{A} \not\hookrightarrow^i \mathcal{A}$ for all $\mathcal{A} \in \boldsymbol{\mathcal{A}}$ and $i \in \mathbb{N}$, with $\hookrightarrow^i$ being the $i$-th composition of $\hookrightarrow$ with itself. We also define the functions $\mathrm{pa}_{\mathfrak{G}} : \boldsymbol{\mathcal{A}} \to \wp\boldsymbol{\mathcal{A}}$ as $\mathrm{pa}_{\mathfrak{G}}(\mathcal{V}) := \{\mathcal{A} \in \boldsymbol{\mathcal{A}} | \mathcal{A} \hookrightarrow \mathcal{V}\}$ and $\mathrm{ch}_{\mathfrak{G}} : \boldsymbol{\mathcal{A}} \to \wp\boldsymbol{\mathcal{A}}$ as $\mathrm{ch}_{\mathfrak{G}}(\mathcal{V}) := \{\mathcal{A} \in \boldsymbol{\mathcal{A}} | \mathcal{V} \hookrightarrow \mathcal{A}\}$ and respectively name them the *parents* and *children* of

1. Therefore, $\mathcal{A} \equiv \mathcal{B}$ should mean that $\mathcal{A}$ and $\mathcal{B}$ are the same variable, whereas $\mathcal{A} = \mathcal{B}$ means that $\mathcal{A}$ is equal to $\mathcal{B}$ everywhere. One the other hand, $\mathbb{P}(\mathcal{A} = \mathcal{B}) = 1$ indicates that the two variables are equal $\mathbb{P}$-almost everywhere.

$\mathcal{V}$. We extend the definition of parents and children by defining the functions $\mathrm{pa}_{\mathfrak{G}} : \wp\boldsymbol{\mathcal{A}} \to \wp\boldsymbol{\mathcal{A}}$ as $\mathrm{pa}_{\mathfrak{G}}(\boldsymbol{\mathcal{V}}) := \bigcup_{\mathcal{V}\in\boldsymbol{\mathcal{V}}} \mathrm{pa}_{\mathfrak{G}}(\mathcal{V})$ and $\mathrm{ch}_{\mathfrak{G}} : \wp\boldsymbol{\mathcal{A}} \to \wp\boldsymbol{\mathcal{A}}$ as $\mathrm{ch}_{\mathfrak{G}}(\boldsymbol{\mathcal{V}}) := \bigcup_{\mathcal{V}\in\boldsymbol{\mathcal{V}}} \mathrm{ch}_{\mathfrak{G}}(\boldsymbol{\mathcal{V}})$. Moreover, we define $\mathrm{root}(\mathfrak{G}) := \{\mathcal{A} \in \boldsymbol{\mathcal{A}} | \mathrm{pa}_{\mathfrak{G}}(\mathcal{A}) = \varnothing\}$, which returns the set of *root nodes* in the DAG $\mathfrak{G}$. Finally, we define $\mathrm{nroot}(\mathfrak{G}) := \boldsymbol{\mathcal{A}} \smallsetminus \mathrm{root}(\mathfrak{G})$ as the set of *non-roots* of $\mathfrak{G}$.

We cover some of the pre-existing species of DAGs and also introduce a new one that serve the purpose of this article: devising algorithms for parameter-optimization of a DAG given an observation with the Gaussian assumption. Generally, the variables in a DAG are distinguished based on observability. They are considered to be either visible or latent. This assumption forms the basis of many real-world scenarios. This motivates us to define *latent variable DAGs* (lvDAGs) as the broadest class of DAGs that we will work with; see Figure 1 for the Venn diagram of lvDAGs and its subclasses.

**Definition 1 (lvDAG)** *Let $\boldsymbol{\mathcal{A}} \equiv \boldsymbol{\mathcal{V}} \dot\cup \boldsymbol{\mathcal{L}}$ be a random vector (where $\dot\cup$ denotes the union of two disjoint sets) composed of the visible $\boldsymbol{\mathcal{V}}$ and latent $\boldsymbol{\mathcal{L}}$ random vectors. Let $\hookrightarrow$ be a homogeneous relation on $\boldsymbol{\mathcal{A}}$. The pair $\mathfrak{G} = \langle \boldsymbol{\mathcal{A}} \equiv \boldsymbol{\mathcal{V}} \dot\cup \boldsymbol{\mathcal{L}}, \hookrightarrow \rangle$ is a* latent variable DAG *(lvDAG) whenever (a) it is a DAG and (b)* $\mathrm{root}(\mathfrak{G}) \subseteq \boldsymbol{\mathcal{L}}$.

In essence, an lvDAG is a DAG with distinguished observed and latent nodes that has no observed variable within its root nodes. This latter condition is because we want to treat all the lvDAGs and their subclasses in a unified manner. If there is a visible node in the root, it is never impossible to add a latent node that points to that visible node and convert the DAG into an lvDAG. Three lvDAGs are depicted in Figure 2. The latent and visible variables are respectively shown by □ and ○ and there is an edge from $\mathcal{A}$ to $\mathcal{B}$ iff. $\mathcal{A}$ points to $\mathcal{B}$.

**Remark 2** *We generically refer to all of the classes of graphs in this article as DAGs. As mentioned, the variables in a DAG can be arbitrarily visible or latent, and we only distinguish between them when these DAGs are lvDAGs. For those lvDAGs that have no latent variables except for the root nodes with at most one child, i.e. $\boldsymbol{\mathcal{L}} \equiv \mathrm{root}(\mathfrak{G})$ and $\mathrm{card}(\mathrm{ch}_{\mathfrak{G}}(\mathcal{L})) \leq 1$ for all $\mathcal{L} \in \boldsymbol{\mathcal{L}}$, we reserve the term* visible DAG *(vDAG). A vDAG is analogical to a Markovian causal model.*

There are many subclasses of lvDAGs in the literature, including marginal DAGs (mDAGs), acyclic directed mixed graphs (ADMGs), bow-free acyclic path diagrams (BAPs), or visible DAGs (vDAGs) (Evans, 2016; Drton et al., 2009; Maathuis et al., 2018); see Figure 1. Among these, the broadest subclass of lvDAGs is the class of mDAGs (Evans, 2016). The importance of mDAGs is that they are stable under the "marginalization" of Bayesian networks when we are completely agnostic about the nature of the latent space. However, as we will show later on, mDAGs are no longer stable under the marginalization when we make further assumptions about the latent space. This means that we need a different class that is more relaxed than mDAGs. This issue shall be dealt with in detail in §3. For completeness, we will bring two equivalent definitions of mDAGs. Then, we will proceed to introduce the relaxed structures that do not suffer from the aforementioned instability problem.

**Definition 3 (mDAG)** *The following two definitions are equivalent.*

*(i) A marginalized DAG (mDAG) is a triple $\mathfrak{G} = \langle \boldsymbol{\mathcal{V}}, \hookrightarrow, \mathfrak{B} \rangle$, where $\langle \boldsymbol{\mathcal{V}}, \hookrightarrow \rangle$ defines a DAG [of observed variables $\boldsymbol{\mathcal{V}}$], and $\mathfrak{B}$ is an abstract simplicial complex on $\boldsymbol{\mathcal{V}}$. The elements of $\mathfrak{B}$ are called* bidirected faces *(Evans, 2016).*

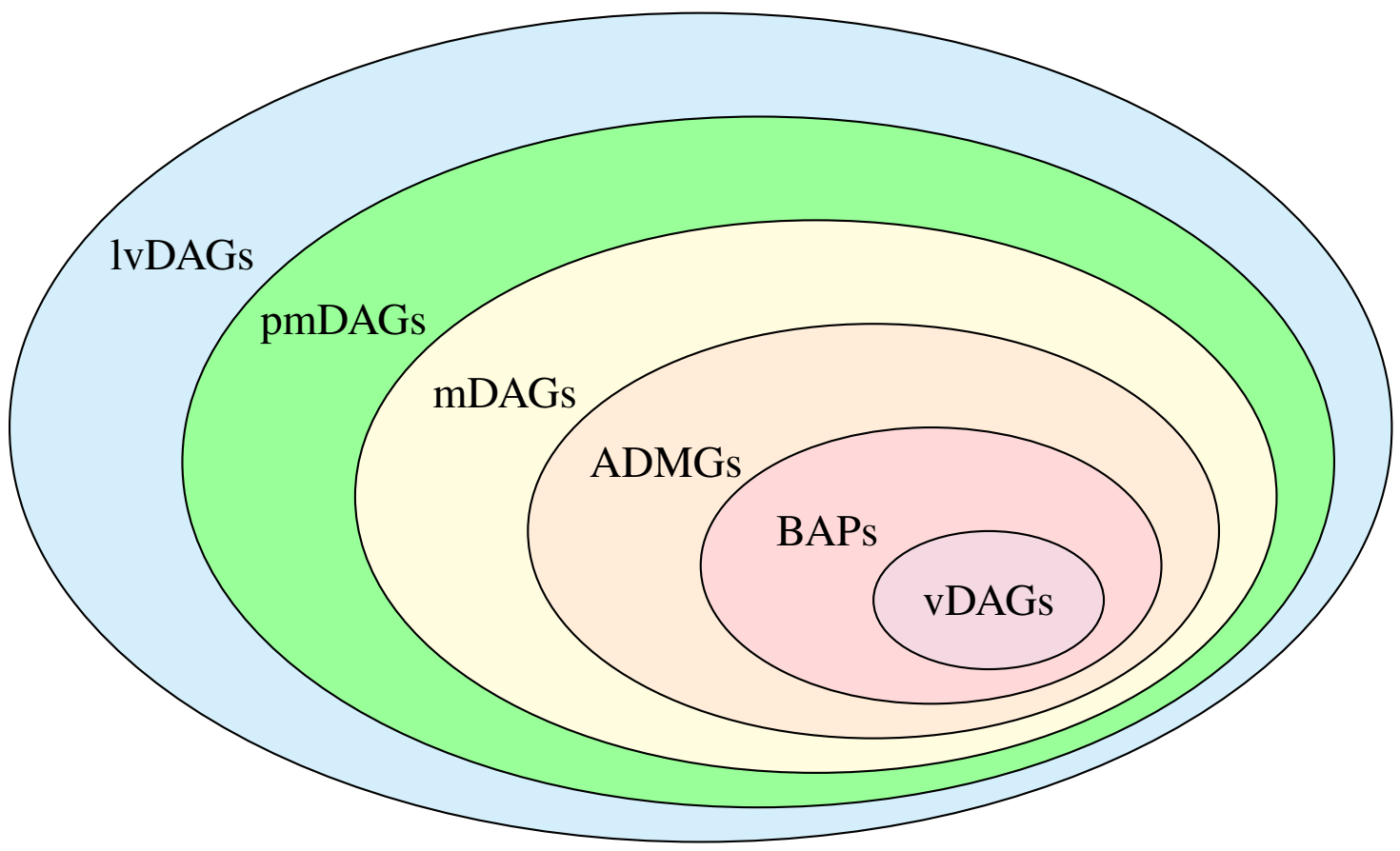

Figure 1: Venn diagram of various classes of latent-variable DAGs. vDAGs $\subset$ BAPs (Drton et al., 2009) $\subset$ ADMGs (Maathuis et al., 2018, p. 47) $\subset$ mDAGs (Evans, 2016) $\subset$ pmDAGs $\subset$ lvDAGs. We primarily work with pmDAGs.

*(ii) Let* $\mathfrak{G} = \langle \boldsymbol{\mathcal{A}} \equiv \boldsymbol{\mathcal{V}} \dot{\cup} \boldsymbol{\mathcal{L}}, \hookrightarrow \rangle$ *be an lvDAG and define the homogeneous relation* $\leq$ *on* $\mathrm{root}(\mathfrak{G})$ *using*

$$\mathcal{I} \leq \mathcal{J} \coloneqq \mathrm{ch}_{\mathfrak{G}}(\mathcal{I}) \subseteq \mathrm{ch}_{\mathfrak{G}}(\mathcal{J}) .$$

*Then,* $\mathfrak{G}$ *is an mDAG iff. (a)* $\boldsymbol{\mathcal{L}} \equiv \mathrm{root}(\mathfrak{G})$ *and (b)* $\langle \mathrm{root}(\mathfrak{G}) , \leq \rangle$ *is an anti-chain.*

Part (ii) of the definition asserts that (a) we require every latent variable to be a root node such that (b) for every root node (latent variable), there is no other root node (latent variable) that points to a subset of its children. An exemplary mDAG is depicted in Figure 2 (a).

**Remark 4** *Punctilious readers are aware that a abstract simplicial complex is inclusive of every subset of the maximal faces. Moreover, every singular subset* $\{\mathcal{V}\} \subseteq \boldsymbol{\mathcal{V}}$ *is in the simplicial complex. On the contrary, our definition excludes the subsets of every maximal faces. Loosely speaking, the second part of Definition 3 is equivalent to the* canonical graphs *of mDAGs in Evans (2016). However, our class of mDAGs (as in Definition 3) does not perfectly align with the canonical DAGs. For example,* ○ *is a canonical DAG in Evans (2016) but not an mDAG here. Conversely,* □⟶○ *is an mDAG according to our definition but is not a canonical DAG in Evans (2016). Nevertheless, one can always form a bijection between the above two definitions, and the results in Evans (2016) that we are concerned with are without loss of generality.*

In §3, we will argue why these structures are not rich enough to capture the margins of all Bayesian networks with extra latent-space assumptions. This lack of richness is a result of the marginal instability of such DAGs. In that respect, we primarily focus on the margins of an important class of Bayesian networks—the Gaussian Bayesian networks. This result justifies using an even richer structure that we call *pre-marginalized DAGs* (pmDAGs). We obtain this richer DAG by relaxing the last constraint in the definition of mDAGs (constraint (b) in Definition 3 (ii)).

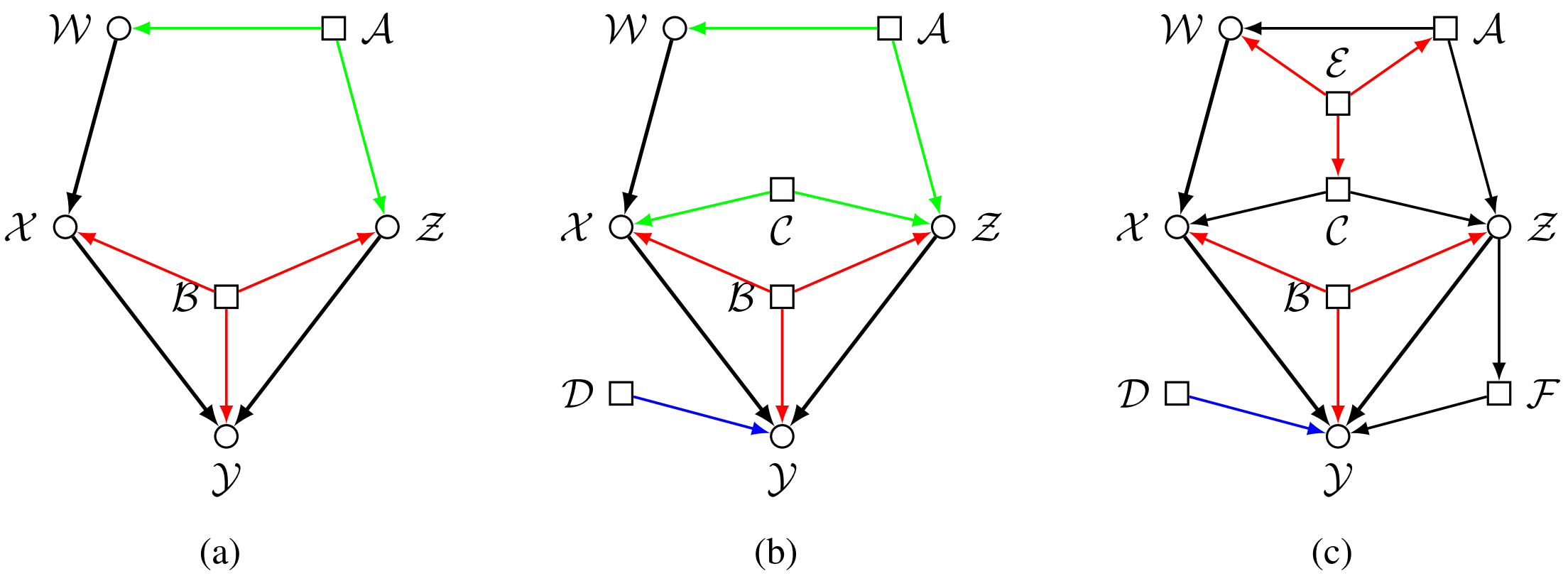

Figure 2: Three lvDAGs. Every □ is a latent variable and every ○ is a visible variable, and there is a directed edge $(\mathcal{I}, \mathcal{J})$ iff. $\mathcal{I}$ points to $\mathcal{J}$ ($\mathcal{I} \hookrightarrow \mathcal{J}$). (a) an mDAG; (b) a pmDAG that is not an mDAG; (c) an lvDAG that is not a pmDAG.

**Definition 5 (pmDAG)** *A pre-marginalized DAG (pmDAG) is an lvDAG* $\mathfrak{G} = \langle \boldsymbol{\mathcal{A}} \equiv \boldsymbol{\mathcal{V}} \dot\cup \boldsymbol{\mathcal{L}}, \hookrightarrow \rangle$ *such that* $\boldsymbol{\mathcal{L}} \equiv \mathrm{root}(\mathfrak{G})$.

From Definitions 1 and 5, it is clear that pmDAGs is a subclass of lvDAGs. From Definitions 5 and 3 (ii) it is readily obvious that mDAGs is a subclass of pmDAGs. Therefore, mDAGs ⊂ pmDAGs ⊂ lvDAGs. We represent the important subclasses of lvDAGs in Figure 1.[2] To see a graphical example of each of these classes, refer to Figure 2.

We conclude our graphical definitions with the *augmentation* of an lvDAG. Augmentation is done by adding an "auxiliary" latent node to as the parent of another node. We will use this operation in various places of this article, especially to relate "deterministic and indeterministic structural systems" of two pmDAGs (see §2.2 for the definitions).

**Definition 6 (augmentation)** *Let* $\mathfrak{G} = \langle \boldsymbol{\mathcal{A}} \equiv \boldsymbol{\mathcal{V}} \dot\cup \boldsymbol{\mathcal{L}}, \hookrightarrow \rangle$ *be an lvDAG. The* augmentation *of* $\mathfrak{G}$ *w.r.t. the random variable* $\mathcal{A} \in \boldsymbol{\mathcal{A}}$, *denoted by* $\iota_{\mathcal{A}}(\mathfrak{G})$, *is an lvDAG* $\langle \boldsymbol{\mathcal{A}}' \equiv \boldsymbol{\mathcal{V}} \dot\cup \boldsymbol{\mathcal{L}}', \hookrightarrow' \rangle$ *constructed as follows:*

*(a) form* $\boldsymbol{\mathcal{L}}' \equiv \boldsymbol{\mathcal{L}} \dot\cup \{\mathcal{L}\}$ *by adding a latent variable* $\mathcal{L} \notin \boldsymbol{\mathcal{A}}$ *to* $\boldsymbol{\mathcal{L}}$,

*(b) define* $\hookrightarrow'$ *on* $\boldsymbol{\mathcal{A}}'$ *in such a way that* $\hookrightarrow$ *is its restriction,*

*(c) finally, make* $\mathcal{L}$ *point only to* $\mathcal{A}$ *and do not allow any* $\mathcal{X} \in \boldsymbol{\mathcal{A}}$ *to point to* $\mathcal{L}$.

*The augmentation of* $\mathfrak{G}$ *w.r.t. the random vector* $\boldsymbol{\mathcal{U}} \subseteq \boldsymbol{\mathcal{A}}$, $\iota_{\boldsymbol{\mathcal{U}}}(\mathfrak{G})$, *is done by augmenting* $\mathfrak{G}$ *once for each* $\mathcal{U} \in \boldsymbol{\mathcal{U}}$ *(in an arbitrary order).*

See an example of augmentation in Figures 3 (a) and (b). We call the added latent variable $\mathcal{L}$ the *auxiliary parent* of $\mathcal{A}$. In order to distinguish this variable, we define the function $\mathrm{aux}_{\mathfrak{G},\mathfrak{G}'} : \boldsymbol{\mathcal{A}} \to \boldsymbol{\mathcal{A}}'$ as $\mathrm{aux}_{\mathfrak{G},\mathfrak{G}'}(\mathcal{V}) := \mathrm{pa}_{\mathfrak{G}'}(\mathcal{V}) \setminus \mathrm{pa}_{\mathfrak{G}}(\mathcal{V})$.[3]

2. Similar to what we have seen with respect to the mDAGs (Remark 4) and its canonical graph, these structures are conventionally defined using (hyper-)edges. In contrast, we consider that they have latent variables with direct edges toward the visible variables. This does not cause any discrepancy as there is a bijection between these structures and the conventional definitions. In the literature some names have been given to structures similar to the latent-variable DAGs, including "canonical graph" (Evans, 2016) or "augmented graph" (Forré and Mooij, 2017).

3. We treat a set of one random variable as a single random variable.

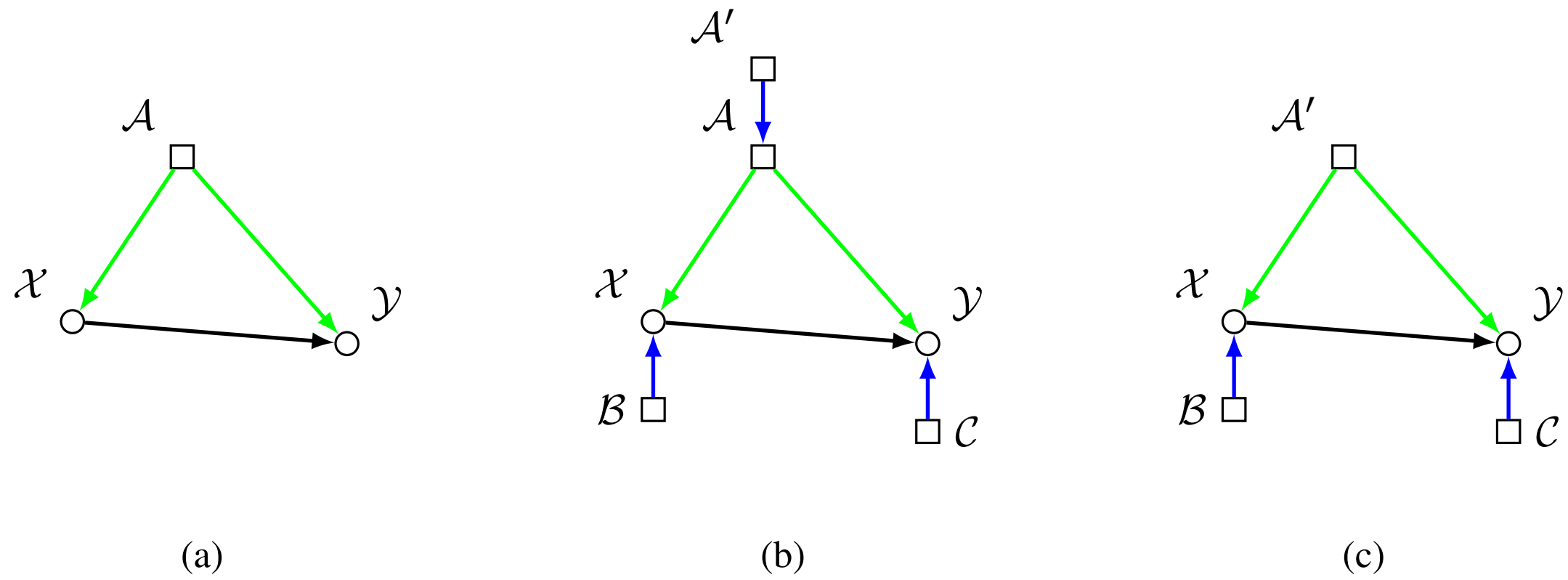

Figure 3: Augmentation (Definition 6) and exogenization (Definition 13) of an lvDAG. (a) an mDAG $\mathfrak{G} = \langle \boldsymbol{\mathcal{A}} \equiv \boldsymbol{\mathcal{V}} \dot\cup \boldsymbol{\mathcal{L}}, \hookrightarrow \rangle$, (b) its augmentation w.r.t. $\boldsymbol{\mathcal{A}}$, i.e. $\iota_{\boldsymbol{\mathcal{A}}}(\mathfrak{G})$, which is an lvDAG, and (c) the deterministic exogenization of $\iota_{\boldsymbol{\mathcal{A}}}(\mathfrak{G})$ w.r.t. $\mathrm{root}(\mathfrak{G})$, i.e. $\mathrm{T}_{\mathrm{root}(\mathfrak{G})}(\iota_{\boldsymbol{\mathcal{A}}}(\mathfrak{G}))$.

## 2.2 Probabilistic Definitions

An lvDAG is the "backbone" that determines functional (deterministic) or probabilistic (indeterministic) relationships between the random variables. The deterministic relationships resemble structural equation models while the indeterministic relationships revoke Bayesian networks. We call these relationships "deterministic structural systems" and "indeterministic structural systems", respectively.[4] Deterministic structural systems are a joint probability distribution of the root nodes paired with a set of functions constructed upon a DAG. Indeterministic structural systems, on the other hand, are a set of Markov kernels. Indeterministic structural systems are the *de facto* generalization of deterministic structural systems and we will use them in place of deterministic structural systems. Indeed, we sometimes "indeterminate" deterministic structural systems before working with them (Definition 12).

We define deterministic structural systems in the following, and then generalize them to the indeterministic structural systems.

**Definition 7 ([deterministic] structural system)** *Let* $\mathfrak{G} = \langle \boldsymbol{\mathcal{A}} \equiv \boldsymbol{\mathcal{V}} \dot\cup \boldsymbol{\mathcal{L}}, \hookrightarrow \rangle$ *be an lvDAG. A* [deterministic] structural system of $\mathfrak{G}$ *is a pair* $\left\langle \mathbb{P}_{\mathrm{root}(\mathfrak{G})}, \Phi_{\mathrm{nroot}(\mathfrak{G})} \right\rangle$ *composed of a probability distribution* $\mathbb{P}_{\mathrm{root}(\mathfrak{G})}$ *w.r.t. which* $\mathrm{root}(\mathfrak{G})$ *is mutually independent and a set* $\Phi_{\mathrm{nroot}(\mathfrak{G})}$ *of (measurable) functions* $\Phi_{\mathcal{N}} : \Omega_{\mathrm{pa}_{\mathfrak{G}}(\mathcal{N})} \to \Omega_{\mathcal{N}}$ *that is indexed by* $\mathrm{nroot}(\mathfrak{G})$. *We abbreviate deterministic structural system to "structural system" and write* $\langle \mathbb{P}, \Phi \rangle$ *as a shorthand for* $\left\langle \mathbb{P}_{\mathrm{root}(\mathfrak{G})}, \Phi_{\mathrm{nroot}(\mathfrak{G})} \right\rangle$.

*We call the set of all structural systems of* $\mathfrak{G}$ *the* deterministic structural system model *(DSM) of* $\mathfrak{G}$ *and denote it using* $\mathfrak{D}(\mathfrak{G})$.

4. In this work, structural systems—deterministic or otherwise—convey no causal meaning on their own, unless stated otherwise. See Remark 10.

A structural system *induces* a probability distribution $\mathbb{P}_{\boldsymbol{\mathcal{A}}}$ using the following system of equations:

$$\mathcal{N} = \Phi_{\mathcal{N}}(\mathrm{pa}_{\mathfrak{G}}(\mathcal{N}))\,, \qquad\qquad \forall\ \mathcal{N} \in \mathrm{nroot}(\mathfrak{G})\,, \tag{1}$$

where the joint probability $\mathbb{P}_{\boldsymbol{\mathcal{A}}}$ is obtained by recursively invoking the push-backs of functions $\Phi_{\mathcal{N}}$.

**Definition 8 (indeterministic structural system)** *Let* $\mathfrak{G} = \langle \boldsymbol{\mathcal{A}} \equiv \boldsymbol{\mathcal{V}} \dot\cup \boldsymbol{\mathcal{L}}, \hookrightarrow \rangle$ *be an lvDAG. An* indeterministic structural system *of* $\mathfrak{G}$ *is a set* $\left\{\kappa_{\mathcal{A}|\mathrm{pa}_{\mathfrak{G}}(\mathcal{A})} : \sigma\Omega_{\mathcal{A}} \times \Omega_{\mathrm{pa}_{\mathfrak{G}}(\mathcal{A})} \to [0,1]\right\}_{\mathcal{A}\in\boldsymbol{\mathcal{A}}}$ *of Markov kernels indexed by* $\boldsymbol{\mathcal{A}}$ *(where* $\Omega_{\varnothing} = \{0\}$*). We denote it using* $\{\kappa_{\mathcal{A}}\}_{\mathcal{A}\in\boldsymbol{\mathcal{A}}}$ *as a shorthand.*

*We call the collection of all indeterministic structural systems of an lvDAG* $\mathfrak{G}$ *the* indeterministic structural system model *(ISM) of* $\mathfrak{G}$ *and denote it using* $\mathfrak{d}(\mathfrak{G})$.

Given an indeterministic structural system $\{\kappa_{\mathcal{A}}\}_{\mathcal{A}\in\boldsymbol{\mathcal{A}}}$, the following equation, called the *recursive factorization*, induces a unique probability measure $\mathbb{P}_{\boldsymbol{\mathcal{A}}}$ on $\boldsymbol{\mathcal{A}}$:

$$\mathbb{P}_{\boldsymbol{\mathcal{A}}}(\mathrm{d}\boldsymbol{x}) = \prod_{\mathcal{A}\in\boldsymbol{\mathcal{A}}} \kappa_{\mathcal{A}}\big(\mathrm{d}x_{\mathcal{A}} \,\big|\, \boldsymbol{x}_{\mathrm{pa}_{\mathfrak{G}}(\mathcal{V})}\big) \tag{2}$$

for all $\boldsymbol{x} = \Omega_{\boldsymbol{\mathcal{A}}}$ (Cowell et al. (2007, Ch. 5.3) and Forré and Mooij (2017, Theorem 3.2.1)).

**Remark 9** *As readers might have acknowledged, we did not explicitly associate a measurable space to the random variables when defining lvDAGs. The probability space depends on the probabilistic assumptions that we hold and gets defined through the deterministic or indeterministic structural system. Therefore, when defining an lvDAG we make no assumption about its nodes except for them being random variables (random elements) in the broad sense.*

Note that, conversely, we can also uniquely define an lvDAG $\mathfrak{G}$ given a deterministic structural system $\langle \mathbb{P}, \Phi \rangle$ or an indeterministic structural system $\{\kappa_{\mathcal{A}}\}_{\mathcal{A}\in\boldsymbol{\mathcal{A}}}$. In that case, we call $\mathfrak{G}$ the corresponding lvDAG of $\langle \mathbb{P}, \Phi \rangle$ or $\{\kappa_{\mathcal{A}}\}_{\mathcal{A}\in\boldsymbol{\mathcal{A}}}$.

**Remark 10** *Despite their similarity, we do not take a structural system as synonymous with the structural "causal" model (SCM). Nor do we think of indeterministic structural systems as "causal" Bayesian networks. No causal assumption is needed for finding the optimal (indeterministic) structural system. For our solution to work, the DAG does not have to be* correspondent *(David, 2002); that is, correspond to the structure of some true/ground-truth causal mechanism/structural causal model. We associate §7 entirely with how our work is related to the causal identifiability problem. For the distinction between causal and non-causal Bayesian networks as well as SCMs, see Pearl et al. (2000) or Bareinboim et al. (2020).*

Next, we enter upon defining the *probabilistic models* of an lvDAG. Parallel to the structural system models, we will have probabilistic models for an lvDAG.

**Definition 11 (probabilistic model)** *Let* $\mathfrak{G} = \langle \boldsymbol{\mathcal{A}} \equiv \boldsymbol{\mathcal{V}} \dot\cup \boldsymbol{\mathcal{L}}, \hookrightarrow \rangle$ *be an lvDAG. We call the set of probability distributions over* $\boldsymbol{\mathcal{A}}$ *that are induced by a deterministic (resp. indeterministic) structural system of* $\mathfrak{G}$*, as per Equation 1 (resp. Equation 2), the* deterministic *(resp.* indeterministic*)* probabilistic model *(DPM (resp. IPM)) of* $\mathfrak{G}$. *We denote this set using* $\mathfrak{Q}(\mathfrak{G})$ *(resp.* $\mathfrak{q}(\mathfrak{G})$*).*

*We show the set of the margins of the members of* $\mathfrak{Q}(\mathfrak{G})$ *(resp.* $\mathfrak{q}(\mathfrak{G})$*) over* $\boldsymbol{\mathcal{U}} \subseteq \boldsymbol{\mathcal{A}}$ *using* $\mathfrak{Q}_{\boldsymbol{\mathcal{U}}}(\mathfrak{G})$ *(resp.* $\mathfrak{q}_{\boldsymbol{\mathcal{U}}}(\mathfrak{G})$*). Formally,* $\mathfrak{Q}_{\boldsymbol{\mathcal{U}}}(\mathfrak{G}) \coloneqq (\mathfrak{Q}(\mathfrak{G}))_{\boldsymbol{\mathcal{U}}}$ *(resp.* $\mathfrak{q}_{\boldsymbol{\mathcal{U}}}(\mathfrak{G}) \coloneqq (\mathfrak{q}(\mathfrak{G}))_{\boldsymbol{\mathcal{U}}}$*). In particular, we call* $\mathfrak{Q}_{\boldsymbol{\mathcal{V}}}(\mathfrak{G})$ *(resp.* $\mathfrak{q}_{\boldsymbol{\mathcal{V}}}(\mathfrak{G})$*) the* marginal DPM *(resp.* marginal IPM*) of the lvDAG* $\mathfrak{G} = \langle \boldsymbol{\mathcal{A}} \equiv \boldsymbol{\mathcal{V}} \dot\cup \boldsymbol{\mathcal{L}}, \hookrightarrow \rangle$.[5]

We denote the mapping that takes a structural system to its induced probability using $\Pi : \mathfrak{D}(\mathfrak{G}) \to \mathfrak{Q}(\mathfrak{G})$, where $\mathfrak{D}(\mathfrak{G})$ is the DSM and $\mathfrak{Q}(\mathfrak{G})$ is the DPM. Therefore, for $\mathfrak{G} = \langle \boldsymbol{\mathcal{A}} \equiv \boldsymbol{\mathcal{V}} \dot\cup \boldsymbol{\mathcal{L}}, \hookrightarrow \rangle$, if the probability measure $\mathbb{P}_{\boldsymbol{\mathcal{A}}}$ is induced by $\langle \mathbb{P}, \Phi \rangle$, we may write $\mathbb{P}_{\boldsymbol{\mathcal{A}}} = \Pi(\langle \mathbb{P}, \Phi \rangle)$. Similar to the structural system, we denote the mapping that takes an indeterministic structural system to its induced probability measure using $\Pi : \mathfrak{d}(\mathfrak{G}) \to \mathfrak{q}(\mathfrak{G})$, with $\mathfrak{d}(\mathfrak{G})$ being the ISM and $\mathfrak{q}(\mathfrak{G})$ being the IPM. Therefore, for the measure $\mathbb{P}_{\boldsymbol{\mathcal{A}}}$ induced by $\{\kappa_{\mathcal{A}}\}_{\mathcal{A} \in \boldsymbol{\mathcal{A}}}$ we may write $\mathbb{P}_{\boldsymbol{\mathcal{A}}} = \Pi(\{\kappa_{\mathcal{A}}\}_{\mathcal{A} \in \boldsymbol{\mathcal{A}}})$.

Now, we consider a straight-forward way that converts a structural system of $\mathfrak{G}$ into the indeterministic structural system of the same lvDAG. This transformation unifies some of the subsequent definitions and theorems.

**Definition 12 (indetermination)** *Let $\mathfrak{G} = \langle \boldsymbol{\mathcal{A}} \equiv \boldsymbol{\mathcal{V}} \dot\cup \boldsymbol{\mathcal{L}}, \hookrightarrow \rangle$ be an lvDAG. We define the function $\mathrm{I}_{\mathfrak{G}} : \mathfrak{D}(\mathfrak{G}) \to \mathfrak{d}(\mathfrak{G})$ as the one that takes in a structural system $\langle \mathbb{P}, \Phi \rangle$ and returns an indeterministic structural system $\{\kappa_{\mathcal{A}}\}_{\mathcal{A} \in \boldsymbol{\mathcal{A}}}$ the Markov kernels of which are defined as follows:*

$$\kappa_{\mathcal{A}}(X \,|\, \boldsymbol{y}) := \begin{cases} \mathbb{P}_{\mathcal{A}}(X) & \mathcal{A} \in \mathrm{root}(\mathfrak{G}), \\ \mathbb{1}_X(\Phi_{\mathcal{A}}(\boldsymbol{y})) & \mathcal{A} \in \mathrm{nroot}(\mathfrak{G}), \end{cases} \tag{3}$$

*for every $X \in \sigma\Omega_{\mathcal{A}}, \boldsymbol{y} \in \Omega_{\mathrm{pa}_{\mathfrak{G}}(\mathcal{A})}$ for all $\mathcal{A} \in \boldsymbol{\mathcal{A}}$ (where $\Omega_{\varnothing} = \{0\}$). Here, $\mathbb{1}$ is the indicator function. We call the mapping $\mathrm{I}_{\mathfrak{G}}$ the* indetermination of $\mathfrak{G}$.

In words, indetermination takes in the structural system and converts it to an indeterministic structural system by forming a Dirac measure for each value of the parents of non-root nodes. It is therefore easy to see that a structural system and its indetermination are in essence the same thing, each of which hand in a different tool (i.e. Equations 1 or 2) to work with.

## 2.3 Gaussian lvDAGs

In above, we introduced some of the necessary notions for this article. We emphasized that a deterministic or indeterministic structural system may be associated to an lvDAG and that they induce a probability distribution. From this point and throughout this article, we restrict attention to the induced distributions that are (jointly) Gaussian. We denote the set of all jointly Gaussian distributions on a generic random vector $\boldsymbol{\mathcal{V}}$ using $\mathfrak{P}^{\mathrm{G}}_{\boldsymbol{\mathcal{V}}}$. We assume that these distributions have the Gaussian density to simplify some theorems that in general hold only almost everywhere. Based on the discussions in Drton et al. (2009), we consider that assuming zero means for the latent and observational space is without loss of generality. We define $\mathfrak{P}^{\mathrm{G}(0)}_{\boldsymbol{\mathcal{V}}} := \left\{ \mathbb{P}_{\boldsymbol{\mathcal{V}}} \in \mathfrak{P}^{\mathrm{G}}_{\boldsymbol{\mathcal{V}}} \,\middle|\, \mathrm{mean}(\boldsymbol{\mathcal{V}}; \mathbb{P}_{\boldsymbol{\mathcal{V}}}) = \mathbf{0} \right\}$ as the set of these distributions.

For a generic lvDAG, $\mathfrak{G} = \langle \boldsymbol{\mathcal{A}} \equiv \boldsymbol{\mathcal{V}} \dot\cup \boldsymbol{\mathcal{L}}, \hookrightarrow \rangle$ we call $\mathfrak{Q}^{\mathrm{G}(0)}(\mathfrak{G}) := \mathfrak{P}^{\mathrm{G}(0)}_{\boldsymbol{\mathcal{A}}} \cap \mathfrak{Q}(\mathfrak{G})$ its *Gaussian DPM*. Moreover, we call

$$\mathfrak{D}^{\mathrm{G}(0)}(\mathfrak{G}) := \left\{ \langle \mathbb{P}, \Phi \rangle \in \mathfrak{D}(\mathfrak{G}) \,\middle|\, \Pi(\langle \mathbb{P}, \Phi \rangle) \in \mathfrak{P}^{\mathrm{G}(0)}_{\boldsymbol{\mathcal{A}}} \right\}$$

the *Gaussian DSM* of $\mathfrak{G}$. The Gaussian DSM is the set of all structural systems that induce a Gaussian distribution with zero means. Given the Gaussian density, every structural system $\langle \mathbb{P}, \Phi \rangle$

5. A distribution in the marginal IPM/DPM is sometimes referred to as "observational" distribution in the context of causal inference.

in Definition 11 that induces a $\mathbb{P}_{\boldsymbol{\mathcal{A}}} \in \mathfrak{Q}(\mathfrak{G})$ is such that $\Phi$ are linear and $\text{root}(\mathfrak{G})$ is $\mathbb{P}$-jointly Gaussian. We therefore define:

$$\mathfrak{D}^{\text{lG}(0)}(\mathfrak{G}) = \left\{ \langle \mathbb{P}, \Phi \rangle \in \mathfrak{D}(\mathfrak{G}) \,\middle|\, \mathbb{P} \in \mathfrak{P}^{\text{G}(0)}_{\text{root}(\mathfrak{G})} \text{ and } \Phi \text{ are linear transformations} \right\}.$$

and assert that $\mathfrak{D}^{\text{lG}(0)}(\mathfrak{G}) = \mathfrak{D}^{\text{G}(0)}(\mathfrak{G})$. To see why this is the case, refer to Appendix B.1 (Theorem 44).

Now that we identified the structural systems in terms of a root distribution and non-root functions, it is appropriate that we also specify every Gaussian structural system using a set of parameters. Assume that $\mathfrak{G} = \langle \boldsymbol{\mathcal{A}} \equiv \boldsymbol{\mathcal{V}} \dot\cup \boldsymbol{\mathcal{L}}, \hookrightarrow \rangle$ is an lvDAG. We specify $\mathbb{P}_{\text{root}(\mathfrak{G})}$ by the variance of each variable $\mathcal{A} \in \text{root}(\mathfrak{G})$, i.e. $\bar{\sigma}^2_{\mathcal{A}} \in [0, \infty)$. We specify the functions $\Phi$ using a set of vectors $\bar{\boldsymbol{w}} = \{\bar{w}_{\mathcal{V}}\}_{\mathcal{V} \in \text{nroot}(\mathfrak{G})}$, where $\bar{w}_{\mathcal{V}} \in \mathbb{R}^{\text{card}(\text{pa}_{\mathfrak{G}}(\mathcal{V}))}$, such that $\Phi_{\mathcal{V}}(\text{pa}_{\mathfrak{G}}(\mathcal{V})) = \bar{w}_{\mathcal{V}} \cdot \text{pa}_{\mathfrak{G}}(\mathcal{V})$, with "$\cdot$" being the inner product. Note that, according to Definition 7, the covariance of each two root nodes is always zero.

It is well-known that the marginal IPM of an lvDAG $\mathfrak{G} = \langle \boldsymbol{\mathcal{A}} \equiv \boldsymbol{\mathcal{V}} \dot\cup \boldsymbol{\mathcal{L}}, \hookrightarrow \rangle$ and the marginal DPM of its augmentation w.r.t. $\boldsymbol{\mathcal{A}}$ are the same. One might therefore ask whether the same is true for the Gaussian IPM and DPM of the two lvDAGs. Similar to what we defined above, for a generic lvDAG, $\mathfrak{G} = \langle \boldsymbol{\mathcal{A}} \equiv \boldsymbol{\mathcal{V}} \dot\cup \boldsymbol{\mathcal{L}}, \hookrightarrow \rangle$ we call $\mathfrak{q}^{\text{G}(0)}(\mathfrak{G}) = \mathfrak{P}^{\text{G}(0)}_{\boldsymbol{\mathcal{A}}} \cap \mathfrak{q}(\mathfrak{G})$ its *Gaussian IPM*. Moreover, we call

$$\mathfrak{d}^{\text{G}(0)}(\mathfrak{G}) = \left\{ \{\kappa_{\mathcal{A}}\}_{\mathcal{A} \in \boldsymbol{\mathcal{A}}} \in \mathfrak{d}(\mathfrak{G}) \,\middle|\, \Pi(\{\kappa_{\mathcal{A}}\}_{\mathcal{A} \in \boldsymbol{\mathcal{A}}}) \in \mathfrak{P}^{\text{G}(0)}_{\boldsymbol{\mathcal{A}}} \right\}$$

the *Gaussian ISM* of $\mathfrak{G}$. Indeed, we can always say that the Gaussian IPM of $\mathfrak{G}$ and the Gaussian DPM of $\mathfrak{G}' = \iota_{\boldsymbol{\mathcal{A}}}(\mathfrak{G})$ are the same. That is,

$$\mathfrak{q}^{\text{G}(0)}(\mathfrak{G}) = \mathfrak{Q}^{\text{G}(0)}_{\boldsymbol{\mathcal{A}}}(\mathfrak{G}'). \tag{4}$$

We show this equivalence in Appendix B.1 (Theorem 48). This hints us that the Gaussian DSM and Gaussian ISM are as powerful as each other, and that the theorems that we prove for the IPM of an lvDAG $\mathfrak{G} = \langle \boldsymbol{\mathcal{A}} \equiv \boldsymbol{\mathcal{V}} \dot\cup \boldsymbol{\mathcal{L}}, \hookrightarrow \rangle$ apply to the DPM of its augmentation w.r.t. $\boldsymbol{\mathcal{A}}$, vice versa.

## 3 Marginal Stability of Gaussian lvDAGs

In this section and before we proceed to the main objective of this article, we establish that generalizing mDAGs to pmDAGs is justified. This amelioration is necessary if we want to parametrize an important class of structural systems that correspond to the Gaussian probabilistic model. In particular, we show that in contrast to our pmDAG structure, margins of mDAGs for this class are not stable when we remove latent variables from lvDAGs to form them. To give you an overview, we consider the marginal Gaussian IPM of mDAGs. By using the natural operation that removes latent variables from an lvDAG to form this mDAG, we observe that the marginal IPM changes. Moreover, we show that as soon as we assume that there is no directed edge between the observed variables, there appear Gaussian margins that no mDAG can induce. That is, there are margins that the marginal Gaussian IPM of no mDAG in this class contains. This leads us to the main result of this section: pmDAGs are better modeling tools than mDAGs when Gaussianity is an assumption.

Two important ingredients that lead to our desired result are the *exogenization* and *coalescence* of an lvDAG, which we take from Evans (2016) and generalize. As opposed to the original definition, we will have two versions of exogenization: deterministic and indeterministic. The original definition of exogenization is the indeterministic one and associates to the IPM, whereas the deterministic one associates to the DPM. The definition of coalescence is the same as the original definition.

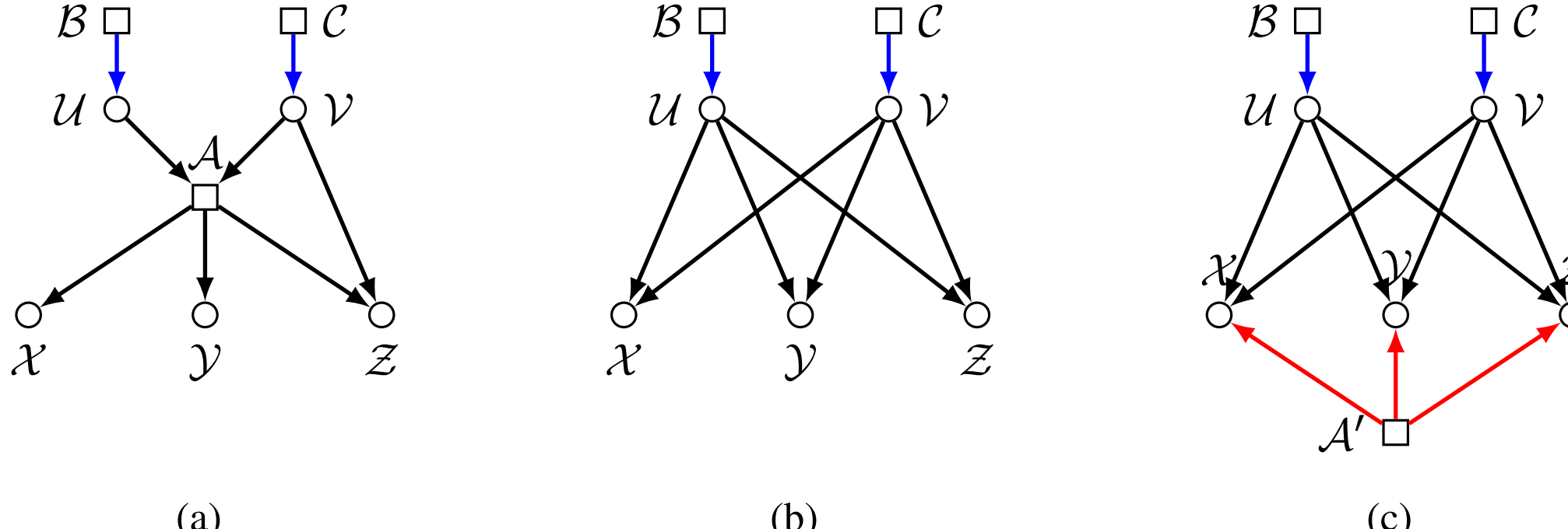

Figure 4: Exogenization of an lvDAG w.r.t. a variable. (a) an lvDAG $\mathfrak{G}$, (b) its deterministic exogenization w.r.t. $\mathcal{A}$, i.e. $\mathrm{T}_{\mathcal{A}}(\mathfrak{G})$, and (c) its indeterministic exogenization w.r.t. $\mathcal{A}$, i.e. $\tau_{\mathcal{A}}(\mathfrak{G})$.

**Definition 13 ((in)deterministic exogenization)** *Let* $\mathfrak{G} = \langle \boldsymbol{\mathcal{A}} \equiv \boldsymbol{\mathcal{V}} \dot\cup \boldsymbol{\mathcal{L}}, \hookrightarrow \rangle$ *be an lvDAG. The* deterministic exogenization *(resp.* indeterministic exogenization*) of* $\mathfrak{G}$ *w.r.t. the variable* $\mathcal{L} \in \boldsymbol{\mathcal{L}} \cap \mathrm{nroot}(\mathfrak{G})$*, denoted by* $\mathrm{T}_{\mathcal{L}}(\mathfrak{G})$ *(resp.* $\tau_{\mathcal{L}}(\mathfrak{G})$*), is an lvDAG* $\langle \boldsymbol{\mathcal{A}}' \equiv \boldsymbol{\mathcal{V}} \dot\cup \boldsymbol{\mathcal{L}}', \hookrightarrow' \rangle$ *constructed as follows:*

*(a) let* $\mathfrak{G}'' = \langle \boldsymbol{\mathcal{A}}'' \equiv \boldsymbol{\mathcal{V}} \dot\cup \boldsymbol{\mathcal{L}}'', \hookrightarrow'' \rangle$ *be equivalent to* $\mathfrak{G}$ *(resp. be the augmentation of* $\mathfrak{G}$ *w.r.t.* $\mathcal{L}$*),*

*(b) let* $\boldsymbol{\mathcal{L}}' \equiv \boldsymbol{\mathcal{L}}'' \setminus \{\mathcal{L}\}$ *and* $\hookrightarrow'$ *be the restriction of* $\hookrightarrow''$ *over* $\boldsymbol{\mathcal{L}}'$*,*

*(c) make the parents of* $\mathcal{L}$ *point to its children, i.e. for each* $\mathcal{A}, \mathcal{B} \in \boldsymbol{\mathcal{A}}''$*, if* $\mathcal{A} \hookrightarrow \mathcal{L}$ *and* $\mathcal{L} \hookrightarrow \mathcal{B}$*, then let* $\mathcal{A} \hookrightarrow'' \mathcal{B}$*.*

*The deterministic exogenization (resp. indeterministic exogenization) of* $\mathfrak{G}$ *w.r.t. the random vector* $\boldsymbol{\mathcal{U}} \subseteq \boldsymbol{\mathcal{L}} \cap \mathrm{nroot}(\mathfrak{G})$*, is done by deterministic exogenization (resp. indeterministic exogenization) of* $\mathfrak{G}$ *once w.r.t. each* $\mathcal{U} \in \boldsymbol{\mathcal{U}}$ *(in an arbitrary order).*

Deterministic exogenization removes a latent non-root node from the lvDAG. Indeterministic exogenization also adds a latent root node in its place. Figure 4 demonstrates deterministic and indeterministic exogenization of a pmDAG w.r.t. a single variable. Now, we are in the position to talk about coalescence; a graph modification that, unlike exogenization, is applied on the root nodes.

**Definition 14 (coalescence)** *Let* $\mathfrak{G} = \langle \boldsymbol{\mathcal{A}}' \equiv \boldsymbol{\mathcal{V}} \dot\cup \boldsymbol{\mathcal{L}}, \hookrightarrow \rangle$ *be an lvDAG. The coalescence of* $\mathfrak{G}$ *w.r.t. the variable* $\mathcal{L} \in \boldsymbol{\mathcal{L}} \cap \mathrm{root}(\mathfrak{G})$*, denoted by* $\delta_{\mathcal{L}}(\mathfrak{G})$*, is an lvDAG* $\langle \boldsymbol{\mathcal{A}} \equiv \boldsymbol{\mathcal{V}} \dot\cup \boldsymbol{\mathcal{L}}', \hookrightarrow' \rangle$ *constructed as follows:*

*(a) if there is a variable* $\mathcal{L}' \in \mathrm{root}(\mathfrak{G})$ *(*$\mathcal{L}' \not\equiv \mathcal{L}$*) such that* $\mathrm{ch}(\mathcal{L}) \subseteq \mathrm{ch}(\mathcal{L}')$ *then let* $\boldsymbol{\mathcal{L}}' \equiv \boldsymbol{\mathcal{L}} \setminus \{\mathcal{L}\}$ *otherwise, let* $\boldsymbol{\mathcal{L}}' \equiv \boldsymbol{\mathcal{L}}$*,*

*(b) then, let* $\hookrightarrow'$ *be the restriction of* $\hookrightarrow$ *over* $\boldsymbol{\mathcal{A}}'$*.*

*The coalescence of* $\mathfrak{G}$ *w.r.t. the random vector* $\boldsymbol{\mathcal{U}} \subseteq \boldsymbol{\mathcal{L}} \cap \mathrm{nroot}(\mathfrak{G})$*, is done by coalescence of* $\mathfrak{G}$ *once w.r.t. each* $\mathcal{U} \in \boldsymbol{\mathcal{U}}$ *(in an arbitrary order).*

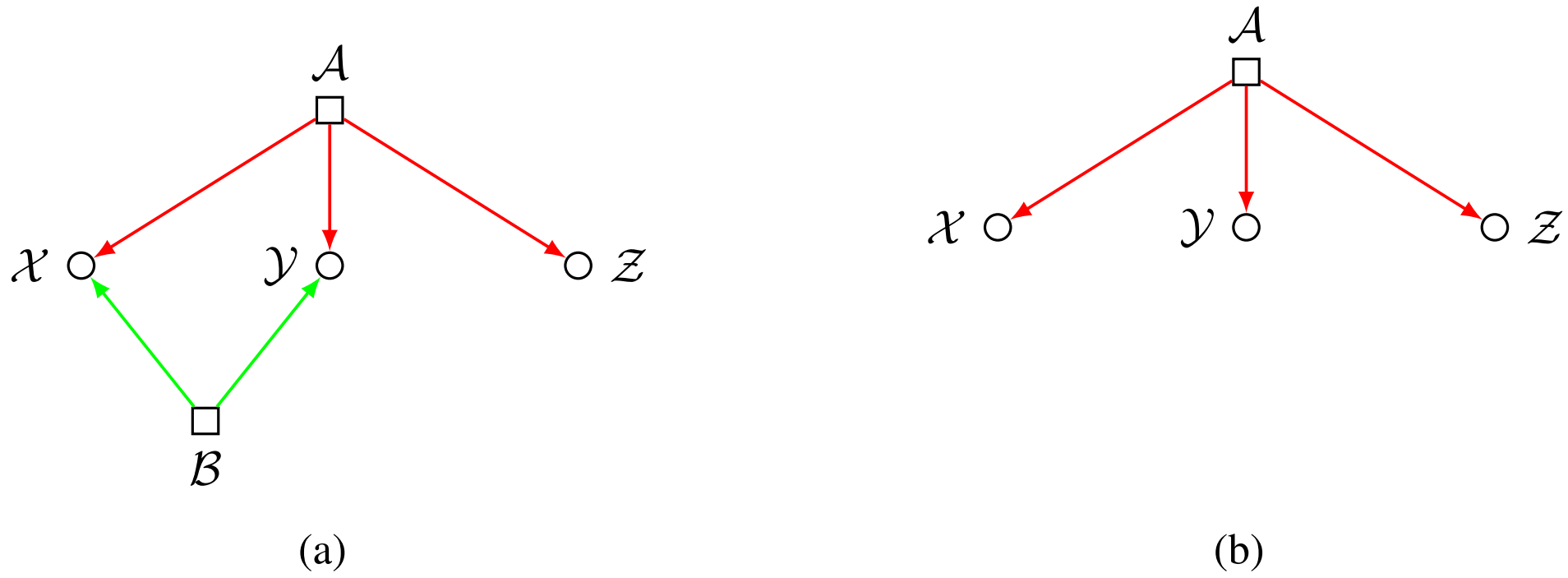

Figure 5: Coalescence of an lvDAG w.r.t. a variable. (a) an lvDAG $\mathfrak{G}$, and (b) its coalescence w.r.t. $\mathcal{B}$, i.e. $\delta_{\mathcal{B}}(\mathfrak{G})$.

Coalescence removes a latent root node under the condition that another root node is connected to all of its children. Similar to exogenization, coalescence is also more easily understood visually. Figure 5 depicts the coalescence of an lvDAG w.r.t. a variable.

It is known that the marginal IPM of an lvDAG does not change under (indeterministic) exogenization or coalescence (Evans, 2016), so by induction an lvDAG can be converted to a pmDAG or mDAG without disrupting its IPM. We inspect these relations for the IPM under the Gaussianity assumption, i.e. for $\mathfrak{q}^{\mathrm{G}(0)}(\mathfrak{G})$. We start with the DPM and use the result to prove the same claim for the IPM.

**Theorem 15** *Let $\mathfrak{G} = \langle \boldsymbol{\mathcal{A}} \equiv \boldsymbol{\mathcal{V}} \dot{\cup} \boldsymbol{\mathcal{L}}, \hookrightarrow \rangle$ be an lvDAG. Then, the Gaussian DPM of $\mathfrak{G}$ is not stable under deterministic exogenization. That is, it does not hold that*

$$\mathfrak{Q}^{\mathrm{G}(0)}_{\boldsymbol{\mathcal{A}} \setminus \{\mathcal{L}\}}(\mathfrak{G}) = \mathfrak{Q}^{\mathrm{G}(0)}(\mathrm{T}_{\mathcal{L}}(\mathfrak{G})),$$

*for any $\mathcal{L} \in \boldsymbol{\mathcal{L}} \cap \mathrm{nroot}(\mathfrak{G})$.*

**Proof** See Proof of Theorem 15 in Appendix C. ■

**Corollary 16** *For an lvDAG $\mathfrak{G} = \langle \boldsymbol{\mathcal{A}} \equiv \boldsymbol{\mathcal{V}} \dot{\cup} \boldsymbol{\mathcal{L}}, \hookrightarrow \rangle$, the Gaussian IPM of $\mathfrak{G}$ is not stable under indeterministic exogenization, i.e.*

$$\mathfrak{q}^{\mathrm{G}(0)}_{\boldsymbol{\mathcal{A}} \setminus \{\mathcal{L}\}}(\mathfrak{G}) = \mathfrak{q}^{\mathrm{G}(0)}(\tau_{\mathcal{L}}(\mathfrak{G})),$$

*for any $\mathcal{L} \in \boldsymbol{\mathcal{L}} \cap \mathrm{nroot}(\mathfrak{G})$ does not hold in general.*

**Proof (sketch)** If we augmented $\mathfrak{G}$, applied deterministic exogenization, then deterministically un-exogenized and un-augmented the resulting lvDAG, we would get the indeterministic exogenization. Each step preserves the IPM. See Proof of Theorem 16 in Appendix C for details. ■

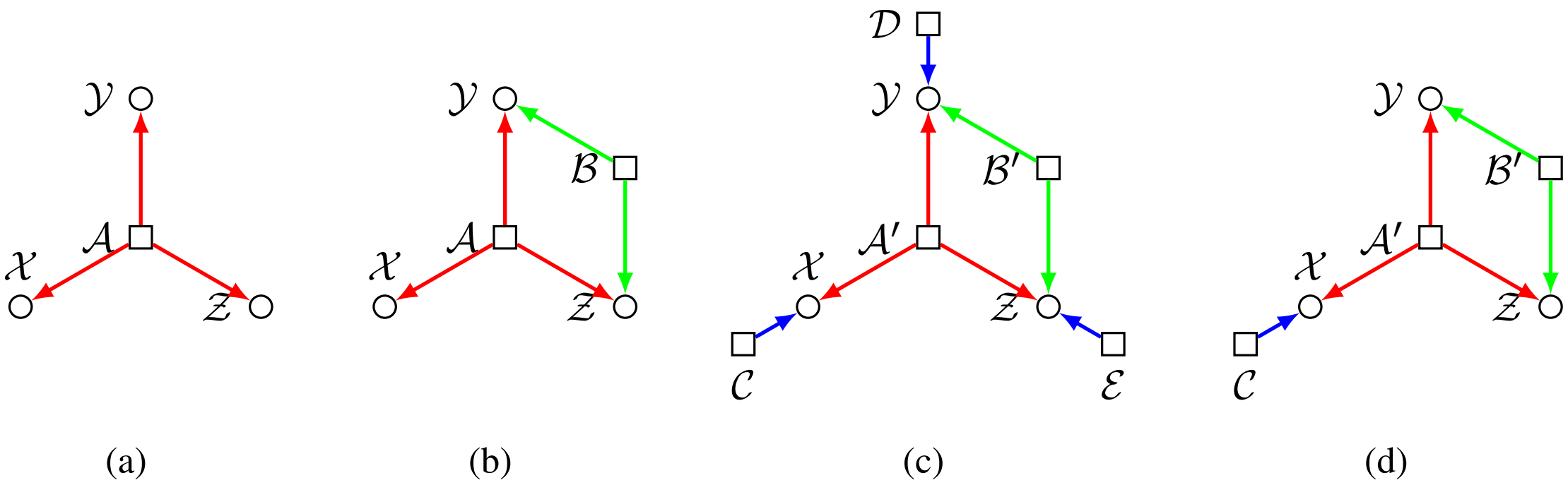

Figure 6: Four pmDAGs. (a) an mDAG $\mathfrak{G}_1 = \langle \boldsymbol{\mathcal{A}}_1 \equiv \boldsymbol{\mathcal{V}} \dot\cup \boldsymbol{\mathcal{L}}_1, \hookrightarrow_1 \rangle$ ($\boldsymbol{\mathcal{V}} \equiv \langle \mathcal{X}, \mathcal{Y}, \mathcal{Z} \rangle$) that is correlation scenario. (b) a pmDAG $\mathfrak{G}_2 = \langle \boldsymbol{\mathcal{A}}_2 \equiv \boldsymbol{\mathcal{V}} \dot\cup \boldsymbol{\mathcal{L}}_2, \hookrightarrow_2 \rangle$ with an additional latent node $\mathcal{B}$ in comparison to $\mathfrak{G}_1$. (c) another pmDAG $\mathfrak{G}_3 = \mathrm{T}_{\mathrm{root}(\mathfrak{G}_2)}(\iota_{\boldsymbol{\mathcal{A}}_2}(\mathfrak{G}))$. (d) a final pmDAG $\mathfrak{G}_4 \subseteq \mathfrak{G}_3$ constructed by removing $\mathcal{D}$ and $\mathcal{E}$ from $\mathfrak{G}_3$.

**Example 1** *Consider the lvDAG* $\mathfrak{G} = \langle \boldsymbol{\mathcal{A}} \equiv \boldsymbol{\mathcal{V}} \dot\cup \boldsymbol{\mathcal{L}}, \hookrightarrow \rangle$ *in Figure 3 (a), its augmentation* $\mathfrak{G}' = \iota_{\boldsymbol{\mathcal{A}}}(\mathfrak{G}) = \langle \boldsymbol{\mathcal{A}}' \equiv \boldsymbol{\mathcal{V}} \dot\cup \boldsymbol{\mathcal{L}}', \hookrightarrow' \rangle$*, and the deterministic exogenization* $\mathfrak{G}'' = \mathrm{T}_{\mathcal{A}}(\mathfrak{G}')$ *of* $\mathfrak{G}'$ *in Figure 3 (c). According to Equation 4,* $\mathfrak{q}^{\mathrm{G}(0)}(\mathfrak{G}) = \mathfrak{Q}^{\mathrm{G}(0)}_{\boldsymbol{\mathcal{A}}}(\mathfrak{G}')$*. Based on Theorem 15, we have* $\mathfrak{Q}^{\mathrm{G}(0)}_{\boldsymbol{\mathcal{A}}' \setminus \{\mathcal{A}\}}(\mathfrak{G}') = \mathfrak{Q}^{\mathrm{G}(0)}(\mathfrak{G}'')$*. By combining these two, we have* $\mathfrak{q}^{\mathrm{G}(0)}_{\boldsymbol{\mathcal{V}}}(\mathfrak{G}) = \mathfrak{Q}^{\mathrm{G}(0)}_{\boldsymbol{\mathcal{V}}}(\mathfrak{G}'')$*. In words, the marginal IPM of* $\mathfrak{G}$ *and the marginal DPM of* $\mathfrak{G}''$ *are the same.*

We showed that the deterministic (indeterministic) exogenization does not change the DPM (IPM) of an lvDAG. We now show that, in general,

$$\mathfrak{q}^{\mathrm{G}(0)}_{\boldsymbol{\mathcal{V}}}(\mathfrak{G}) \neq \mathfrak{q}^{\mathrm{G}(0)}_{\boldsymbol{\mathcal{V}}}(\delta_{\mathcal{L}}(\mathfrak{G})) .$$

That is, we cannot arbitrarily coalesce latent root nodes without disrupting the IPM. We prove this by showing that if $\mathfrak{G} = \langle \boldsymbol{\mathcal{A}} \equiv \boldsymbol{\mathcal{V}} \dot\cup \boldsymbol{\mathcal{L}}, \hookrightarrow \rangle$ is an mDAG with $\mathrm{card}(\boldsymbol{\mathcal{V}}) \geq 2$ that is a "correlation scenario" (Fritz, 2012) (i.e. with the property that for all $\mathcal{V} \in \boldsymbol{\mathcal{V}}$ we have $\mathrm{pa}_{\mathfrak{G}}(\mathcal{V}) \subseteq \mathrm{root}(\mathfrak{G})$) then the pmDAG $\mathfrak{G}' = \mathrm{T}_{\mathrm{root}(\mathfrak{G})}(\iota_{\boldsymbol{\mathcal{A}}}(\mathfrak{G}))$ is "unsaturated" (i.e. $\mathfrak{Q}^{\mathrm{G}(0)}_{\boldsymbol{\mathcal{V}}}(\mathfrak{G}) \subset \mathfrak{P}^{\mathrm{G}(0)}_{\boldsymbol{\mathcal{V}}}$) while the un-coalesced pmDAGs of $\mathfrak{G}'$ can be saturated. This is shown in the following Theorem. We then conclude the desired result as a corollary.

**Theorem 17** *Let* $\mathfrak{G} = \langle \boldsymbol{\mathcal{A}} \equiv \boldsymbol{\mathcal{V}} \dot\cup \boldsymbol{\mathcal{L}}, \hookrightarrow \rangle$ *be an mDAG with* $n = \mathrm{card}(\boldsymbol{\mathcal{V}}) > 2$ *such that for all* $\mathcal{V} \in \boldsymbol{\mathcal{V}}$ *we have* $\mathrm{pa}_{\mathfrak{G}}(\mathcal{V}) \subseteq \boldsymbol{\mathcal{L}}$ *(i.e.* $\mathfrak{G}$ *is a correlation scenario). Also, let* $\mathfrak{G}' = \mathrm{T}_{\mathrm{root}(\mathfrak{G})}(\iota_{\boldsymbol{\mathcal{A}}}(\mathfrak{G}))$*. Then,* $\mathfrak{Q}^{\mathrm{G}(0)}_{\boldsymbol{\mathcal{V}}}(\mathfrak{G}') \subset \mathfrak{P}^{\mathrm{G}(0)}_{\boldsymbol{\mathcal{V}}}$*. That is, some Gaussian distributions (with zero means) are not induced by such a pmDAG.*

**Proof** See Proof of Theorem 17 in Appendix C. ■

It follows that converting a Gaussian pmDAG to an mDAG corresponds to altering its IPM. We demonstrate this result in the following corollary and example.

**Corollary 18** *For an mDAG* $\mathfrak{G} = \langle \boldsymbol{\mathcal{A}} \equiv \boldsymbol{\mathcal{V}} \dot{\cup} \boldsymbol{\mathcal{L}}, \hookrightarrow \rangle$ *with* $n = \operatorname{card}(\boldsymbol{\mathcal{V}}) > 2$ *that is a correlation scenario,* $\mathfrak{q}_{\boldsymbol{\mathcal{V}}}^{\mathrm{G}(0)}(\mathfrak{G}) \subset \mathfrak{P}_{\boldsymbol{\mathcal{V}}}^{\mathrm{G}(0)}$.

**Proof** According to Equation 4 and Theorem 15, Theorem 17 is sufficient for $\mathfrak{q}_{\boldsymbol{\mathcal{V}}}^{\mathrm{G}(0)}(\mathfrak{G}) \subset \mathfrak{P}_{\boldsymbol{\mathcal{V}}}^{\mathrm{G}(0)}$. ■

**Example 2** *Consider* $\mathfrak{G}_1$ *and* $\mathfrak{G}_2$ *in Figures 6(a) and (b). With the help of* $\mathfrak{G}_3$ *and* $\mathfrak{G}_4$ *we want to see whether the IPM* $\mathfrak{q}_{\boldsymbol{\mathcal{V}}}^{\mathrm{G}(0)}(\mathfrak{G}_2)$ *remains unchanged under the coalescence* $\mathfrak{G}_1 = \delta_{\mathcal{B}}(\mathfrak{G}_2)$*, i.e. whether* $\mathfrak{q}_{\boldsymbol{\mathcal{V}}}^{\mathrm{G}(0)}(\mathfrak{G}_1) = \mathfrak{q}_{\boldsymbol{\mathcal{V}}}^{\mathrm{G}(0)}(\mathfrak{G}_2)$.

*(a) According to Corollary 18, the lvDAG* $\mathfrak{G}_1$ *is unsaturated, i.e.* $\mathfrak{q}_{\boldsymbol{\mathcal{V}}}^{\mathrm{G}(0)}(\mathfrak{G}_1) \subset \mathfrak{P}_{\boldsymbol{\mathcal{V}}}^{\mathrm{G}(0)}$.

*(b) We prove that* $\mathfrak{Q}_{\boldsymbol{\mathcal{V}}}^{\mathrm{G}(0)}(\mathfrak{G}_4) = \mathfrak{P}_{\boldsymbol{\mathcal{V}}}^{\mathrm{G}(0)}$*. To show this, let* $\mathfrak{C}_3$ *be the cone of* 3*-dimensional semi-definite matrices. Let* $\bar{\Sigma} = \operatorname{diag}\left(\begin{bmatrix} \bar{\sigma}^2_{\mathcal{A}'} & \bar{\sigma}^2_{\mathcal{B}'} & \bar{\sigma}^2_{\mathcal{C}} \end{bmatrix}\right)$ *and assume that* $\boldsymbol{\mathcal{V}} = \Phi_{\boldsymbol{\mathcal{V}}}(\boldsymbol{\mathcal{L}}) = \bar{W}^\top \boldsymbol{\mathcal{L}}$*, i.e.*

$$\begin{bmatrix} \mathcal{X} \\ \mathcal{Y} \\ \mathcal{Z} \end{bmatrix} = \begin{bmatrix} \bar{w}_{\mathcal{A}'\mathcal{X}} & 0 & \bar{w}_{\mathcal{C}\mathcal{X}} \\ \bar{w}_{\mathcal{A}'\mathcal{Y}} & \bar{w}_{\mathcal{B}'\mathcal{X}} & 0 \\ \bar{w}_{\mathcal{A}'\mathcal{Z}} & \bar{w}_{\mathcal{B}'\mathcal{Z}} & 0 \end{bmatrix} \begin{bmatrix} \mathcal{A} \\ \mathcal{B} \\ \mathcal{C} \end{bmatrix}.$$

*We show that* $\mathfrak{G}_4$ *induces all Gaussian distributions. Fix* $\Sigma \in \mathfrak{C}_3$ *as the covariance of* $\boldsymbol{\mathcal{V}}$ *and let* $L(\Sigma) = \begin{bmatrix} a & & \\ b & d & \\ c & e & f \end{bmatrix}$ *be its Cholesky decomposition* ($\Sigma = L(\Sigma)L(\Sigma)^\top$)*. It is enough to show that the system of equations* $\bar{W}^\top \bar{\Sigma} \bar{W} = L(\Sigma)L(\Sigma)^\top$ *is solvable for* $W$*: if* $\Sigma$ *is positive definite, let* $\bar{w}_{\mathcal{C}\mathcal{X}} = \pm adf\sqrt{\frac{\bar{\sigma}^2_{\mathcal{A}'}}{(be-cd)^2+(bf)^2+(df)^2}}$*; then substitute other* $\bar{w}_{..}$ *variables accordingly. Repeat the process for covariance matrices that are not positive definite.*

*(c)* $\mathfrak{G}_3$ *has two extra latent variables* $\mathcal{D}$ *and* $\mathcal{E}$ *in comparison to* $\mathfrak{G}_4$*, and therefore,* $\mathfrak{Q}_{\boldsymbol{\mathcal{V}}}^{\mathrm{G}(0)}(\mathfrak{G}_4) \subseteq \mathfrak{Q}_{\boldsymbol{\mathcal{V}}}^{\mathrm{G}(0)}(\mathfrak{G}_3) = \mathfrak{P}_{\boldsymbol{\mathcal{V}}}^{\mathrm{G}(0)}$ *(Lemma 50 in Appendix C). Moreover,* $\mathfrak{G}_3 = \mathrm{T}_{\mathrm{root}(\mathfrak{G}_2)}(\iota_{\boldsymbol{\mathcal{A}}_2}(\mathfrak{G}_2))$*. Therefore, we have* $\mathfrak{q}_{\boldsymbol{\mathcal{V}}}^{\mathrm{G}(0)}(\mathfrak{G}_2) = \mathfrak{Q}_{\boldsymbol{\mathcal{V}}}^{\mathrm{G}(0)}(\mathfrak{G}_3) = \mathfrak{P}_{\boldsymbol{\mathcal{V}}}^{\mathrm{G}(0)}$.

*We have shown* $\mathfrak{q}_{\boldsymbol{\mathcal{V}}}^{\mathrm{G}(0)}(\mathfrak{G}_1) \subset \mathfrak{q}_{\boldsymbol{\mathcal{V}}}^{\mathrm{G}(0)}(\mathfrak{G}_2)$*. Yet,* $\mathfrak{G}_1 = \delta_{\mathcal{B}}(\mathfrak{G}_2)$*. Therefore, the margin has not remained stable under the coalescence.*

The conclusion of the above discussion is that we cannot arbitrarily coalesce root nodes or exogenize non-root nodes of an lvDAG to construct an mDAG without disrupting the marginal Gaussian IPM. While exogenization relaxes constraints on the IPM/DPM, coalescence imposes further constraints on the IPM/DPM. Formally, for an lvDAG $\mathfrak{G} = \langle \boldsymbol{\mathcal{A}} \equiv \boldsymbol{\mathcal{V}} \dot{\cup} \boldsymbol{\mathcal{L}}, \hookrightarrow \rangle$, we have $\mathfrak{q}_{\boldsymbol{\mathcal{V}}}^{\mathrm{G}(0)}(\mathfrak{G}) \neq \mathfrak{q}_{\boldsymbol{\mathcal{V}}}^{\mathrm{G}(0)}(\delta_{\mathcal{Q}}(\mathfrak{G}))$ and $\mathfrak{q}_{\boldsymbol{\mathcal{V}}}^{\mathrm{G}(0)}(\mathfrak{G}) \neq \mathfrak{q}_{\boldsymbol{\mathcal{V}}}^{\mathrm{G}(0)}(\tau_{\mathcal{L}}(\mathfrak{G}))$. Similarly, the marginal Gaussian DPMs of lvDAGs are also unstable under coalescence or exogenization. To remove the restriction imposed by coalescence we use a class of lvDAGs that is more relaxed than mDAGs. A natural choice will be pre-marginalized DAGs (pmDAGs). Note that, we assume that the current pmDAG is not the result of an exogenization process that has limited the IPM/DPM—this is aligned with the current literature or linear models.

## 4 Problem Definition

The main objective of this article is to propose an algorithm that finds a structural system of a pmDAG that is in some sense optimal given a sequence of observations $V$. An *observation* is the measurement of the visible variables—that are $\boldsymbol{\mathcal{V}}$ in $\mathfrak{G} = \langle \boldsymbol{\mathcal{A}} \equiv \boldsymbol{\mathcal{V}} \dot\cup \boldsymbol{\mathcal{L}}, \hookrightarrow \rangle$. The complete DSM is generally too large for the von Neumann architecture to be able to search in. Therefore, we naturally consider a sub-model and assign a set of parameters to each structural system in that sub-model, then we search within the space of all possible sets of parameters until we find an optimal one. This procedure is called the *parameter-optimization* of a pmDAG and corresponds to finding an optimal structural system of $\mathfrak{G}$. We mark this optimal structural system of $\mathfrak{G}$ with a superscript asterisk, i.e. we denote it using $\langle \mathbb{P}^*, \Phi^* \rangle$ (or, equivalently, $\langle \mathbb{P}, \Phi \rangle^*$).

The semantics of optimality can be defined in many ways and be varied from scenario to scenario. Among the possible ways one can define optimality, one of the frequent ways of considering a deterministic structural system to be optimal is when its parameters are set according to the maximum likelihood estimation (MLE). MLE is the *de facto* standard for learning probabilistic models from an iid sequence of observations (Murphy, 2023). In this section, we formally define the MLE. Our definition of the MLE allows for the inclusion of arbitrary assumptions as constraints. In this sense, finding the MLE is a constrained optimization problem. Based on this definition, we will formally define the problem which we are interested in solving.

**Definition 19 (maximum likelihood estimation)** *Let $\mathfrak{G} = \langle \boldsymbol{\mathcal{A}} \equiv \boldsymbol{\mathcal{V}} \dot\cup \boldsymbol{\mathcal{L}}, \hookrightarrow \rangle$ be a pmDAG and $V$ a sequence of (iid) observations $(\boldsymbol{v})_{\boldsymbol{v} \in V}$ of $\boldsymbol{\mathcal{V}}$. Then, we have:*

*(i) The* likelihood function *of $\mathfrak{G}$ w.r.t. $V$ is any function $\ell_{\mathfrak{G},V} : \mathfrak{D}(\mathfrak{G}) \to [0, \infty)$ that satisfies*

$$\ell_{\mathfrak{G},V}(\langle \mathbb{P}, \Phi \rangle) = \prod_{\boldsymbol{v} \in V} \frac{\mathrm{d}}{\mathrm{d}\boldsymbol{v}} \mathbb{P}_{\boldsymbol{\mathcal{V}}}\left( \left\{ \boldsymbol{v}' \in \Omega_{\boldsymbol{\mathcal{V}}} \,\middle|\, \boldsymbol{v}' \leq \boldsymbol{v} \right\} \right) \qquad \Longleftarrow \qquad \mathbb{P}_{\boldsymbol{\mathcal{A}}} = \Pi(\langle \mathbb{P}, \Phi \rangle),$$

*for all deterministic structural systems $\langle \mathbb{P}, \Phi \rangle \in \mathfrak{D}(\mathfrak{G})$ of $\mathfrak{G}$. In short notation, we may write:*

$$\ell_{\mathfrak{G},V}(\langle \mathbb{P}, \Phi \rangle) \coloneqq \prod_{\boldsymbol{v} \in V} \frac{\mathrm{d}}{\mathrm{d}\boldsymbol{v}} \Pi(\langle \mathbb{P}, \Phi \rangle)_{\boldsymbol{\mathcal{V}}}(\boldsymbol{\mathcal{V}} \leq \boldsymbol{v}).$$

*(ii) Let $\mathfrak{I} \subseteq \mathfrak{D}(\mathfrak{G})$ be an arbitrary collection subsetting $\mathfrak{D}(\mathfrak{G})$. The* maximum likelihood estimation *(MLE) of $\mathfrak{G}$ constrained on $\mathfrak{I}$ and w.r.t. $V$ is the operation defined as:*

$$\mathrm{MLE}_{\mathfrak{G},V}(\mathfrak{I}) \coloneqq \underset{\langle \mathbb{P}, \Phi \rangle \in \mathfrak{I}}{\arg\max} \left\{ \ell_{\mathfrak{G},V}(\langle \mathbb{P}, \Phi \rangle) \right\}.$$

*In the most relaxed case, $\mathfrak{I} = \mathfrak{D}(\mathfrak{G})$, and the MLE is not constrained by any other assumption than those imposed by the pmDAG.*[6]

The MLE of $\mathfrak{G}$—as in many (constrained) optimization problems—is not necessarily unique. (We discuss the consequence of this non-uniqueness in §7.) A typical optimization algorithm tries to find a single indeterministic structural system in $\mathrm{MLE}_{\mathfrak{G},V}(\mathfrak{I})$. This is exactly what our algorithm does in §6.

6. We assume that the maximum likelihood estimation always exists.

We let the MLE be defined on arbitrary sets $\mathfrak{I}$. This generalization is helpful when there are assumptions about the nature of the probability space. For example, if one wants to find $\langle\mathbb{P},\Phi\rangle^*$ such that the induced probability $\mathbb{P}_{\boldsymbol{\mathcal{A}}} = \Pi\big(\langle\mathbb{P},\Phi\rangle^*\big)$ is Gaussian with zero means, they let $\mathfrak{I} = \mathfrak{D}^{\mathrm{G}(0)}(\mathfrak{G})$.

In view of the above discussion, we formally define the problem that we want to solve in this article. Instead of proceeding with $\mathfrak{D}^{\mathrm{G}(0)}(\mathfrak{G})$, we restrict attention to a subset of $\mathfrak{D}^{\mathrm{G}(0)}(\mathfrak{G})$ defined as follows:

$$\mathfrak{D}^{\mathrm{G}(0,1)}(\mathfrak{G}) := \Big\{\langle\mathbb{P},\Phi\rangle \in \mathfrak{D}^{\mathrm{G}(0)}(\mathfrak{G}) \,\Big|\, \forall\ \mathcal{A} \in \mathrm{root}(\mathfrak{G}) \,:\, \bar{\sigma}^2_{\mathcal{A}} = 1\Big\};$$

the subset where each root node is standard Gaussian. This modification is without loss of generality as the stability of the DPM under deterministic exogenization suggests. To see the details, refer to Appendix B.2. Let $\mathfrak{G} = \langle\boldsymbol{\mathcal{A}} \equiv \boldsymbol{\mathcal{V}} \dot\cup \boldsymbol{\mathcal{L}}, \hookrightarrow\rangle$ be a pmDAG. Given a sequence of observations $V$, we would like to find a $\langle\mathbb{P},\Phi\rangle^* \in \mathfrak{D}^{\mathrm{G}(0,1)}(\mathfrak{G})$ such that:

$$\langle\mathbb{P},\Phi\rangle^* \in \mathrm{MLE}_{\mathfrak{G},V}\Big(\mathfrak{D}^{\mathrm{G}(0,1)}(\mathfrak{G})\Big). \tag{5}$$

That is, among all linear Gaussian structural systems with standard Gaussian root nodes, we seek one of those that minimize the likelihood function of $\mathfrak{G}$ w.r.t. the set of observations $V$.

In the next subsection, we explain the relationship between the MLE and the Kullback-Leibler (KL) divergence. In this article, we will ultimately use the KL divergence to compute the MLE.

### 4.1 The MLE and KL Divergence

It is widely known that asymptotically the MLE amounts to minimizing the Kullback-Leibler (KL) divergence of the observed probability from the induced one (Wasserman, 2013, Ch. 9.5). Formally,

$$\mathrm{MLE}_{\mathfrak{G},V}(\mathfrak{D}(\mathfrak{G})) - \underset{\langle\mathbb{P},\Phi\rangle\in\mathfrak{D}(\mathfrak{G})}{\arg\min}\ \Big\{\mathrm{KL}\Big(\mathbb{P}^V_{\boldsymbol{\mathcal{V}}} \,\Big\|\, \Pi(\langle\mathbb{P},\Phi\rangle)_{\boldsymbol{\mathcal{V}}}\Big)\Big\} \xrightarrow{\text{p.}} 0, \qquad \text{as } \mathrm{card}(V) \longrightarrow \infty.$$

where $\mathbb{P}^V_{\boldsymbol{\mathcal{V}}}$ is the probability distribution over the set of observed variables $\boldsymbol{\mathcal{V}}$ given the sequence of observations $V$. We prove a slightly different proposition for the Gaussian case.

**Theorem 20** *Let $\mathfrak{G} = \langle\boldsymbol{\mathcal{A}} \equiv \boldsymbol{\mathcal{V}} \dot\cup \boldsymbol{\mathcal{L}}, \hookrightarrow\rangle$ be a pmDAG and $V$ a sequence of observations $(v)_{v\in V}$ of $\boldsymbol{\mathcal{V}}$. Let $\mathbb{P}^V_{\boldsymbol{\mathcal{V}}}$ be a measure of the estimated Gaussian probability* $\mathrm{Normal}\big(\hat{\mu}(V),\hat{\Sigma}(V)\big)$*, where $\hat{\mu}(V) :=$* $\mathrm{mean}(V)$ *and* $\hat{\Sigma}(V) := \big({(\mathrm{card}(V)-1)}/{\mathrm{card}(V)}\big)\,\mathrm{cov}(V)$ *are the sample mean and the (biased) sample covariance, respectively. Then, we have:*

$$\mathrm{MLE}_{\mathfrak{G},V}(*) = \underset{\langle\mathbb{P},\Phi\rangle\in *}{\arg\min}\Big\{\mathrm{KL}\Big(\Pi(\langle\mathbb{P},\Phi\rangle)_{\boldsymbol{\mathcal{V}}} \,\Big\|\, \mathbb{P}^V_{\boldsymbol{\mathcal{V}}}\Big)\Big\}, \tag{6}$$

*where* $* \equiv \mathfrak{D}^{\mathrm{G}(0,1)}(\mathfrak{G})$.

**Proof** See Proof of Theorem 20 in Appendix C. ■

Theorem 20 states that if we want to find the deterministic structural system with Gaussianity and linearity, we can use the KL divergence instead of directly computing the MLE. In the remainder of this article, especially in §6, we restrict attention to finding a member of the left-hand side of Equation 6.

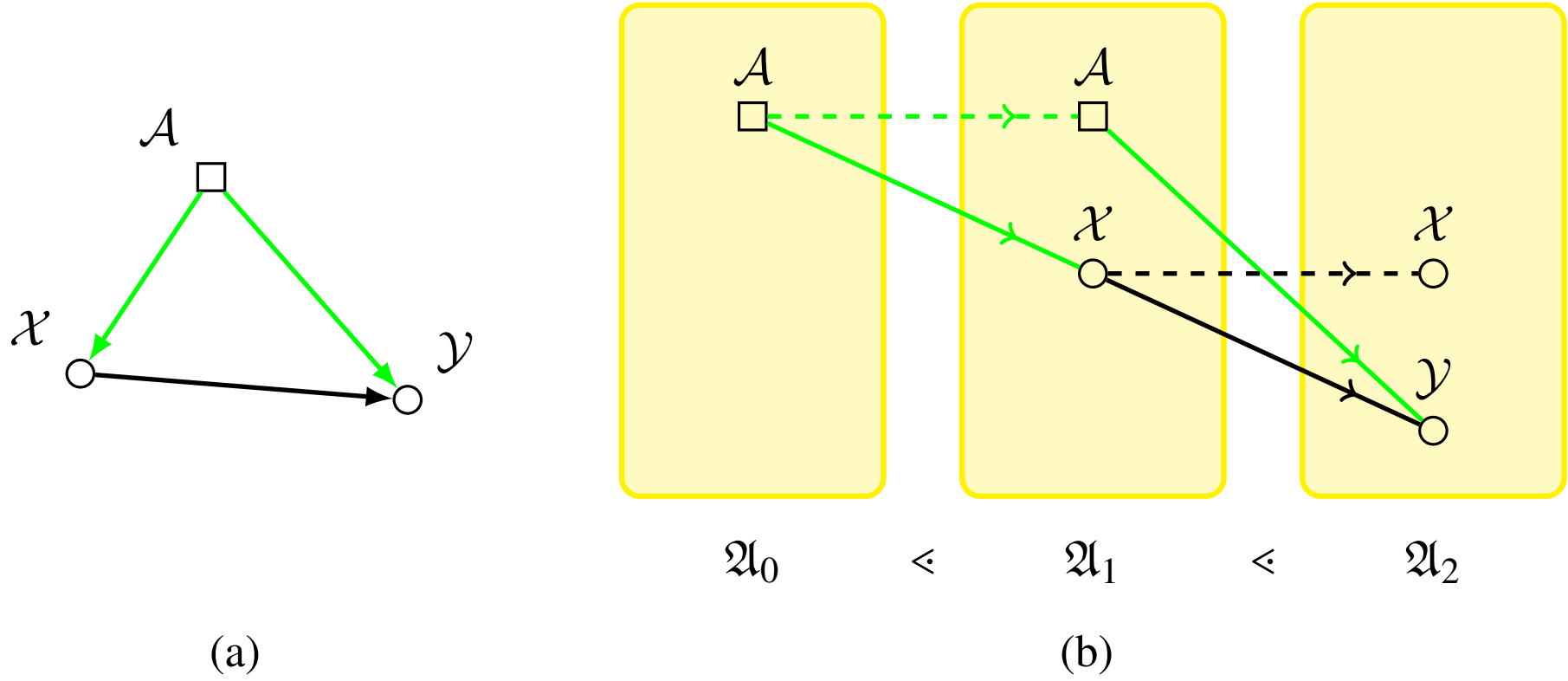

Figure 7: A simple pmDAG and its synchronization. (a) a pmDAG $\mathfrak{G}$ (a bow), and (b) its synchronization $\Xi(\mathfrak{G})$.

**Remark 21** *A hidden premise in the Proof of Theorem 20, which might not be readily obvious, is that* $\mathbb{P}^{V}_{\boldsymbol{\mathcal{V}}} \in \Pi\big(\mathfrak{D}^{\mathrm{G}(0)}(\mathfrak{G})\big)$. *It is in this case that we can assume the minimum KL divergence of the search space is symmetric; see Lemma 51 in Appendix C.*

In the next section (§5), we introduce the "synchronization" of a pmDAG, which makes the parameter-optimization straightforward. As we will see, although the MLE is central to our work, parameter-optimization is not limited to the MLE, and our technique is as well capable of performing other optimization methods based on different optimality semantics.

## 5 Synchronization of pmDAGs

Under certain conditions, the set of indeterministic structural systems of a pmDAG is amenable to gradient-based optimization. For parameterizing the structural system of a pmDAG, we introduce a transformation of these DAGs that make the optimization straightforward. This transformation converts a pmDAG into a "medium" object that benefits from favorable characteristics of neural networks while preserving crucial properties of the pmDAG. These medium objects are called *synchronizations*.

In what follows, we work with finite chains (totally ordered sets) $\langle \mathfrak{O}, \leq \rangle$. A chain $\langle \mathfrak{O}, \leq \rangle$ is $\leq$-order-isomorphic to $\{0, \cdots, \mathrm{card}(\mathfrak{O}) - 1\}$ and we simply refer to its elements using a subscript $\mathfrak{O}_l$ ($l \in \{0, \cdots, \mathrm{card}(\mathfrak{O}) - 1\}$).

**Definition 22 (synchronization)** *A* synchronization *of a pmDAG* $\mathfrak{G} = \langle \boldsymbol{\mathcal{A}} \equiv \boldsymbol{\mathcal{V}} \dot{\cup} \boldsymbol{\mathcal{L}}, \hookrightarrow \rangle$, *denoted by* $\Xi(\mathfrak{G})$, *is a triple* $\langle \mathfrak{G}, \mathfrak{A}, \leq \rangle$ *composed of* $\mathfrak{G}$ *itself and a chain* $\langle \mathfrak{A}, \leq \rangle$ *with* $\mathfrak{A} \subseteq \wp \boldsymbol{\mathcal{A}}$, *constructed as follows:*

*(1) let* $\boldsymbol{\mathcal{P}} \equiv \varnothing$, $\boldsymbol{\mathcal{C}} \equiv \boldsymbol{\mathcal{A}}$, $\boldsymbol{\mathcal{S}} = \mathrm{root}(\mathfrak{G})$ *and* $l = 0$.

*(2) do the following until* $\boldsymbol{\mathcal{C}} \equiv \varnothing$.

- *(a) construct the DAG* $\mathfrak{H} = \langle \boldsymbol{\mathcal{C}}, \hookrightarrow \upharpoonright_{\boldsymbol{\mathcal{C}}} \rangle$,
- *(b) let* $\boldsymbol{\mathcal{C}} \equiv \boldsymbol{\mathcal{C}} \smallsetminus \boldsymbol{\mathcal{S}}$,
- *(c) let* $\mathfrak{A}_l \equiv \boldsymbol{\mathcal{S}} \cup (\boldsymbol{\mathcal{P}} \cap \boldsymbol{\mathcal{V}}) \cup (\boldsymbol{\mathcal{P}} \cap \boldsymbol{\mathcal{L}} \cap \mathrm{pa}_{\mathfrak{G}}(\boldsymbol{\mathcal{C}}))$,
- *(d) let* $\boldsymbol{\mathcal{P}} \equiv \boldsymbol{\mathcal{P}} \cup \boldsymbol{\mathcal{S}}$,

*(e) choose* $\varnothing \subset \boldsymbol{\mathcal{S}} \subseteq \mathrm{root}(\mathfrak{H})$ *and let* $l = l + 1$.

*We define the* depth *of* $\Xi(\mathfrak{G})$ *as* $\|\Xi(\mathfrak{G})\| := \mathrm{card}(\mathfrak{A})$*. Moreover, we call each* $\mathfrak{A}_l$ *(*$0 \le l < \|\Xi(\mathfrak{G})\|$*) in above a* layer *of the* $\Xi(\mathfrak{G})$.

Synchronization assigns the root nodes to $\mathfrak{A}_0$. It then gradually dissembles $\mathfrak{G}$ by iteratively removing (arbitrary) subsets of the root variables from its remainder. These emergent root variables along with some "relevant" variables are iteratively added to the sets that form the elements of the chain $\langle \mathfrak{A}, \le \rangle$. The relevant variables are either the visible variables that have previously been visited (i.e. $\boldsymbol{\mathcal{P}} \cap \boldsymbol{\mathcal{V}}$ in (c)) or the latent variables that (have previously been visited and) still have a child in the remnant of $\mathfrak{G}$ (i.e. $\boldsymbol{\mathcal{P}} \cap \boldsymbol{\mathcal{L}} \cap \mathrm{pa}_{\mathfrak{G}}(\boldsymbol{\mathcal{C}})$ in (c)). Notice that, depending on the choices of $\boldsymbol{\mathcal{S}}$, this procedure generates different $\Xi(\mathfrak{G})$'s. We will further elaborate on the synchronization in the following, but if Figure 7 (a) depicts $\mathfrak{G}$, the yellow boxes in 7 (b) are the constructed layers $\mathfrak{A}_l$ in $\Xi(\mathfrak{G}) = \langle \mathfrak{G}, \mathfrak{A}, \le \rangle$ $(0 \le l < \|\Xi(\mathfrak{G})\|)$.

**Remark 23** *The synchronization of a pmDAG produces non-unique structures that form an isomorphism class. When we write* $\Xi(\mathfrak{G})$*, we refer to a generic member of this class. Our theorems are general enough to apply to any of these members. Moreover, for each of these isomorphism classes there is only one pmDAG that generates every member of the class.*

*In our implementation in §6, however, we construct a synchronization with the minimum number of layers by letting* $\boldsymbol{\mathcal{S}} \equiv \mathrm{root}(\mathfrak{H})$ *in Definition 22-(2) (e). This construction might not always be the most efficient construction for the task of parameter optimization.*

Let $\Xi(\mathfrak{G}) = \langle \mathfrak{G}, \mathfrak{A}, \le \rangle$ be the synchronization of $\mathfrak{G} = \langle \boldsymbol{\mathcal{A}} \equiv \boldsymbol{\mathcal{V}} \dot\cup \boldsymbol{\mathcal{L}}, \hookrightarrow \rangle$. Similar to the parents function in lvDAGs, a synchronization of a pmDAG comes with "layer-wise parents" functions. We define this function via the *first appearance* function $\mathrm{app}_{\le} : \boldsymbol{\mathcal{A}} \to \{0, \cdots, \|\Xi(\mathfrak{G})\| - 1\}$, which is itself defined as $\mathrm{app}_{\le}(\mathcal{A}) := \min\{0 \le l < \|\Xi(\mathfrak{G})\| \,|\, \mathcal{A} \in \mathfrak{A}_l\}$. $\mathrm{app}_{\le}(\mathcal{A})$ returns the index of the lowest layer on which $\mathcal{A}$ appears. The *layer-wise parents* functions are denoted by $\mathrm{pa}_{\Xi(\mathfrak{G})|l} : \mathfrak{A}_l \to \wp\mathfrak{A}_{l-1}$, $0 \le l < \|\Xi(\mathfrak{G})\|$ and defined as:

$$\mathrm{pa}_{\Xi(\mathfrak{G})|l}(\mathcal{A}) := \begin{cases} \mathrm{pa}_{\mathfrak{G}}(\mathcal{A}) & \mathrm{app}_{\le}(\mathcal{A}) = l, \\ \{\mathcal{A}\} & \text{otherwise.} \end{cases}$$

In words, the layer-wise parents of $\mathcal{A}$ are equivalent to its parents in $\mathfrak{G}$ when it first appears and are equal to $\{\mathcal{A}\}$ after that. We extend the definition of layer-wise parents by defining $\mathrm{pa}_{\Xi(\mathfrak{G})|l} : \wp\mathfrak{A}_l \to \wp\mathfrak{A}_{l-1}$, $0 \le l < \|\Xi(\mathfrak{G})\|$ as $\mathrm{pa}_{\Xi(\mathfrak{G})|l}(\boldsymbol{\mathcal{U}}) := \bigcup_{\mathcal{U} \in \boldsymbol{\mathcal{U}}} \mathrm{pa}_{\Xi(\mathfrak{G})|l}(\mathcal{U})$. These functions will be used as an essential tool for working for synchronizations. Similar to an lvDAG, its synchronization also has a visual (semi-graphical) representation.

**Definition 24 (visualization of synchronization)** *We draw a synchronization* $\Xi(\mathfrak{G}) = \langle \mathfrak{G}, \mathfrak{A}, \le \rangle$ *by following the instruction provided below:*

*(a) Draw each set* $\mathfrak{A}_l$ *(*$0 \le l < \|\Xi(\mathfrak{G})\|$*) from left to right.*

*(b) For each* $\mathfrak{A}_l$ *(*$0 < l < \|\Xi(\mathfrak{G})\|$*),*

*for each* $\mathcal{A} \in \mathfrak{A}_l$ *and* $\mathcal{B} \in \mathrm{pa}_{\Xi(\mathfrak{G})|l}(\mathcal{A})$*, draw an arrow from* $\mathcal{B} \in \mathfrak{A}_{l-1}$ *to* $\mathcal{A}$*, using a solid line (*$\longrightarrow$*) if* $\mathrm{app}_{\le}(\mathcal{A}) = l$*, or a dashed line (*$\dashrightarrow$*) otherwise.*

To see an example of this graphical representation, please refer to Figure 7, where the pmDAG in 7 (a) has been synchronized and it synchronization has been depicted in 7 (b). This procedure is well-defined, in the sense that for any $\mathcal{A} \in \mathfrak{A}_l$ we have $\mathrm{pa}_{\Xi(\mathfrak{G})|l}(\mathcal{A}) \subseteq \mathfrak{A}_{l-1}$, and therefore, we can always pick $\mathcal{B} \in \mathrm{pa}_{\Xi(\mathfrak{G})|l}(\mathcal{A})$ from $\mathfrak{A}_{l-1}$. We prove this in Lemma 52 in Appendix C.

At this point, synchronizations obviously serve as feed-forward neural networks and lvDAGs. There are indeed two observations that are worth mentioning:

(i) As the visual representation (Figure 7 (b)) suggests, the synchronization of a pmDAG can be thought of as an lvDAG. It contains latent and visible variables (that repeat in a multiset) and edges (solid or dashed) that connect them without making a cycle. These edges can be interpreted as equivalent to the relation $\hookrightarrow$ in the lvDAG.

(ii) The synchronization of a pmDAG can also be considered a feed-forward neural network. It is composed of layers $\mathfrak{A}_l$ ($0 \leq l < \|\Xi(\mathfrak{G})\|$) and some "weights" between consecutive layers that are composed of the edges. These "weights" implicate dense layers in the neural networks.

Notice that we do not claim that every feed-forward neural network can be representative of a pmDAG, but rather, we can transform every pmDAG to a special type of feed-forward neural network called synchronization. In essence, this duality between the two structures in one-directional.

We transformed pmDAGs to synchronizations. We also need a map from the pmDAGs' ISM to that of the synchronization (as the ISM is more general than the DSM). The following definition relates the indeterministic structural system of a pmDAG to that of its synchronization.

**Definition 25 (indeterministic structural system of synchronization)** *Let* $\mathfrak{G} = \langle \boldsymbol{\mathcal{A}} \equiv \boldsymbol{\mathcal{V}} \dot{\cup} \boldsymbol{\mathcal{L}}, \hookrightarrow \rangle$ *be an lvDAG,* $\{\kappa_{\mathcal{A}}\}_{\mathcal{A} \in \boldsymbol{\mathcal{A}}}$ *be an indeterministic structural system of* $\mathfrak{G}$*, and* $\Xi(\mathfrak{G}) = \langle \mathfrak{G}, \mathfrak{A}, \leq \rangle$ *be* $\mathfrak{G}$*'s synchronization. The* indeterministic structural system of $\Xi(\mathfrak{G})$ corresponding to $\{\kappa_{\mathcal{A}}\}_{\mathcal{A} \in \boldsymbol{\mathcal{A}}}$ *is the set of Markov kernels* $\left\{\kappa_{\mathcal{A}|l} : \sigma\Omega_{\mathcal{A}} \times \Omega_{\mathrm{pa}_{\Xi(\mathfrak{G})|l}(\mathcal{A})} \to [0,1]\right\}_{\mathcal{A} \in \mathfrak{A}_l, 0 \leq l < \|\Xi(\mathfrak{G})\|}$ *each defined as:*

$$\kappa_{\mathcal{A}|l}(X|\boldsymbol{y}) := \begin{cases} \kappa_{\mathcal{A}}(X|\boldsymbol{y}) & \mathrm{app}_{\leq}(\mathcal{A}) = l, \\ \mathbb{1}_X(\boldsymbol{y}) & \textit{otherwise}, \end{cases} \tag{7}$$

*for all* $X \in \sigma\Omega_{\mathcal{A}}$*,* $\boldsymbol{y} \in \Omega_{\mathrm{pa}_{\Xi(\mathfrak{G})|l}(\mathcal{A})}$*, for* $\mathcal{A} \in \mathfrak{A}_l$ *and* $0 \leq l < \|\Xi(\mathfrak{G})\|$*. We use* $\left\{\kappa_{\mathcal{A}|l}\right\}_{\mathcal{A} \in \mathfrak{A}_l, 0 \leq l < \|\Xi(\mathfrak{G})\|}$ *as the shorthand notation.*

*We denote the mapping that takes* $\{\kappa_{\mathcal{A}}\}_{\mathcal{A} \in \boldsymbol{\mathcal{A}}} \in \mathfrak{d}(\mathfrak{G})$ *and returns the corresponding indeterministic structural system* $\left\{\kappa_{\mathcal{A}|l}\right\}_{\mathcal{A} \in \mathfrak{A}_l, 0 \leq l < \|\Xi(\mathfrak{G})\|}$ *using* $\Xi_{\mathfrak{G}} : \mathfrak{d}(\mathfrak{G}) \to \Xi(\mathfrak{d}(\mathfrak{G}))$*, where we denote the range of this mapping using* $\Xi(\mathfrak{d}(\mathfrak{G}))$*.*[7]

For the synchronization $\Xi(\mathfrak{G}) = \langle \mathfrak{G}, \mathfrak{A}, \leq \rangle$ of the pmDAG $\mathfrak{G} = \langle \boldsymbol{\mathcal{A}} \equiv \boldsymbol{\mathcal{V}} \dot{\cup} \boldsymbol{\mathcal{L}}, \hookrightarrow \rangle$ we may write the inverse of the indeterministic structural system $\Xi_{\mathfrak{G}} : \mathfrak{d}(\mathfrak{G}) \to \Xi(\mathfrak{d}(\mathfrak{G}))$ defined in Definition 25 using the following equation:

$$\kappa_{\mathcal{A}}(X|\boldsymbol{y}) := \kappa_{\mathcal{A}|\mathrm{app}_{\leq}(\mathcal{A})}(X|\boldsymbol{y}), \tag{8}$$

for all $X \in \sigma\Omega_{\mathcal{A}}$, $\boldsymbol{y} \in \Omega_{\mathrm{pa}_{\mathfrak{G}}(\mathcal{A})}$, for $\mathcal{A} \in \boldsymbol{\mathcal{A}}$.

7. In this sense, we can roughly see the synchronization procedure $\Xi$ as a functor that takes the category of Bayesian networks into the category of synchronized Bayesian networks.

Notice that in the indeterministic structural system of $\Xi(\mathfrak{G})$ corresponding to that of $\mathfrak{G}$ there is one and only one equation for each variable $\mathcal{A}$ that is equivalent to a Markov kernel in the indeterministic structural system of $\mathfrak{G}$. For example, $\kappa_{\mathcal{Y}}$ in Figure 7 (a) and $\kappa_{\mathcal{Y}|2}$ in Figure 7 (b) are the same Markov kernels. This implies that we might be able to parameterize $\mathfrak{G}$ by finding an optimal $\{\kappa_{\mathcal{A}|l}\}_{\mathcal{A}\in\mathfrak{A}_l,\, 0\leq l<\|\Xi(\mathfrak{G})\|}$. This is indeed true, as we will formally prove soon (Theorem 26).

The lvDAG facet of synchronizations has that the constructed indeterministic structural system of a synchronization induces a probability distribution using the recursive factorization in a similar manner to Equation 2. Based on Equation 7 this distribution is "coherent" in the sense that for $\mathcal{P}\in\mathfrak{A}_{l_1}$ and $\mathcal{Q}\in\mathfrak{A}_{l_2}$ $(0\leq l_1, l_2<\|\Xi(\mathfrak{G})\|)$, $\mathcal{P}\equiv\mathcal{Q}$ implies $\mathcal{P}=\mathcal{Q}$. The margin of this probability over $\boldsymbol{\mathcal{A}}$ is the same as the measure induced by the lvDAG. We denote the mapping that brings every indeterministic structural system of the synchronization to the induced probability measure over $\boldsymbol{\mathcal{A}}$ using $\Pi:\Xi(\mathfrak{d}(\mathfrak{G}))\to\mathfrak{q}(\mathfrak{G})$. For this induction, Equation 2 is applicable. It must therefore be trivial that:

$$\Pi(\Xi_{\mathfrak{G}}(\{\kappa_{\mathcal{A}}\}_{\mathcal{A}\in\boldsymbol{\mathcal{A}}}))=\Pi(\{\kappa_{\mathcal{A}}\}_{\mathcal{A}\in\boldsymbol{\mathcal{A}}}). \tag{9}$$

We would like to find the optimal indeterministic structural system of the pmDAG $\mathfrak{G}$ and we use the synchronization of $\mathfrak{G}$ for doing so. In what follows, we prove that the optimal indeterministic structural system of $\mathfrak{G}$ can be derived by minimizing the distance between the observed distribution and the distribution induced by the synchronization.

**Theorem 26** *Let* $\mathfrak{G}=\langle\boldsymbol{\mathcal{A}}\equiv\boldsymbol{\mathcal{V}}\dot\cup\boldsymbol{\mathcal{L}},\hookrightarrow\rangle$ *be a pmDAG and* $V$ *a sequence of observations* $(\boldsymbol{v})_{\boldsymbol{v}\in V}$ *of* $\boldsymbol{\mathcal{V}}$. *We let* $\mathfrak{I}\subseteq\mathfrak{D}(\mathfrak{G})$ *be an arbitrary collection and* Err *an arbitrary distance between two distributions. Then, we have:*

$$\underset{\langle\mathbb{P},\Phi\rangle\in\mathfrak{I}}{\arg\min}\left\{\mathrm{Err}\Big(\Pi(\langle\mathbb{P},\Phi\rangle)_{\boldsymbol{\mathcal{V}}};\mathbb{P}^{V}_{\boldsymbol{\mathcal{V}}}\Big)\right\}=\mathrm{I}_{\mathfrak{G}}^{-1}\left(\Xi_{\mathfrak{G}}^{-1}\left(\underset{*\in\Xi_{\mathfrak{G}}(\mathrm{I}_{\mathfrak{G}}(\mathfrak{I}))}{\arg\min}\left\{\mathrm{Err}\Big(\Pi(*);\mathbb{P}^{V}_{\boldsymbol{\mathcal{V}}}\Big)\right\}\right)\right), \tag{10}$$

*where* $*\equiv\{\kappa_{\mathcal{A}|l}\}_{\mathcal{A}\in\mathfrak{A}_l,\, 0\leq l<\|\Xi(\mathfrak{G})\|}$. Err *can be the KL divergence.*

**Proof** Trivial based on Equation 9 and the fact that $\mathrm{I}_{\mathfrak{G}}$ in Definition 12 and $\Xi_{\mathfrak{G}}$ in Definition 25 are injective. ■

Now, let us get back to Theorem 20 in §4. We may simply combine it with Theorem 26 and, as the most important claim of this section, conclude that we can indeed find the MLE of a pmDAG by utilizing its synchronization and finding the indeterministic structural system of the synchronization that minimizes the KL divergence.

**Corollary 27** *Let* $\mathfrak{G}=\langle\boldsymbol{\mathcal{A}}\equiv\boldsymbol{\mathcal{V}}\dot\cup\boldsymbol{\mathcal{L}},\hookrightarrow\rangle$ *be a pmDAG and* $V$ *a sequence of observations* $(\boldsymbol{v})_{\boldsymbol{v}\in V}$ *of* $\boldsymbol{\mathcal{V}}$. *Then, we have:*

$$\mathrm{MLE}_{\mathfrak{G},V}(*)=\mathrm{I}_{\mathfrak{G}}^{-1}\left(\Xi_{\mathfrak{G}}^{-1}\left(\underset{**\in\Xi_{\mathfrak{G}}(\mathrm{I}_{\mathfrak{G}}(*))}{\arg\min}\left\{\mathrm{KL}\Big(\Pi(**)\,\|\,\mathbb{P}^{V}_{\boldsymbol{\mathcal{V}}}\Big)\right\}\right)\right),$$

*where* $*\equiv\mathfrak{D}^{\mathrm{G}(0,1)}(\mathfrak{G})$ *and* $**\equiv\{\kappa_{\mathcal{A}|l}\}_{\mathcal{A}\in\mathfrak{A}_l,\, 0\leq l<\|\Xi(\mathfrak{G})\|}$.

Before concluding this section, we introduce a nice property of synchronizations. The *synchronized factorization property* gives a general rule for parameterizing a synchronization.

**Theorem 28 (synchronized factorization property)** *Let* $\Xi(\mathfrak{G}) = \langle \mathfrak{G}, \mathfrak{A}, \leq \rangle$ *be the synchronization of the lvDAG* $\mathfrak{G}$. *Moreover, let* $\left\{\kappa_{\mathcal{A}|l}\right\}_{\mathcal{A}\in\mathfrak{A}_l, 0\leq l<\|\Xi(\mathfrak{G})\|} \in \Xi(\eth(\mathfrak{G}))$ *be a indeterministic structural system of* $\Xi(\mathfrak{G})$. *If* $\left\{\kappa_{\mathcal{A}|l}\right\}_{\mathcal{A}\in\mathfrak{A}_l, 0\leq l<\|\Xi(\mathfrak{G})\|}$ *induces* $\mathbb{P}_{\boldsymbol{\mathcal{A}}}$, *then:*

$$\mathbb{P}_{\boldsymbol{\mathcal{U}}}(\boldsymbol{X}) = \mathbb{E}_{\mathrm{pa}_{\Xi(\mathfrak{G})|l}(\boldsymbol{\mathcal{U}})}\left[\prod_{\mathcal{U}\in\boldsymbol{\mathcal{U}}} \kappa_{\mathcal{U}|l}(X_{\mathcal{U}}|\cdot)\right], \qquad \forall\ \boldsymbol{X} = \prod_{\mathcal{U}\in\boldsymbol{\mathcal{U}}} X_{\mathcal{U}}, X_{\mathcal{U}} \in \sigma\Omega_{\mathcal{U}} \tag{11}$$

*holds for any* $\boldsymbol{\mathcal{U}} \subseteq \mathfrak{A}_l$ $(0 \leq l < \|\Xi(\mathfrak{G})\|)$. *We call Equation 11 the* synchronized factorization property *(SFP) of* $\Xi(\mathfrak{G})$.

**Proof** See Proof of SFP in Appendix C. ■

The SFP is specifically useful when we want to parameterize pmDAGs with discrete distributions. However, we restrict attention to the Gaussian distributions. The next example shows how the SFP is used:

**Example 3** *Assume that* $\mathfrak{G} = \langle \boldsymbol{\mathcal{A}} \equiv \boldsymbol{\mathcal{V}} \dot{\cup} \boldsymbol{\mathcal{L}}, \hookrightarrow \rangle$ *is the pmDAG in Figure 7(a). The indeterministic structural system* $\{\kappa_{\mathcal{A}}\}_{\mathcal{A}\in\boldsymbol{\mathcal{A}}}$ *is given and the marginal probability distribution* $\mathbb{P}_{\boldsymbol{\mathcal{V}}}$ *is required. The required probability distribution is obtained by applying the synchronized factorization property on* $\Xi(\mathfrak{G})$ *twice:*

$$\begin{aligned}
\mathbb{P}_{\{\mathcal{A},\mathcal{X}\}}(A, X) &= \int_{\Omega_{\mathcal{A}}} \kappa_{\mathcal{A}|1}(A|a)\ \kappa_{\mathcal{X}|1}(X|a)\,\mathbb{P}_{\mathcal{A}}(\mathrm{d}a),\\
\mathbb{P}_{\{\mathcal{X},\mathcal{Y}\}}(X, Y) &= \iint_{\Omega_{\{\mathcal{A},\mathcal{X}\}}} \kappa_{\mathcal{X}|2}(X|x)\ \kappa_{\mathcal{X}|2}(Y|a,x)\,\mathbb{P}_{\{\mathcal{A},\mathcal{X}\}}(\mathrm{d}a, \mathrm{d}x),
\end{aligned}$$

*for* $A \in \sigma\Omega_{\mathcal{A}}$, $X \in \sigma\Omega_{\mathcal{X}}$, *and* $Y \in \sigma\Omega_{\mathcal{Y}}$.

## 6 Parameter-optimization of pmDAGs with Gaussian Distributions

As described in §4, we would like to find an structural system that is (not necessarily uniquely) optimal. As a reminder, we are going to propose a solution for finding an indeterministic structural system of Equation 5. When we are working in the parametric setting, as in the Gaussian setting, this objective is equivalent to finding only the parameters of the indeterministic structural system that is optimal.

This constructs $\left\{\kappa_{\mathcal{A}|l}\right\}_{\mathcal{A}\in\mathfrak{A}_l, 0\leq l<\|\Xi(\mathfrak{G})\|} = \Xi_{\mathfrak{G}}(\mathrm{I}_{\mathfrak{G}}(\langle\mathbb{P}, \Phi\rangle))$ for $\langle\mathbb{P}, \Phi\rangle \in \mathfrak{D}^{\mathrm{G}(0,1)}(\mathfrak{G})$, which is the indeterministic structural system corresponding to the structural system $\langle\mathbb{P}, \Phi\rangle$ of $\mathfrak{G}$ (Definition 25):

$$\kappa_{\mathcal{A}|l}\left((-\infty, v]\,\middle|\,\mathrm{pa}_{\Xi(\mathfrak{G})|l}(\mathcal{A})\right) := \begin{cases} \varphi(v) & l = 0,\\ \mathbb{1}_{(-\infty,v]}\left(W_{(l)}^{\top}\ \mathrm{pa}_{\Xi(\mathfrak{G})|l}(\mathcal{A})\right) & l > 0. \end{cases} \tag{12}$$

where $\varphi$ is the standard Gaussian cumulative distribution function (CDF). The parameters of the kernels $\left\{\kappa_{\mathcal{A}|l}\right\}_{\mathcal{A}\in\mathfrak{A}_l, 0<l<\|\Xi(\mathfrak{G})\|}$ are the set of weight matrices $\boldsymbol{W} = \left\{W_{(l)}\right\}_{0\leq l<\|\Xi(\mathfrak{G})\|}$ $(W_{(l)} = \left[w_{(l)\mathcal{I}\mathcal{J}}\right],$

$\mathcal{I} \in \mathfrak{A}_{l-1}$, $\mathcal{J} \in \mathfrak{A}_l$) subject to:

$$w_{(l)\mathcal{I}\mathcal{J}} = \begin{cases} 1 & \mathcal{I} \equiv \mathcal{J}, \\ 0 & \mathcal{I} \notin \mathrm{pa}_{\Xi(\mathfrak{G})|l}(\mathcal{J}). \end{cases} \tag{13}$$

We can therefore assume that $\kappa_{\mathfrak{A}_l|l}$ $(0 < l < \|\Xi(\mathfrak{G})\|)$ of Definition 25 are linear transformations with parameters $W_{(l)}$. Equation 12, analogical to a dense (linear) neural-network layer, leads us to the neural network view of the synchronization. We primarily take a neural network approach to find the optimal parameters $\boldsymbol{W}$ of $\Xi(\mathfrak{G})$.

In light of the duality between the synchronization of a pmDAG and a neural network, we start by looking at the problem from the perspective of training a feed-forward neural network. This allows us to retrieve some initial set of equations. Then, we move forward from that naïve view and propose our devised algorithm that outperforms the neural network-based algorithm. Once we have climbed the ladder, we kick it off.[8] We name the implemented algorithm, the "SN$^2$" (structural system neural network) algorithm.

## 6.1 Neural Network-Based Solution

A feed-forward neural network minimizes a loss function using gradient-based methods. If we denote the (biased) sample covariance matrix using $\hat{\Sigma}$, a *loss function* is any function $\mathrm{Err} : \mathrm{dom}(\boldsymbol{W}) \times \mathrm{dom}(\hat{\Sigma}) \to \mathbb{R}$ ($\mathrm{dom}(\cdot)$ being the domain of $\cdot$) the minimizing parameters of which is desired:

$$\boldsymbol{W}^* = \operatorname*{arg\,min}_{\boldsymbol{W}} \left\{ \mathrm{Err}\left(\boldsymbol{W}; \hat{\Sigma}\right) \right\}, \tag{14}$$

where $\boldsymbol{W}^*$ is the set of parameters of $\left\{\Phi_{\mathcal{A}|l}\right\}^*_{\mathcal{A}\in\mathfrak{A}_l, 0\le l<\|\Xi(\mathfrak{G})\|}$. We review possible definitions of Err in full details in §6.1.1. The canonical loss function will obviously be the KL divergence, as it will give us the maximum likelihood estimation (Theorems 20 and 26).

The neural network-based parameterizer optimizes the loss function using iterative gradient descent. That is, it obtains $\boldsymbol{W}^*$ by computing $\mathrm{Err}\left(\boldsymbol{W}; \hat{\Sigma}\right)$ and its derivative w.r.t. $\boldsymbol{W}$, and updating $\boldsymbol{W}$ iteratively. The solver has a forward phase, where $\mathrm{Err}\left(\boldsymbol{W}; \hat{\Sigma}\right)$ is computed, and a backward phase, where $\nabla_{\boldsymbol{W}} \mathrm{Err}\left(\boldsymbol{W}; \hat{\Sigma}\right)$ is computed.

We denote the covariance of $\mathfrak{A}_l$ by $\Sigma_l(\boldsymbol{W})$. We let $\Sigma_0(\boldsymbol{W}) = \mathbf{I}_{\mathrm{card}(\mathfrak{A}_0)\times\mathrm{card}(\mathfrak{A}_0)}$ by definition. For $0 < l < \|\Xi(\mathfrak{G})\|$, $\Sigma_l(\boldsymbol{W})$ is defined using the recursive equation:

$$\Sigma_l(\boldsymbol{W}) \;=\; W_{(l)}^{\top}\; \Sigma_{l-1}(\boldsymbol{W})\; W_{(l)}. \tag{15}$$

On way to derive Equation 15 is using the SFP. It can also derived using the functional relationships between the layers.

In the backward phase, the constant parameters in $\boldsymbol{W}$, as described by Equation 13, need to remain unchanged. Therefore, we define the set of the *masking* matrices $\boldsymbol{M} = \left\{M_{(l)}\right\}_{0<l<\|\Xi(\mathfrak{G})\|}$ that project $\boldsymbol{W}$ to the non-constant parameter space, with $M_{(l)} = \left[m_{(l)\mathcal{I}\mathcal{J}}\right]$ ($\mathcal{I} \in \mathfrak{A}_{l-1}$, $\mathcal{J} \in \mathfrak{A}_l$), and:

$$m_{(l)\mathcal{I}\mathcal{J}} = \begin{cases} 1 & \mathcal{I} \in \mathrm{pa}_{\Xi(\mathfrak{G})|l}(\mathcal{J}) \wedge (\mathcal{I} \not\equiv \mathcal{J}), \\ 0 & \text{otherwise.} \end{cases}$$

8. A metaphor set out by Ludwig Wittgenstein (Wittgenstein, 2023).

**Algorithm 1:** The neural network-based solver.

**gets** : $(\boldsymbol{W}, \boldsymbol{M})$: initial weights, set of weight derivation masks,
$(\eta, I, \varepsilon)$ : optimization rate, maximum # of iterations, minimum loss improvement,
$\hat{\Sigma}$ : sample covariance matrix.

**returns** : $(\boldsymbol{W}^*, \Sigma^*)$: set of optimized weights, optimal covariance matrix.

1 **begin**
2 $L \leftarrow \|\Xi(\mathfrak{G})\|$ `// Initialization`
3 $\mathrm{Err}_{(0)} \leftarrow \infty$
4 **for** $l \leftarrow 1, \ldots, L$ **do**
5 $W_{(l)} \leftarrow W_{(l)} + M_{(l)} \circ \mathrm{NormalRand}\big(M_{(l)}.\mathrm{shape}, \mu = -1\big)$
6 **for** $i \leftarrow 1, \ldots, I$ **do**
7 forward() `// Forward phase`
8 $\mathrm{Err}_{(i)} \leftarrow \mathrm{tr}\big(\hat{\Sigma}^{-1}\Sigma\big) - \ln|\Sigma|$ `// (Equation 26; alt. 28 or 30)`
9 **if** $\big|\mathrm{Err}_{(i)} - \mathrm{Err}_{(i-1)}\big| \leq \varepsilon$ **then**
10 **return**
11 $\Delta\Sigma \leftarrow \hat{\Sigma}^{-1} - \Sigma^{-1}$ `// (Equation 27; alt. 29 or 31)`
12 backward() `// Backward phase`
13 **return**
14 **begin** *forward()*
15 $\Sigma \leftarrow \mathbf{I}$
16 **for** $l \leftarrow 1, \ldots, L$ **do**
17 $\Lambda_{(l)} \leftarrow \Sigma\, W_{(l)}$ `// (Equation 16)`
18 $\Sigma \leftarrow W_{(l)}^{\top}\, \Lambda_{(l)}$ `// (Equation 17)`
19
20 **begin** *backward()*
21 **for** $l \leftarrow L, \ldots, 1$ **do**
22 $\Delta W \leftarrow M_{(l)} \circ \big(\Lambda_{(l)}\, \Delta\Sigma\big)$ `// (Equation 18)`
23 $\Delta\Sigma \leftarrow 2\, W_{(l)}\, \Delta\Sigma\, W_{(l)}^{\top}$ `// (Equation 19)`
24 $W_{(l)} \leftarrow \mathrm{Optimize}\big(W_{(l)}, \Delta W, \eta\big)$ `// (Equation 20)`

Letting

$$\Lambda_{(l)} = \Sigma_{l-1}(\boldsymbol{W})\, W_{(l)}, \tag{16}$$

allows us to redefine 15 as

$$\Sigma_l(\boldsymbol{W}) \;=\; W_{(l)}^{\top}\;\Lambda_{(l)}, \tag{17}$$

decompose the gradient to partial derivatives for each layer $0 < l < \|\Xi(\mathfrak{G})\|$, and perform a step-by-step gradient calculation in the backward phase:

$$\frac{\partial \mathrm{Err}(\boldsymbol{W}; S)}{\partial W_{(l)}} \propto M_{(l)} \circ \left(\Lambda_{(l)}\, \frac{\partial \mathrm{Err}(\boldsymbol{W}; S)}{\partial \Sigma_l(\boldsymbol{W})}\right), \tag{18}$$

**Algorithm 2:** The neural network-based solver forward and backward phase with accumulated weights.

1 **begin** *forward()*
2 $\quad W^{\text{acc}}_{(0)} \leftarrow \mathbf{I}$
3 $\quad$ **for** $l \leftarrow 1, \ldots, L$ **do**
4 $\quad\quad W^{\text{acc}}_{(l)} \leftarrow W^{\text{acc}}_{(l-1)}\, W_{(l)}$ // (Equation 22)
5 $\quad \Sigma \leftarrow W^{\text{acc}\top}_{(L)}\, W^{\text{acc}}_{(L)}$ // (Equation 23)
6
7 **begin** *backward()*
8 $\quad \Omega \leftarrow W^{\text{acc}}_{(L)}\, \Delta\Sigma$
9 $\quad$ **for** $l \leftarrow L, \ldots, 1$ **do**
10 $\quad\quad \Delta W \leftarrow M_{(l)} \circ \left( W^{\text{acc}}_{(l-1)}{}^{\top}\, \Omega \right)$ // (Equation 25)
11 $\quad\quad \Omega \leftarrow \Omega\, W^{\top}_{(l)}$ // (Equation 24)
12 $\quad\quad W_{(l)} \leftarrow \text{Optimize}\left( W_{(l)}, \Delta W, \eta \right)$ // (Equation 20)

where $\circ$ is the Hadamard product and,

$$\frac{\partial \text{Err}(\boldsymbol{W}; S)}{\partial \Sigma_{l-1}(\boldsymbol{W})} = 2\, W_{(l)}\, \frac{\partial \text{Err}(\boldsymbol{W}; S)}{\partial \Sigma_{l}(\boldsymbol{W})}\, W^{\top}_{(l)}. \tag{19}$$

Once random values are assigned to $\boldsymbol{W}$, the solver begins the forward phase, in which it computes $\Sigma_l(\boldsymbol{W})$ (Eq. 15) step-wisely from layer 1 through layer $\|\Xi(\mathfrak{G})\|$, and then, (a monotonic function of) Err (see §6.1.1). In the backward phase, it takes the partial derivative of (the monotonic function of) Err w.r.t. $\Sigma_{\|\Xi(\mathfrak{G})\|}(\boldsymbol{W})$ (*ibid.*) and backpropagates it over each layer (Eq. 18), then updates $\boldsymbol{W}$ by subtracting (a multiplier of) the partial derivatives from $\boldsymbol{W}$, starting from layer $\|\Xi(\mathfrak{G})\| - 1$ back to layer 1. The updating procedure is done using an optimizer function, i.e.

$$\boldsymbol{W}^{(i)} = \text{Optimize}\left( \boldsymbol{W}^{(i-1)}, \underset{\boldsymbol{W}}{\nabla} \text{Err}(\boldsymbol{W}; S)\,; \eta \right). \tag{20}$$

Here, $\eta$ is the hyper-parameters of the optimizer. In the simplest case, Optimize obeys the following equation:

$$\text{Optimize}(\cdot) = \boldsymbol{W}^{(i-1)} - \eta \cdot \underset{\boldsymbol{W}}{\nabla} \text{Err}(\boldsymbol{W}; S), \tag{21}$$

with $\eta$ being a positive real number indicating the learning rate. Yet, any other optimizer is applicable. This entire process forms Algorithm 1. We call this the *covariance method*. Alternatively, one can use what we call the *accumulation method*. Define the set $\boldsymbol{W}^{\text{acc}} = \left\{ W^{\text{acc}}_{(l)} \right\}_{0 \le l < \|\Xi(\mathfrak{G})\|}$ of the *accumulated weight matrices* as:

$$W^{\text{acc}}_{(l)} = \begin{cases} \mathbf{I} & l = 0, \\ \prod_{i=1}^{l} W_{(i)} & \text{otherwise.} \end{cases} \tag{22}$$

It is easy to show that:

$$\Sigma_{\|\Xi(\mathfrak{G})\|-1}(\boldsymbol{W}) = {W^{\text{acc}}_{(\|\Xi(\mathfrak{G})\|-1)}}^{\top} W^{\text{acc}}_{(\|\Xi(\mathfrak{G})\|-1)}. \tag{23}$$

Moreover, we define new matrices $\boldsymbol{\Omega} = \{\Omega_{(l)}\}$ $(0 < l \leq \|\Xi(\mathfrak{G})\|)$:

$$\Omega_{(l)} = \begin{cases} W_{(l)} \left( \dfrac{\partial \text{Err}(\boldsymbol{W};S)}{\partial \Sigma_l(\boldsymbol{W})} \right) & l = \|\Xi(\mathfrak{G})\| - 1, \\ \Omega_{(l+1)} \, W_{(l)}^{\top} & \text{otherwise,} \end{cases} \tag{24}$$

which hand in the derivatives:

$$\frac{\partial \text{Err}(\boldsymbol{W};S)}{\partial W_{(l)}} \propto M_{(l)} \circ \left( W_{(l)}^{\text{acc}\top} \, \Omega_{(l)} \right), \tag{25}$$

This alternative approach is in Algorithm 2.

**Remark 29** *The ultimate goal is finding a structural system in* $\mathfrak{D}^{\mathrm{G}(0,1)}(\mathfrak{G})$ *(Equation 5). We use an implicit premise in our neural network approach. For our solution to work, the* Optimize *function should not generate a set of weight matrices* $\boldsymbol{W}$ *corresponding to an indeterministic structural system that is not in* $\mathfrak{D}^{\mathrm{G}(0,1)}(\mathfrak{G})$*. Here, this is automatically achieved, because for arbitrary* $\boldsymbol{W}$ *we get an indeterministic structural system of synchronization* $\{\kappa_{\mathcal{A}|l}\}_{\mathcal{A}\in\mathfrak{A}_l, 0\leq l<\|\Xi(\mathfrak{G})\|}$ *that can be converted to a structural system that is in* $\mathfrak{D}^{\mathrm{G}(0,1)}(\mathfrak{G})$*. For implementing similar algorithms with different assumptions, this premise should be considered carefully.*

#### 6.1.1 LOSS FUNCTIONS

The neural network-based method allows for various loss functions (see Hellinger (1909); Bhattacharyya (1946); Rubner et al. (1998); Kullback and Leibler (1951) for some variations). The loss function Err is arbitrary. KL divergence (Kullback and Leibler, 1951) has a closed-form formula for multivariate Gaussian distributions and it maximizes the likelihood of the sample data; see Corollary 27. Bhattacharyya distance (Bhattacharyya, 1946) provides a closed-formed formula for multivariate Gaussian variables, too. For Hellinger or squared Hellinger loss functions, the closed-form formulas exist, too. Conversely, the Earth-mover's distance relies on iterative methods that compute the square root of the (sample) covariance matrix and its derivative (Lin and Maji, 2017).

If $\text{KL}(\boldsymbol{\mathcal{A}} \| \boldsymbol{\mathcal{B}})$ is the KL divergence of random vector $\boldsymbol{\mathcal{A}}$ from $\boldsymbol{\mathcal{B}}$, or in case of two Gaussian distributions, their covariance matrices, then $\text{Err} = \text{Err}_{\text{KL}}$ must satisfy in 14 the following:

$$\boldsymbol{W}^* = \arg\min_{\boldsymbol{W}} \left\{ \text{KL}\left( \Sigma_{\|\Xi(\mathfrak{G})\|-1}(\boldsymbol{W}) \,\|\, \hat{\Sigma} \right) \right\}.$$

We define $\text{Err}_{\text{KL}}$ as:

$$\text{Err}_{\text{KL}}(\boldsymbol{W};\hat{\Sigma}) = \text{tr}\left( \hat{\Sigma}^{-1} \Sigma_{\|\Xi(\mathfrak{G}')\|-1}(\boldsymbol{W}) \right) - \ln \left| \Sigma_{\|\Xi(\mathfrak{G})\|-1}(\boldsymbol{W}) \right|, \tag{26}$$

with the partial derivative of $\text{Err}_{\text{KL}}$ with respect to the covariance matrix in the last layer of $\Xi(\mathfrak{G})$, $\Sigma_{\|\Xi(\mathfrak{G})\|-1}(\boldsymbol{W})$:

$$\frac{\partial \text{Err}_{\text{KL}}(\boldsymbol{W};\hat{\Sigma})}{\partial \Sigma_{\|\Xi(\mathfrak{G})\|-1}(\boldsymbol{W})} = \hat{\Sigma}^{-1} - \Sigma_{\|\Xi(\mathfrak{G})\|-1}(\boldsymbol{W})^{-1}. \tag{27}$$

For the Bhattacharyya loss function, let $\mathrm{Bha}(\boldsymbol{\mathcal{A}} \| \boldsymbol{\mathcal{B}})$ be the Bhattacharyya distance between two random vectors $\boldsymbol{\mathcal{A}}$ and $\boldsymbol{\mathcal{B}}$. The loss function must satisfy in 14 the following:

$$\boldsymbol{W}^* = \underset{\boldsymbol{W}}{\arg\min} \left\{ \mathrm{Bha}\left( \Sigma_{\|\Xi(\mathfrak{G})\|-1}(\boldsymbol{W}) \,\|\, \hat{\Sigma} \right) \right\},$$

We define it as an alternative to Equation 26 as:

$$\mathrm{Err}_{\mathrm{Bha}}(\boldsymbol{W}; \hat{\Sigma}) = \left( \frac{1}{2} \right)^{\mathrm{card}\left( \mathfrak{A}_{\|\Xi(\mathfrak{G})\|-1} \right)} \frac{\left| \Sigma_{\|\Xi(\mathfrak{G})\|-1}(\boldsymbol{W}) + \hat{\Sigma} \right|}{\left| \Sigma_{\|\Xi(\mathfrak{G})\|-1}(\boldsymbol{W})^{\frac{1}{2}} \right|}, \tag{28}$$

where $A^{\frac{1}{2}}$ is the square root of the positive (semi-)definite matrix $A$. The above Equation replaces Equation 27 with the following:

$$\frac{\partial \mathrm{Err}_{\mathrm{Bha}}(\boldsymbol{W}; \hat{\Sigma})}{\partial \Sigma_{\|\Xi(\mathfrak{G})\|-1}(\boldsymbol{W})} = \left( \Sigma_{\|\Xi(\mathfrak{G})\|-1}(\boldsymbol{W}) + \hat{\Sigma} \right)^{-1} - \frac{1}{2} \Sigma_{\|\Xi(\mathfrak{G})\|-1}(\boldsymbol{W})^{-1}. \tag{29}$$

Finally, if we use the squared Hellinger distance, denoted by $\mathrm{SH}(\boldsymbol{\mathcal{A}} \| \boldsymbol{\mathcal{B}})$, then $\mathrm{Err} = \mathrm{Err}_{\mathrm{SH}}$ must satisfy:

$$\boldsymbol{W}^* = \underset{\boldsymbol{W}}{\arg\min} \left\{ \mathrm{SH}\left( \Sigma_{\|\Xi(\mathfrak{G})\|-1}(\boldsymbol{W}) \,\|\, \hat{\Sigma} \right) \right\}.$$

The squared Hellinger distance defines a third loss function as:

$$\mathrm{Err}_{\mathrm{SH}}(\boldsymbol{W}; \hat{\Sigma}) = 1 - 2^{\frac{\mathrm{card}\left( \mathfrak{A}_{\|\Xi(\mathfrak{G})\|-1} \right)}{2}} \left| \Sigma_{\|\Xi(\mathfrak{G})\|-1}(\boldsymbol{W}) \right|^{1/4} \left| \Sigma_{\|\Xi(\mathfrak{G})\|-1}(\boldsymbol{W}) + \hat{\Sigma} \right|^{-1/2}, \tag{30}$$

which is an alternative to Equation 26. The following partial derivative of $\mathrm{Err}_{\mathrm{SH}}$ takes the place of Equation 27:

$$\frac{\partial \mathrm{Err}_{\mathrm{SH}}(\boldsymbol{W}; \hat{\Sigma})}{\partial \Sigma_{\|\Xi(\mathfrak{G})\|-1}(\boldsymbol{W})} = \frac{\left( \left| \Sigma_{\|\Xi(\mathfrak{G})\|-1}(\boldsymbol{W}) \right| - \left| \hat{\Sigma} \right| \right)}{\left( \left| \Sigma_{\|\Xi(\mathfrak{G})\|-1}(\boldsymbol{W}) \right| + \left| \hat{\Sigma} \right| \right)^{3/2}} \left| \Sigma_{\|\Xi(\mathfrak{G})\|-1}(\boldsymbol{W}) \right|^{1/4} \Sigma_{\|\Xi(\mathfrak{G})\|-1}(\boldsymbol{W})^{-\top}. \tag{31}$$

### 6.2 The SN$^2$ Algorithm

We took a look at the problem from a point of view that was purely based on neural networks. At first, it might seem that any algorithm implemented based on this perspective takes a massive amount of memory. To give you an intuition, in Figure 7, the parameters of $\mathfrak{G}$ is composed of the vector $\bar{w}_{\mathcal{X}}$ of length 1 and the vector $\bar{w}_{\mathcal{Y}}$ of length 2. However, when we transform it to a synchronization, as in Figure 7 (b), the set of (constant or variable) parameters is composed of the matrices $W_{(1)}$ and $W_{(2)}$, which are respectively $1 \times 2$ and $2 \times 2$, doubling the computational space.

Contrary to this impression and as shown in this subsection, with our Gaussian assumption, an algorithm can be implemented with almost no memory addition. In order to implement a solver with reduced memory foot-print, we will need to see what parameters are *preserved* in either the forward or backward phases, and which are unused. The following lemmata will lead us to the desired results. The main results of this subsection are in Corollaries 34 and 37 followed by Remark 38.

**Lemma 30** (**$\Sigma$–$\Lambda$ alternation**) *Let* $\Xi(\mathfrak{G}) = \langle \mathfrak{G}, \mathfrak{A}, \leq \rangle$ *be the synchronization of* $\mathfrak{G}$ *and* $0 < l < \|\Xi(\mathfrak{G})\|$. *If* $\mathcal{P} \in \mathfrak{A}_{l-1}$ *and* $\mathcal{P}, \mathcal{R} \in \mathfrak{A}_l$, *then* $\lambda_{(l)\mathcal{P}\mathcal{R}} = \sigma_{(l)\mathcal{P}\mathcal{R}}$.

**Proof**

$$\begin{aligned} \sigma_{(l)\mathcal{P}\mathcal{R}} &= \left[ W_{(l)}^{\top} \, \Sigma_{(l-1)} \, W_{(l)} \right]_{\mathcal{P}\mathcal{R}} && \text{(Equation 15)} \\ &= \sum_{\mathcal{U} \in \mathfrak{A}_{l-1}} \sum_{\mathcal{V} \in \mathfrak{A}_{l-1}} w_{(l)\mathcal{U}\mathcal{P}} \, \sigma_{(l-1)\mathcal{U}\mathcal{V}} \, w_{(l)\mathcal{V}\mathcal{R}}. \end{aligned}$$

We have $w_{(l)\mathcal{U}\mathcal{P}} = \delta_{\mathcal{U}\mathcal{P}}$, where $\delta_{..}$ is the Kronecker delta function. This means:

$$\begin{aligned} \sigma_{(l)\mathcal{P}\mathcal{R}} &= \sum_{\mathcal{V} \in \boldsymbol{\mathcal{U}}_{(l)}} \sigma_{(l-1)\mathcal{P}\mathcal{V}} \, w_{(l)\mathcal{V}\mathcal{R}} \\ &= \lambda_{(l)\mathcal{P}\mathcal{R}}. && \text{(Equation 16)} \; \blacksquare \end{aligned}$$

**Lemma 31** (**$\Sigma$ preservation**) *Let* $\mathfrak{G} = \langle \boldsymbol{\mathcal{A}} \equiv \boldsymbol{\mathcal{V}} \dot{\cup} \boldsymbol{\mathcal{L}}, \hookrightarrow \rangle$ *be a pmDAG and* $\Xi(\mathfrak{G}) = \langle \mathfrak{G}, \mathfrak{A}, \leq \rangle$ *its synchronization. Let* $0 < l < \|\Xi(\mathfrak{G})\| - z$ *with* $z \geq 0$. *Let* $\mathcal{P}, \mathcal{Q} \in \boldsymbol{\mathcal{A}}$ *be two random variables such that* $\mathcal{P}, \mathcal{Q} \in \mathfrak{A}_l$ *and* $\mathcal{P}, \mathcal{Q} \in \mathfrak{A}_{l+z}$. *Then,* $\sigma_{(l+z)\mathcal{P}\mathcal{Q}} = \sigma_{(l)\mathcal{P}\mathcal{Q}}$.

**Proof** The result is trivial for $z = 0$. If $z \geq 1$, then we know from Lemma 53 that $\mathcal{P}, \mathcal{Q} \in \mathfrak{A}_{l+z-1}$. Without loss of generality, we assume $\sigma_{(l+z-1)\mathcal{P}\mathcal{Q}} = \sigma_{(l)\mathcal{P}\mathcal{Q}}$ and prove $\sigma_{(l+z)\mathcal{P}\mathcal{Q}} = \sigma_{(l+z-1)\mathcal{P}\mathcal{Q}}$.

$$\begin{aligned} \sigma_{(l+z)\mathcal{P}\mathcal{Q}} &= \left[ W_{(l+z)}^{\top} \, \Sigma_{(l+z-1)} \, W_{(l+z)} \right]_{\mathcal{P}\mathcal{Q}} && \text{(Equation 15)} \\ &= \sum_{\mathcal{U} \in \mathfrak{A}_{l+z-1}} \sum_{\mathcal{V} \in \mathfrak{A}_{l+z-1}} w_{(l+z)\mathcal{U}\mathcal{P}} \, \sigma_{(l+z-1)\mathcal{U}\mathcal{V}} \, w_{(l+z)\mathcal{V}\mathcal{Q}} \\ &= \sum_{\mathcal{U} \in \mathfrak{A}_{l+z-1}} \sum_{\mathcal{V} \in \mathfrak{A}_{l+z-1}} \delta_{\mathcal{U}\mathcal{P}} \, \sigma_{(l+z-1)\mathcal{U}\mathcal{V}} \, \delta_{\mathcal{V}\mathcal{Q}} \\ &= \sigma_{(l+z-1)\mathcal{P}\mathcal{Q}}. \end{aligned}$$

Since equality is transitive, the consequent is proven. $\blacksquare$

**Lemma 32** (**$\Lambda$ preservation**) *Let* $\mathfrak{G} = \langle \boldsymbol{\mathcal{A}} \equiv \boldsymbol{\mathcal{V}} \dot{\cup} \boldsymbol{\mathcal{L}}, \hookrightarrow \rangle$ *be a pmDAG and* $\Xi(\mathfrak{G}) = \langle \mathfrak{G}, \mathfrak{A}, \leq \rangle$ *its synchronization. Let* $1 < l < \|\Xi(\mathfrak{G})\| - z$ *with* $z \geq 0$. *Let* $\mathcal{P}, \mathcal{Q} \in \boldsymbol{\mathcal{A}}$ *be two random variables such that* $\mathcal{P} \in \mathfrak{A}_{l-1}, \mathfrak{A}_{l+z-1}$ *and* $\mathcal{Q} \in \mathfrak{A}_l, \mathfrak{A}_{l+z}$. *Then,* $\lambda_{(l+z)\mathcal{P}\mathcal{Q}} = \lambda_{(l)\mathcal{P}\mathcal{Q}}$.

**Proof** For $z = 0$ the result is trivial. For $z > 0$, the result is simply derived from Lemmata 30 and 31. We have $\mathcal{P} \in \mathfrak{A}_l, \mathfrak{A}_{l+z}$ such that:

$$\begin{aligned} \lambda_{(l)\mathcal{P}\mathcal{Q}} &= \sigma_{(l)\mathcal{P}\mathcal{Q}} && \text{(Lemma 30)} \\ &= \sigma_{(l+z)\mathcal{P}\mathcal{Q}} && \text{(Lemma 31)} \\ &= \lambda_{(l+z)\mathcal{P}\mathcal{Q}}. && \text{(Lemma 30)} \; \blacksquare \end{aligned}$$

**Lemma 33 ($\boldsymbol{W}$ reduction)** *Let $\mathfrak{G} = \langle \boldsymbol{\mathcal{A}} \equiv \boldsymbol{\mathcal{V}} \dot\cup \boldsymbol{\mathcal{L}}, \hookrightarrow \rangle$ be a pmDAG and $\Xi(\mathfrak{G}) = \langle \mathfrak{G}, \mathfrak{A}, \leq \rangle$ its synchronization. Let $1 < l < \|\Xi(\mathfrak{G})\| - z$ with $z \geq 0$. If $\mathcal{P} \in \mathfrak{A}_{l-1}$ and $\mathcal{Q}_l$ are two random variables then $w_{(l)\mathcal{P}\mathcal{Q}}$ is a parameter only if $\mathrm{app}_{\leq}(\mathcal{Q}) = l$ and $\mathcal{P} \in \mathrm{pa}_{\mathfrak{G}}(\mathcal{Q})$.*

**Proof** This is trivial and can be derived from Equation 13. ■

**Corollary 34** *(i) From Lemmata 31 and 32, both $\Sigma$ and $\Lambda$ are preserved between two variables as the algorithm progresses in the forward phase. They are independent of the layer. Therefore it is needed to compute the values $\sigma_{\mathcal{R}\mathcal{S}}$ and $\lambda_{\mathcal{R}\mathcal{S}}$ for each pair of variables $\mathcal{R}$ and $\mathcal{S}$ once. By assumption, $\Sigma_{(0)} = \mathbf{I}$. Therefore, if $\mathcal{P}, \mathcal{Q} \in \mathfrak{A}_0$, $\sigma_{\mathcal{P}\mathcal{Q}}$ is not a parameter for $\left\{\kappa_{\mathcal{A}|l}\right\}_{\mathcal{A} \in \mathfrak{A}_l, 0 \leq l < \|\Xi(\mathfrak{G})\|}$. This means that if $\mathfrak{G} = \langle \boldsymbol{\mathcal{A}} \equiv \boldsymbol{\mathcal{V}} \dot\cup \boldsymbol{\mathcal{L}}, \hookrightarrow \rangle$, $\sigma_{\mathcal{P}\mathcal{Q}}$ must be computed only for $\mathcal{P} \in \boldsymbol{\mathcal{A}}$ and $\mathcal{Q} \in \boldsymbol{\mathcal{V}}$. It is clear from Lemma 30 that the same is true for $\lambda_{\mathcal{P}\mathcal{Q}}$.*

*(ii) From Lemma 33, $w_{(l)\mathcal{X}\mathcal{Y}}$ must be stored only when $\mathcal{Y}$ is an appears first. Therefore, for each $\mathcal{X} \in \boldsymbol{\mathcal{A}}$ and $\mathcal{Y} \in \boldsymbol{\mathcal{V}}$ of variables, at most one $w_{(l)\mathcal{X}\mathcal{Y}}$ must be stored. Note also that $w_{(l)\mathcal{X}\mathcal{Y}}$ is not a parameter when both $\mathcal{X}$ and $\mathcal{X}$ are in $\boldsymbol{\mathcal{L}}$.*

Although both $\Lambda$ and $\Sigma$ are preserved in the forward phase, this is not the case for $\partial\mathrm{Err}(\boldsymbol{W};\hat\Sigma)/\partial\Sigma_l(\boldsymbol{W})$. This means that for $\mathcal{P}, \mathcal{Q} \in \boldsymbol{\mathcal{A}}$, the equation $\partial\mathrm{Err}(\boldsymbol{W};\hat\Sigma)/\partial\sigma_{(l_1)\mathcal{P}\mathcal{Q}} = \partial\mathrm{Err}(\boldsymbol{W};\hat\Sigma)/\partial\sigma_{(l_2)\mathcal{P}\mathcal{Q}}$ does not necessarily hold for any two layers $0 < l_1, l_2 < \|\Xi(\mathfrak{G})\|$. Nevertheless, this does not cause any memory overhead in our algorithm; the values of $\partial\mathrm{Err}(\boldsymbol{W};\hat\Sigma)/\partial\sigma_{(l)\mathcal{P}\mathcal{Q}}$ can be constructed in and then removed from the memory layer-wisely while the Optimize function is performed in each layer.

**Lemma 35 ($\partial\mathrm{Err}/\partial\Sigma_{(l)}$ sufficiency)** *Let $\mathfrak{G} = \langle \boldsymbol{\mathcal{A}} \equiv \boldsymbol{\mathcal{V}} \dot\cup \boldsymbol{\mathcal{L}}, \hookrightarrow \rangle$ be a pmDAG and $\Xi(\mathfrak{G}) = \langle \mathfrak{G}, \mathfrak{A}, \leq \rangle$ its synchronization. For $0 < l < \|\Xi(\mathfrak{G})\|$, $\partial\mathrm{Err}/\partial\Sigma_{(l-1)}$ and $\partial\mathrm{Err}/\partial W_{(l)}$ only depend on $\partial\mathrm{Err}/\partial\Sigma_{(l)}$. That is:*

$$\left\{ \frac{\partial\mathrm{Err}\big(\boldsymbol{W};\hat\Sigma\big)}{\partial\Sigma_{l-1}(\boldsymbol{W})}, \frac{\partial\mathrm{Err}\big(\boldsymbol{W};\hat\Sigma\big)}{\partial W_{(l)}} \right\} \perp\!\!\!\perp \frac{\partial\mathrm{Err}\big(\boldsymbol{W};\hat\Sigma\big)}{\partial\Sigma_m(\boldsymbol{W})} \,\Bigg|\, \frac{\partial\mathrm{Err}\big(\boldsymbol{W};\hat\Sigma\big)}{\partial\Sigma_l(\boldsymbol{W})},$$

*for all $l < m < \|\Xi(\mathfrak{G})\|$.*

**Proof** Follows directly from Equations 18 and 19, because both $\partial\mathrm{Err}(\boldsymbol{W};S)/\partial\Sigma_{l-1}(\boldsymbol{W})$ and $\partial\mathrm{Err}(\boldsymbol{W};S)/\partial W_{(l-1)}$ are functions of $\partial\mathrm{Err}(\boldsymbol{W};S)/\partial\Sigma_l(\boldsymbol{W})$. ■

**Lemma 36 ($\partial\mathrm{Err}/\partial\sigma$ disregardance)** *Let $\mathfrak{G} = \langle \boldsymbol{\mathcal{A}} \equiv \boldsymbol{\mathcal{V}} \dot\cup \boldsymbol{\mathcal{L}}, \hookrightarrow \rangle$ be a pmDAG and $\Xi(\mathfrak{G}) = \langle \mathfrak{G}, \mathfrak{A}, \leq \rangle$ its synchronization. Let $\mathcal{A}, \mathcal{B} \in \boldsymbol{\mathcal{L}}$ and $\{\mathcal{X}, \mathcal{Y}\} \notin \boldsymbol{\mathcal{L}}$. For $0 < l < \|\Xi(\mathfrak{G})\|$, $\mathcal{X}, \mathcal{Y} \in \mathfrak{A}_{l-1}$ and $\mathcal{A}, \mathcal{B} \in \mathfrak{A}_l$ implies:*

*(a)* $\dfrac{\partial\mathrm{Err}\big(\boldsymbol{W};\hat\Sigma\big)}{\partial W_{(l)}} \perp\!\!\!\perp \dfrac{\partial\mathrm{Err}\big(\boldsymbol{W};\hat\Sigma\big)}{\partial\sigma_{(l)\mathcal{A}\mathcal{B}}}$, *and* *(b)* $\dfrac{\partial\mathrm{Err}\big(\boldsymbol{W};\hat\Sigma\big)}{\partial\sigma_{(l-1)\mathcal{X}\mathcal{Y}}} \perp\!\!\!\perp \dfrac{\partial\mathrm{Err}\big(\boldsymbol{W};\hat\Sigma\big)}{\partial\sigma_{(l)\mathcal{A}\mathcal{B}}}$.

**Proof**

(a) Assume $\mathcal{P} \in \mathfrak{A}_{l-1}$ and $\mathcal{Q} \in \mathfrak{A}_l$. If $\mathcal{Q} \in \mathfrak{A}_{l-1}$, then ${\partial \mathrm{Err}(\boldsymbol{W};S)}/{\partial w_{(l-1)\mathcal{P}\mathcal{Q}}} = 0$ because $m_{(l)\mathcal{P}\mathcal{Q}} = 0$ in Equation 18. Otherwise,

$$\frac{\partial \mathrm{Err}\big(\boldsymbol{W};\hat{\Sigma}\big)}{\partial w_{(l)\mathcal{P}\mathcal{Q}}} = \sum_{\mathcal{U}\in\mathfrak{A}_l} \lambda_{(l)\mathcal{P}\mathcal{U}} \frac{\partial \mathrm{Err}\big(\boldsymbol{W};\hat{\Sigma}\big)}{\partial \sigma_{(l)\mathcal{U}\mathcal{Q}}},$$

is not a function of ${\partial \mathrm{Err}(\boldsymbol{W};\hat{\Sigma})}/{\partial \sigma_{(l)\mathcal{A}\mathcal{B}}}$ as $\mathcal{B} \in \boldsymbol{\mathcal{L}}$ but $\mathcal{Q} \notin \boldsymbol{\mathcal{L}}$. Therefore, ${\partial \mathrm{Err}(\boldsymbol{W};\hat{\Sigma})}/{\partial W_{(l-1)}} \perp\!\!\!\perp {\partial \mathrm{Err}(\boldsymbol{W};\hat{\Sigma})}/{\partial \sigma_{(l)\mathcal{A}\mathcal{B}}}$.

(b) Without loss of generality, assume $\mathcal{X} \notin \boldsymbol{\mathcal{L}}$. For ${\partial \mathrm{Err}(\boldsymbol{W};\hat{\Sigma})}/{\partial \sigma_{(l-1)\mathcal{X}\mathcal{Y}}}$ to depend on ${\partial \mathrm{Err}(\boldsymbol{W};\hat{\Sigma})}/{\partial \sigma_{(l)\mathcal{A}\mathcal{B}}}$ we need that $\mathcal{X} \hookrightarrow \mathcal{A}$ or $\mathcal{X} \hookrightarrow \mathcal{B}$. Yet, $\mathcal{A}, \mathcal{B} \in \boldsymbol{\mathcal{L}}$, and neither $\mathcal{X} \hookrightarrow \mathcal{A}$ nor $\mathcal{X} \hookrightarrow \mathcal{B}$. Therefore, ${\partial \mathrm{Err}(\boldsymbol{W};\hat{\Sigma})}/{\partial \sigma_{(l-1)\mathcal{X}\mathcal{Y}}} \perp\!\!\!\perp {\partial \mathrm{Err}(\boldsymbol{W};\hat{\Sigma})}/{\partial \sigma_{(l)\mathcal{A}\mathcal{B}}}$. ■

**Corollary 37** *(i) From Lemma 36 (a), when computing* ${\partial \mathrm{Err}(\boldsymbol{W};\hat{\Sigma})}/{\partial \Sigma_{l-1}(\boldsymbol{W})}$ *and* ${\partial \mathrm{Err}(\boldsymbol{W};\hat{\Sigma})}/{\partial W_{(l-1)}}$*, there is no need to have* ${\partial \mathrm{Err}(\boldsymbol{W};\hat{\Sigma})}/{\partial \Sigma_m(\boldsymbol{W})}$ *stored for any other layer than $m = l$. From Lemma 36 (b), there is no need to have* ${\partial \mathrm{Err}(\boldsymbol{W};\hat{\Sigma})}/{\partial \sigma_{(l)\mathcal{A}\mathcal{B}}}$ *for any $\mathcal{A}, \mathcal{B} \in \boldsymbol{\mathcal{L}}$.*

*(ii) Notice that if $w_{(l)\mathcal{P}\mathcal{Q}}$ is not a parameter, there is no need to store the corresponding value of* ${\partial \mathrm{Err}(\boldsymbol{W};\hat{\Sigma})}/{w_{(l)\mathcal{P}\mathcal{Q}}}$. ${\partial \mathrm{Err}(\boldsymbol{W};\hat{\Sigma})}/{W_{(l)}}$ *takes the same size as $W_{(l)}$.*

The above results show that the $\mathrm{SN}^2$ algorithm only needs to compute some values that all fit in $\mathrm{card}(\boldsymbol{\mathcal{V}}) \times (\mathrm{card}(\boldsymbol{\mathcal{V}}) + \mathrm{card}(\boldsymbol{\mathcal{L}}))$ matrices. We formally present it in the following remark.

**Remark 38** *Given a pmDAG $\mathfrak{G} = \langle \boldsymbol{\mathcal{A}} \equiv \boldsymbol{\mathcal{V}} \dot{\cup} \boldsymbol{\mathcal{L}}, \hookrightarrow \rangle$ assume the $SN^2$ (the covariance method) algorithm is run for $\Xi(\mathfrak{G})$. The space complexity of the algorithm is* $O(\mathrm{card}(\boldsymbol{\mathcal{V}}) \times (\mathrm{card}(\boldsymbol{\mathcal{V}}) + \mathrm{card}(\boldsymbol{\mathcal{L}})))$*; if $\Sigma$, $\Lambda$, $\boldsymbol{W}$, $\frac{\partial \mathrm{Err}(\boldsymbol{W};\hat{\Sigma})}{\partial \boldsymbol{W}}$, $\frac{\partial \mathrm{Err}(\boldsymbol{W};\hat{\Sigma})}{\partial \Sigma_l(\boldsymbol{W})}$, and $\frac{\partial \mathrm{Err}(\boldsymbol{W};\hat{\Sigma})}{\partial \Sigma_{l+1}(\boldsymbol{W})}$ are stored in dense matrices, each take* $\mathrm{card}(\boldsymbol{\mathcal{V}}) \times (\mathrm{card}(\boldsymbol{\mathcal{V}}) + \mathrm{card}(\boldsymbol{\mathcal{L}}))$ *entries (Corollaries 34 and 37).*

The actual implementation of our algorithm, the $\mathrm{SN}^2$ algorithm, takes the structure of a pmDAG $\mathfrak{G} = \langle \boldsymbol{\mathcal{A}} \equiv \boldsymbol{\mathcal{V}} \dot{\cup} \boldsymbol{\mathcal{L}}, \hookrightarrow \rangle$ in the form of a $\mathrm{card}(\boldsymbol{\mathcal{A}}) \times \mathrm{card}(\boldsymbol{\mathcal{V}})$ binary matrix and an observed sample covariance matrix $\hat{\Sigma}$ corresponding to the distribution $\mathbb{P}^V_{\boldsymbol{\mathcal{V}}}$ and randomly returns (the parameters $\bar{\boldsymbol{w}} = \{\bar{\boldsymbol{w}}_{\mathcal{V}}\}_{\mathcal{V}\in\mathrm{nroot}(\mathfrak{G})}$ of) an optimal indeterministic structural system $\langle \mathbb{P}, \Phi \rangle^* \in \mathrm{MLE}_{\mathfrak{G},V}\big(\mathfrak{D}^{\mathrm{G}(0,1)}(\mathfrak{G})\big)$ as in Equation 5.

### 6.3 Implementation

We have implemented the algorithm on CUDA so that it runs in parallel for each variable on layer $l$. The algorithm is an extension of the PyTorch® library. Our PyTorch®-compatible CUDA implementation of the $\mathrm{SN}^2$ algorithm is accessible via https://github.com/msaremi/sn2.

## 7 Causal Identifiability

So far, we considered no causal meaning for a structural system and its corresponding lvDAG. These structures are however an essential tool that are widely used in causal inference. In this section, we talk about a central question in this realm called the *[causal] effect identifiability* (Bareinboim et al. (2020, Def. 18) and Pearl et al. (2000, p. 77)). As we will make it formal, effect identifiability assumes an unknown "causally correspondent" structural system and inspects whether layer $\mathfrak{L}_2$ (interventional) queries can be identified if knowledge of its pmDAG and of the marginal distribution over the visible variables is present (Bareinboim et al., 2020). How one discovers the pmDAG in a problem domain is not an easy task *per se* (Dawid, 2010) and goes beyond the scope of this paper. Moreover, notice that finding an analytical solution to the problem of effect identifiability when the structural functions of Equation 1 are linear is considered a difficult task (Wright, 1921; Simon and Simon, 1977; Fisher, 1966; Drton et al., 2011) and our Gaussian case is not an exception. (Note that Gaussianity implies linearity; see Theorem 44.)

As the focal point of this section, we assume an underlying deterministic structural system that generates the data and call it the *[causally] correspondent* (ground-truth) deterministic structural system. In essence, in a correspondent structural system, each variable $\mathcal{N} \in \mathrm{nroot}(\mathfrak{G})$ calculates the function $\Phi_{\mathcal{N}}$ in Equation 1 and takes its value accordingly based on the values of its parents. This function is sometimes called a mechanism and is considered independent of the mechanism of other variables (Peters et al., 2017, ch. 2). We denote the correspondent structural system using $\langle \mathbb{P}, \Phi \rangle^+$. We assume that the lvDAG of this correspondent structural system is a pmDAG, call it the *[causally] correspondent* pmDAG, and denote it using $\mathfrak{G}^+$. We also assume that $\langle \mathbb{P}, \Phi \rangle^+ \in \mathfrak{D}^{\mathrm{G}(0,1)}(\mathfrak{G}^+)$, which means that the structural system is Gaussian. Note that if our correspondent lvDAG is not a pmDAG, we can convert it to a pmDAG using the process of deterministic exogenization introduced in §3. Also, remember that standardization of root variables is without loss of generality (see §B.2).

The causal identifiability problem is, therefore, as follows: The pmDAG $\mathfrak{G}^+ = \langle \boldsymbol{\mathcal{A}} \equiv \boldsymbol{\mathcal{V}} \dot\cup \boldsymbol{\mathcal{L}}, \hookrightarrow \rangle$ is known. The structural system $\langle \mathbb{P}, \Phi \rangle^+$ is unknown, yet we know $\langle \mathbb{P}, \Phi \rangle^+ \in \mathfrak{D}^{\mathrm{G}(0,1)}(\mathfrak{G}^+)$. Moreover, the induced distribution $\mathbb{P}_{\boldsymbol{\mathcal{V}}}$ is known. The distribution of $\boldsymbol{\mathcal{E}} \subseteq \boldsymbol{\mathcal{V}}$ if we intervened on $\boldsymbol{\mathcal{T}} \subseteq \boldsymbol{\mathcal{V}}$ and enforced them to take the value $\boldsymbol{t} \in \Omega_{\boldsymbol{\mathcal{T}}}$ is required. The identifiability problem asks whether this distribution can be uniquely identified. For discussing the effect identifiability, we will be working through some definitions and lemmata. We start with the definition of *mutilation*, which enforces variables to take specific values.

**Definition 39 (mutilation of structural system)** *Let $\langle \mathbb{P}, \Phi \rangle$ be a structural system of pmDAG $\mathfrak{G} = \langle \boldsymbol{\mathcal{A}} \equiv \boldsymbol{\mathcal{V}} \dot\cup \boldsymbol{\mathcal{L}}, \hookrightarrow \rangle$. The* mutilation *of $\langle \mathbb{P}, \Phi \rangle$ w.r.t. $\mathcal{V} = v$ ($\mathcal{V} \in \boldsymbol{\mathcal{V}}$, $v \in \Omega_{\mathcal{V}}$), denoted by $\langle \mathbb{P}, \Phi \rangle_{\mathcal{V}=v} = \langle \mathbb{P}', \Phi' \rangle$, is a structural system defined as follows. We define a new latent variable $\mathcal{E}_{\mathcal{V}} \notin \boldsymbol{\mathcal{A}}$ with $\mathbb{P}'_{\mathcal{E}_{\mathcal{V}}}(V) := \mathbb{1}_V(v)$ for all $V \in \sigma\Omega_{\mathcal{V}}$, and let*

$$\mathbb{P}'(\boldsymbol{X} \times V) := \mathbb{P}_{\boldsymbol{\mathcal{L}}}(\boldsymbol{X})\ \mathbb{P}'_{\mathcal{E}_{\mathcal{V}}}(V),$$

*for all $\boldsymbol{X} \in \sigma\Omega_{\boldsymbol{\mathcal{L}}}$ and $V \in \sigma\Omega_{\mathcal{E}_{\mathcal{V}}}$. Moreover, we define $\Phi'$ as*

$$\Phi'_{\mathcal{A}} := \begin{cases} \mathrm{id}_{\mathcal{E}_{\mathcal{V}}} & \mathcal{A} \equiv \mathcal{V}, \\ \Phi_{\mathcal{A}} & \mathcal{A} \not\equiv \mathcal{V}, \end{cases}$$

*for all $\mathcal{A} \in \mathrm{nroot}(\mathfrak{G})$.*

*The mutilation of $\langle \mathbb{P}, \Phi \rangle$ w.r.t. $\boldsymbol{\mathcal{U}} = \boldsymbol{u}$ ($\boldsymbol{\mathcal{U}} \subseteq \boldsymbol{\mathcal{V}}$ and $\boldsymbol{u} \in \Omega_{\boldsymbol{\mathcal{U}}}$), denoted by $\langle \mathbb{P}, \Phi \rangle_{\boldsymbol{\mathcal{U}}=\boldsymbol{u}}$, is done by mutilating $\langle \mathbb{P}, \Phi \rangle$ once for each $\mathcal{U} \in \boldsymbol{\mathcal{U}}$ (in an arbitrary order). The mutilation of a set $\mathfrak{D} \subseteq \mathfrak{D}(\mathfrak{G})$ of structural systems w.r.t. $\boldsymbol{\mathcal{U}} = \boldsymbol{u}$ is defined as $\mathfrak{D}_{\boldsymbol{\mathcal{U}}=\boldsymbol{u}} := \{\langle \mathbb{P}, \Phi \rangle_{\boldsymbol{\mathcal{U}}=\boldsymbol{u}} \,|\, \langle \mathbb{P}, \Phi \rangle \in \mathfrak{D}\}$.*

The mutilation of a structural system is what we introduced as performing an intervention: it enforces a set of observed variables $\boldsymbol{\mathcal{T}} \subseteq \boldsymbol{\mathcal{V}}$ to take specific values $\boldsymbol{t}$. Similar to the original structural system, the mutilated structural system also induces a distribution. As discussed, the task of effect identifiability is to test whether a margin of this induced probability distribution over an observed subset $\mathcal{E}$ of $\mathcal{V}$ is identifiable given the correspondent pmDAG $\mathfrak{G}^+$ and marginal distribution $\mathbb{P}_{\mathcal{V}}$. If identifiable, we call this distribution the *interventional distribution of* $\mathcal{E} \subseteq \mathcal{V}$ *in* $\langle \mathbb{P}, \Phi \rangle^+_{\mathcal{T}=t}$ and denote it using $\mathbb{P}_{\mathcal{E}|\mathrm{do}(\mathcal{T}=t)}$.

Let $\mathfrak{G} = \langle \mathcal{A} \equiv \mathcal{V} \dot\cup \mathcal{L}, \hookrightarrow \rangle$ be a pmDAG. For the set $\mathfrak{D} \subseteq \mathfrak{D}(\mathfrak{G})$ and the probability measure $\mathbb{P}_{\mathcal{V}}$ over $\mathcal{V}$, we define the function $\mathrm{SOL}_{\mathfrak{G}}$ as:

$$\mathrm{SOL}_{\mathfrak{G}}(\mathfrak{D}, \mathbb{P}_{\mathcal{V}}) \coloneqq \{ \langle \mathbb{P}, \Phi \rangle \in \mathfrak{D} \,|\, \Pi(\langle \mathbb{P}, \Phi \rangle)_{\mathcal{V}} = \mathbb{P}_{\mathcal{V}} \},$$

which is the set of all structural systems of $\mathfrak{G}$ in $\mathfrak{D}$ that induce $\mathbb{P}_{\mathcal{V}}$. We formally define the effect identifiability problem and subsequently explain it in normal text.

**Definition 40 (effect identifiability)** *We assume that (a)* $\mathbb{P}_{\mathcal{V}} \coloneqq \Pi(\langle \mathbb{P}, \Phi \rangle^+)_{\mathcal{V}}$*, and (b)* $\langle \mathbb{P}, \Phi \rangle^+ \in \mathfrak{D}$*. Then, knowing* $\mathfrak{G}^+$*,* the interventional distribution of $\mathcal{E} \subseteq \mathcal{V}$ in $\langle \mathbb{P}, \Phi \rangle^+_{\mathcal{T}=t}$ ($\mathcal{T} \subseteq \mathcal{V}$) ($\mathbb{P}_{\mathcal{E}|\mathrm{do}(\mathcal{T}=t)}$, for short) is identifiable from $\mathfrak{D}$ and $\mathbb{P}_{\mathcal{V}}$ *iff. the cardinality of the set*

$$\Pi(\mathrm{SOL}_{\mathfrak{G}^+}(\mathfrak{D}, \mathbb{P}_{\mathcal{V}})_{\mathcal{T}=t})_{\mathcal{E}} \tag{32}$$

*is* 1.

The mental picture of Equation 32 is as follows. We denote the set of all structural systems of $\mathfrak{G}$ in $\mathfrak{D}$ that are capable of inducing $\mathbb{P}_{\mathcal{V}}$ using $\mathrm{SOL}_{\mathfrak{G}^+}(\mathfrak{D}, \mathbb{P}_{\mathcal{V}})$. For the mutilation of each of these structural systems, an interventional distribution is induced. These distributions are written as $\Pi(\mathrm{SOL}_{\mathfrak{G}}(\mathfrak{D}, \mathbb{P}_{\mathcal{V}})_{\mathcal{T}=t})$. We are interested in the margin of this distribution over $\mathcal{E}$. Now, if the induced probabilities are unique over $\mathcal{E}$, i.e. if $\mathrm{card}(*) = 1$, we say that the interventional distribution of $\mathcal{E}$ given $\mathcal{T} = t$ is identifiable.

**Remark 41** *For the interventional distribution of* $\mathcal{E} \subseteq \mathcal{V}$ *in* $\langle \mathbb{P}, \Phi \rangle^+_{\mathcal{T}=t}$ *in Definition 40 and when* $\mathfrak{D} = \mathfrak{D}(\mathfrak{G}^+)$*, whether or not it is identifiable, the notation* $\mathbb{P}(\mathcal{E} | \mathrm{do}(\mathcal{T} = t))$ *is usually used in the Pearlian causal inference regime (Pearl et al., 2000; Pearl, 2012). It is an* $\mathfrak{L}_2$ *(interventional) query asking "what would the effect of enforcing* $\mathcal{T} = t$ *be on* $\mathcal{E}$*," and it is sometime identifiable given that one has observed V (or, more specifically, has learned* $\mathbb{P}_{\mathcal{V}}$*).*

Now, we specialize the aforementioned problem of identifiability to the Gaussian case by letting $\mathfrak{D} = \mathfrak{D}^{\mathrm{G}(0,1)}(\mathfrak{G})$ in Equation 32. We are now in a position to define the main theorem of this section that relates effect identifiability to the results of the previous sections.

**Theorem 42** *Let* $\mathfrak{G}^+ = \langle \mathcal{A} \equiv \mathcal{V} \dot\cup \mathcal{L}, \hookrightarrow \rangle$ *and* $\langle \mathbb{P}, \Phi \rangle^+ \in \mathfrak{D}^{\mathrm{G}(0,1)}(\mathfrak{G}^+)$*. Moreover, let V be a sequence of observations* $(v)_{v \in V}$ *of* $\mathcal{V}$ *such that* $\Pi(\langle \mathbb{P}, \Phi \rangle^+)_{\mathcal{V}} = \mathbb{P}^V_{\mathcal{V}}$*. Then, we have:*

$$\mathrm{SOL}_{\mathfrak{G}^+}(\mathfrak{D}, \mathbb{P}^V_{\mathcal{V}}) = \mathrm{MLE}_{\mathfrak{G}^+, V}(\mathfrak{D}).$$

**Proof** If $\Pi(\langle \mathbb{P}, \Phi \rangle^+)_{\mathcal{V}} = \mathbb{P}^V_{\mathcal{V}}$, this means that the pmDAG is capable of inducing $\mathbb{P}^V_{\mathcal{V}}$. Indeed, every structural system of the MLE induces $\mathbb{P}^V_{\mathcal{V}}$ and every structural system in $\mathfrak{D}^{\mathrm{G}(0,1)}(\mathfrak{G}^+)$ that induces $\mathbb{P}^V_{\mathcal{V}}$ is in the MLE. But, this is by definition $\mathrm{SOL}_{\mathfrak{G}}(\mathfrak{D}, \mathbb{P}^V_{\mathcal{V}})$. ■

**Algorithm 3:** The meta-algorithm for the effect identifiability problem.

**gets** : $\mathfrak{G}^+ = \langle \boldsymbol{\mathcal{A}} \equiv \boldsymbol{\mathcal{V}} \dot\cup \boldsymbol{\mathcal{L}}, \hookrightarrow \rangle$: the correspondent pmDAG,
$\boldsymbol{\mathcal{T}}, \boldsymbol{\mathcal{E}} \subseteq \boldsymbol{\mathcal{V}}$ : a pair of random vectors,
$\boldsymbol{t} \in \Omega_{\boldsymbol{\mathcal{T}}}$ : values for the random vector $\boldsymbol{\mathcal{T}}$,
$\mathbb{P}^V_{\boldsymbol{\mathcal{V}}} \in \mathfrak{P}^{\mathrm{G}(0)}_{\boldsymbol{\mathcal{V}}}$ : an observational distribution,
Fn : a parameter optimization algorithm (including SN$^2$),
$I \in \mathbb{N}$ : maximum number of iterations.

**returns** : **true** or **false**: **false** if $\mathbb{P}_{\boldsymbol{\mathcal{E}}|\mathrm{do}(\boldsymbol{\mathcal{T}}=\boldsymbol{t})}$ is detected as not identifiable from $\mathfrak{D}^{\mathrm{G}(0,1)}(\mathfrak{G}^+)$ and $\mathbb{P}^V_{\boldsymbol{\mathcal{V}}}$; **true**, otherwise.

1 **begin**
2 $\quad \langle \mathbb{P}, \Phi \rangle^* \leftarrow \mathrm{Fn}\big(\mathfrak{G}, \mathbb{P}^V_{\boldsymbol{\mathcal{V}}}\big)$
3 $\quad$ **if** $\mathrm{KL}\Big(\Pi\big(\langle \mathbb{P}, \Phi \rangle^*\big)_{\boldsymbol{\mathcal{V}}} \,\big\|\, \mathbb{P}^V_{\boldsymbol{\mathcal{V}}}\Big) > 0$ **then**
4 $\qquad$ **return false**
5 $\quad$ **for** $i \leftarrow 1, \ldots, I$ **do**
6 $\qquad \langle \mathbb{P}', \Phi' \rangle^* \leftarrow \mathrm{Fn}\big(\mathfrak{G}, \mathbb{P}^V_{\boldsymbol{\mathcal{V}}}\big)$
7 $\qquad$ **if** $\mathrm{KL}\Big(\Pi\big(\langle \mathbb{P}, \Phi \rangle^*_{\boldsymbol{\mathcal{T}}=\boldsymbol{t}}\big)_{\boldsymbol{\mathcal{E}}} \,\big\|\, \Pi\big(\langle \mathbb{P}', \Phi' \rangle^*_{\boldsymbol{\mathcal{T}}=\boldsymbol{t}}\big)_{\boldsymbol{\mathcal{E}}}\Big) > 0$ **then**
8 $\qquad\quad$ **return false**
9 $\quad$ **return true**

Theorem 42 suggests that if the solution space is not empty, the SN$^2$ algorithm is indeed capable of extracting a single member of the solution space. That is, we can obtain a structural system that induces the observed probability. The effect identifiability, as in Equation 32, requires the complete set of $\mathrm{SOL}_{\mathfrak{G}^+}\big(\mathfrak{D}^{\mathrm{G}(0,1)}(\mathfrak{G}^+), \mathbb{P}^V_{\boldsymbol{\mathcal{V}}}\big)$. We show that the problem of causal identifiability is asymptotically possible using the SN$^2$ algorithm when it is run multiple times. We can develop a meta-algorithm[9] that uses the SN$^2$ algorithm to asymptotically test whether the causal effect of one variable on another variable is identifiable. This will be the subject of the next subsection.

### 7.1 Causal Identifiability Meta-algorithm

Based on the discussion above, we propose a meta-algorithm that detects whether the distribution $\mathbb{P}_{\boldsymbol{\mathcal{E}}|\mathrm{do}(\boldsymbol{\mathcal{T}})}$ ($\boldsymbol{\mathcal{T}}, \boldsymbol{\mathcal{E}} \subseteq \boldsymbol{\mathcal{V}}$) is identifiable or not. In the most abstract sense, let us assume that Fn is an algorithm that takes the pmDAG $\mathfrak{G}^+ = \langle \boldsymbol{\mathcal{A}} \equiv \boldsymbol{\mathcal{V}} \dot\cup \boldsymbol{\mathcal{L}}, \hookrightarrow \rangle$ and the observational distribution $\mathbb{P}^V_{\boldsymbol{\mathcal{V}}} \in \mathfrak{P}^{\mathrm{G}(0)}_{\boldsymbol{\mathcal{V}}}$ and returns a $\langle \mathbb{P}, \Phi \rangle^* \in \mathrm{MLE}_{\mathfrak{G},V}\big(\mathfrak{D}^{\mathrm{G}(0,1)}(\mathfrak{G}^+)\big)$ at random. A candidate for such an algorithm is the SN$^2$ algorithm; it takes the structure of a pmDAG $\mathfrak{G}$ and the observed covariance matrix $\hat{\Sigma}$ and returns an optimal set of parameters $\bar{\boldsymbol{w}}$ that correspond to $\langle \mathbb{P}, \Phi \rangle^*$.

The meta-algorithm, Algorithm 3, works as follows. Initially, it retrieves an optimal structural system by means of the Fn algorithm. We assume that the Fn algorithm always converges. (This simplifying assumption is, however, not necessarily true for gradient descent-based algorithms such

9. We chose the name "meta-algorithm" to indicate that this algorithm is mounted to the SN$^2$ algorithm or any other algorithm. Indeed, one of the inputs of this meta-algorithm is another algorithm.

as the SN[2] algorithm.) The meta-algorithm then compares the induced probability distribution $\Pi\big(\langle \mathbb{P}, \Phi \rangle^*\big)_{\mathcal{V}}$ with the observed one $\mathbb{P}^V_{\mathcal{V}}$. If they are not equal, it means that the solution space is empty (see Theorem 42), and therefore, $\mathbb{P}_{\mathcal{E}|\mathrm{do}(\mathcal{T})}$ is not identifiable from $\mathfrak{D}^{\mathrm{G}(0,1)}(\mathfrak{G}^+)$ and $\mathbb{P}^V_{\mathcal{V}}$; the meta-algorithm returns **false**. Otherwise, the Fn algorithm is run multiple times and returns multiple deterministic structural systems $\langle \mathbb{P}', \Phi' \rangle^*$ at random. This is where we check whether the mutilated structural systems induce the same distribution over $\mathcal{E}$. If the the distributions are not equal (KL divergence is greater than 0), the meta-algorithm will detect $\mathbb{P}_{\mathcal{E}|\mathrm{do}(\mathcal{T})}$ as non-identifiable from $\mathfrak{D}^{\mathrm{G}(0,1)}(\mathfrak{G}^+)$ and $\mathbb{P}^V_{\mathcal{V}}$ and returns **false**.

If the algorithm does not return unequal distributions over $\mathcal{E}$ in $I$ iterations, the meta-algorithm will not recognize $\mathbb{P}_{\mathcal{E}|\mathrm{do}(\mathcal{T})}$ as non-identifiable and returns **true**. Notice that this meta-algorithm is refutatively sound in the sense that if the meta-algorithm returns **false**, then $\mathbb{P}_{\mathcal{E}|\mathrm{do}(\mathcal{T})}$ is definitely not identifiable from $\mathfrak{G}^+$, $\mathfrak{D}$, and $\mathbb{P}^V_{\mathcal{V}}$. On the other hand, this meta-algorithm is almost-surely asymptotically correct[10] in the sense that the greater the $I$, the higher the confidence that $\mathbb{P}_{\mathcal{E}|\mathrm{do}(\mathcal{T})}$ is identifiable from $\mathfrak{D}^{\mathrm{G}(0,1)}(\mathfrak{G}^+)$ and $\mathbb{P}^V_{\mathcal{V}}$ if the algorithm returns **true**, such that we can be almost certain about the identifiability as the number of iterations $I$ approaches $\infty$.

## 8 Experimental Results

We divide this section into two parts. In §8.1, we measure the performance of our SN[2] algorithm. In §8.2 we study the effect identifiability problem that we discussed in §7. Our tests were conducted on an NVIDIA GeForce GTX 970M CUDA 10.2 machine with 3GB of VRAM and a Core i7-6700HQ CPU. In addition to these, in Appendix D, we compare the numerical stability of different loss functions using the two proposed accumulation and covariance methods.

### 8.1 Ablation Study

We study the performance of the SN[2] algorithm in both the forward and backward phase. Our experiment compares the covariance and accumulation methods. For these tests, we set the number of threads per block to 256 and used the maximum shared memory for both the forward and backward CUDA kernels equal to 16KBs.

We conditioned our tests on three parameters, which define the characteristics of random pmDAGs. The first parameter was the number of visible variables $v \in \mathbb{N}$, which we set to 16, 32, 64, 128, 256, and 512. For each of these values, we defined the abundance of latent variables $l^* \in [0,1)$ from which the number of latent variables $l$ is derived using the following formula:

$$l := \left\lfloor \frac{l^*}{1 - l^*} v \right\rfloor + v.$$

When $l^* = 0$, for each visible variable we only have one auxiliary latent variable as the parent of a single visible variable. As $l^*$ grows, the number of latent variables $l$ tends to $\infty$. For our experiments, we assigned 0 and $1/2$ to $l^*$. Finally, we define the edge density $e^* \in [0,1]$, which

10. Xia et al. (2021) find whether $\mathbb{P}(\mathcal{E}|\mathrm{do}(\mathcal{T} = t))$ is identifiable by trying to extract a minimum and a maximum value for this expression. If the minimum and maximum values are above the error threshold, they report non-identifiability; otherwise they report identifiability. In the parametric case, like ours, one can always achieve this by defining an end-to-end model that penalizes for similar distributions of $\Pi\big(\langle \mathbb{P}, \Phi \rangle^*_{\mathcal{T}=t}\big)_{\mathcal{E}}$ in two "twin" networks. Note that, however, achieving this diversity is never guaranteed, making their solution also almost-surely asymptotically correct.

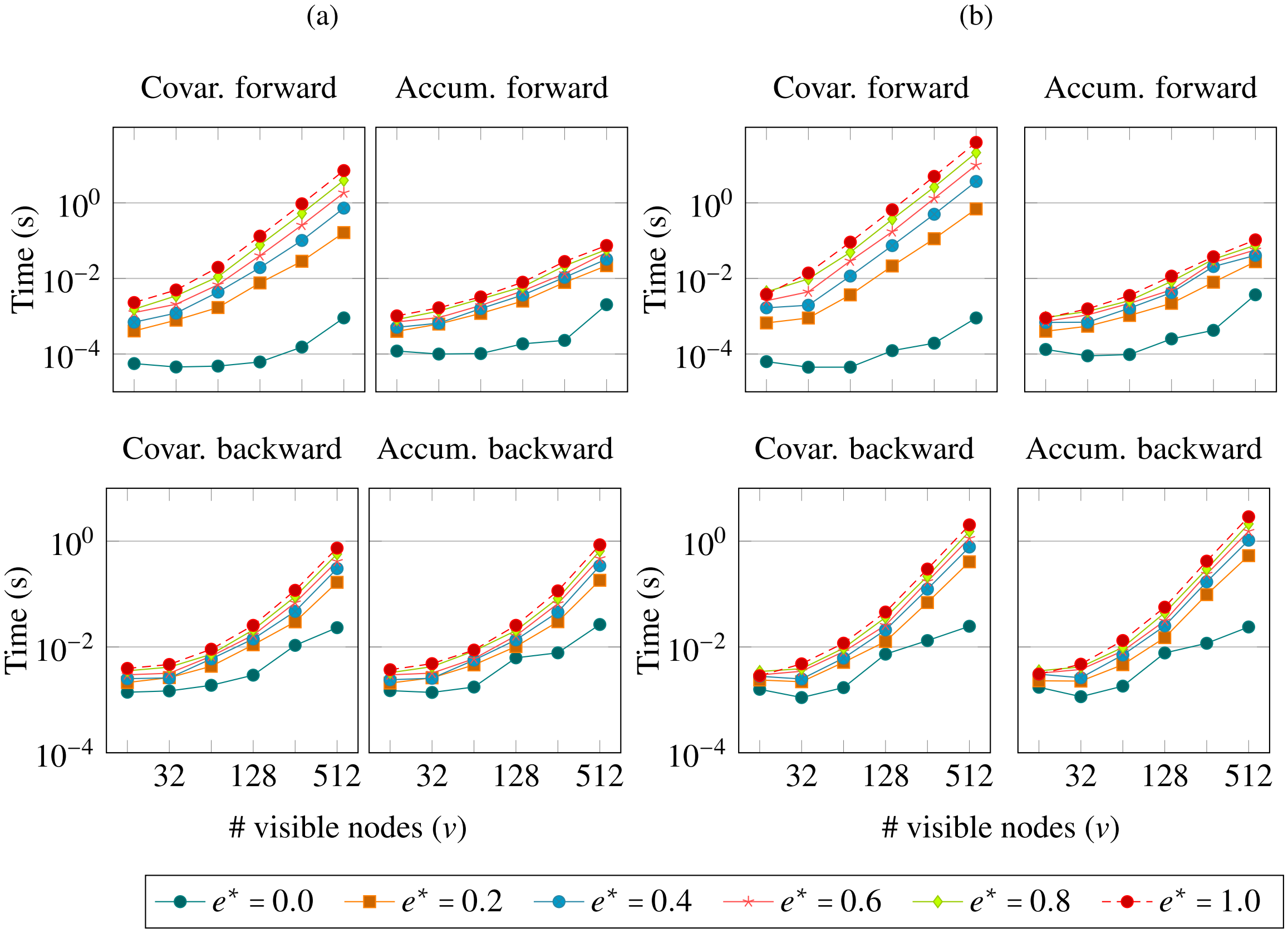

Figure 8: Log-log plots of the average performance time of the $SN^2$ algorithm using both the covariance method and the accumulation method computed for the forward and backward phases. (a) $l^* = 0$ (Markovian pmDAG) and (b) $l^* = 1/2$.

determines the number of edges $e$ on the pmDAG using the following formula:

$$e = \left\lfloor \left( lv + \frac{v(v-1)}{2} \right) e^* \right\rfloor + v.$$

If $e^*$ is set to 0.0 the only edges of the pmDAG are those that connect the auxiliary latent nodes to the visible nodes. When $e^* = 1.0$, the pmDAG is fully connected. $e^*$ takes the values of 0.0, 0.2, 0.4, 0.6, 0.8, and 1.0.

For each values of $v$, $l^*$ and $e^*$, we averaged the performance time on 5 randomly generated pmDAGs. Moreover, we used the standard KL divergence error ($\mathrm{Err} = \mathrm{Err}_{\mathrm{KL}}$) as our loss function. We measured the average performance time on the forward and backward phases separately for each of the covariance and accumulation methods. Figure 8 shows the results of our ablation study. As seen, the performance time of the forward phase for both the covariance and accumulation methods grows almost linearly with respect to the number of visible nodes, whereas the same performance time of the backward phase grows super-linearly. Another important observation is that the performance of the accumulation method is significantly higher than the covariance method, especially when the edge density of the pmDAG is high. For example, the time ratio of the covariance method to the accumulation method with $v = 512$, $l^* = {}^1\!/_2$ and $e^* = 0.4$ in the forward phase is ${}^{3.7\mathrm{s}}\!/_{0.04\mathrm{s}} \approx 92$. Therefore, using the accumulation method in these cases is recommended.

### 8.2 The Identifiability Problem

Given a pmDAG $\mathfrak{G} = \langle \boldsymbol{\mathcal{A}} \equiv \boldsymbol{\mathcal{V}} \dot{\cup} \boldsymbol{\mathcal{L}}, \hookrightarrow \rangle$ in Figure 9 we construct the corresponding pmDAG $\mathfrak{G}^+ = \mathrm{T}_{\mathrm{root}(\mathfrak{G})}(\iota_{\boldsymbol{\mathcal{A}}}(\mathfrak{G}))$. Our assumption is that $\mathfrak{G}^+$ is a correspondent pmDAG. Here, the objective is to utilize the SN$^2$ algorithm to see whether the interventional distribution of $\mathcal{Y} \in \boldsymbol{\mathcal{V}}$ given $\mathcal{X} = 0$ ($\mathcal{X} \in \boldsymbol{\mathcal{V}}$) is identifiable from $\mathfrak{G}^+$ and the (biased) sample covariance matrix $\hat{\Sigma}_{\boldsymbol{\mathcal{V}}}$. This distribution is denoted using $\mathbb{P}(\mathcal{Y} | \mathrm{do}(\mathcal{X}))$ in the Pearlian causal inference regime. For the problem of identifiability, we used the same eight pmDAGs that have been experimented on in Xia et al. (2021, §4). The first four pmDAGs are called the back-door, front-door, M and napkin, where $\mathbb{P}(\mathcal{Y} | \mathrm{do}(\mathcal{X}))$ is known to be identifiable, when no assumption is held about the distribution of variables. The second four pmDAGs are called the bow, extended bow, instrumental variable (IV) and bad M. The interventional distribution $\mathbb{P}(\mathcal{Y} | \mathrm{do}(\mathcal{X}))$ is known not to be identifiable in these pmDAGs. With the Gaussian assumption, the interventional distribution of $\mathcal{Y} \in \boldsymbol{\mathcal{V}}$ given $\mathcal{X} = x$ ($x \in \Omega_{\mathcal{X}}$) is identifiable for the IV pmDAG, too, except for when $\bar{w}'_{\mathcal{Z}\mathcal{X}} = 0$ (see Figure 9 (e)). For the other cases, the identifiability is the same as when there is no Gaussian assumption.

As described below, Figure 9 measures the KL divergence of the induced distribution from the observed distribution and the KL divergence of the induced $\mathbb{P}_{\mathcal{Y} | \mathrm{do}(\mathcal{X}=0)}$ from the corresponding ground-truth distribution over $12{,}000$ iterations (epochs).

We first assigned random standard-normally distributed parameters to a pmDAG $\mathfrak{G}^+$ and induced the sample marginal covariance matrix $\hat{\Sigma}_{\boldsymbol{\mathcal{V}}}$ from that pmDAG. This pmDAG and the corresponding structural system act as the correspondent pmDAG and structural system, and we denote this structural system using $\langle \mathbb{P}, \Phi \rangle^+$. Then, for parameter-optimization of a pmDAG we constructed a pmDAG isomorphic to the correspondent pmDAG with standard-normally distributed random parameters. After that, we ran the SN$^2$ algorithm many times until the KL divergence $\mathrm{KL}\big(\Sigma \,\big\|\, \hat{\Sigma}\big)$ error converged within the $12{,}000$ epochs in ten of the experiments ($\mathrm{KL}\big(\Sigma \,\big\|\, \hat{\Sigma}\big) \leq 10^{-5}$). For training, we used the Adamax optimizer with learning rate equal to $10^{-3}$ (Kingma and Ba, 2014). We depicted the first, fifth, and ninth deciles of the error on the $\mathrm{Err}_{\mathrm{KL}}$ plot of each pmDAG.

For each pmDAG, we also measured the KL divergence of the induced interventional distribution $\mathbb{P}_{\mathcal{Y} | \mathrm{do}(\mathcal{X}=0)}$ of $\mathcal{Y}$ from the ground-truth distribution $\mathbb{P}^+_{\mathcal{Y} | \mathrm{do}(\mathcal{X}=0)}$. We define this divergence as:

$$\mathrm{KL}_{\mathcal{Y} | \mathrm{do}(\mathcal{X}=0)} := \mathrm{KL}\Big(\Pi\big(\langle \mathbb{P}, \Phi \rangle'_{\mathcal{X}=0}\big)_{\mathcal{Y}} \,\Big\|\, \Pi\big(\langle \mathbb{P}, \Phi \rangle^+_{\mathcal{X}=0}\big)_{\mathcal{Y}}\Big).$$

We used the same deciles for these values, too. As a result of Theorem 42, it is expected that the KL divergence of the induced interventional distribution of $\mathcal{Y}$ given $\mathcal{X} = 0$ from its ground-truth distribution converges for the identifiable cases. Interestingly, for tor the five identifiable cases this distance has converged and for the three non-identifiable cases this distance has not converged. This is indicative of the proposed meta-algorithm being capable of distinguishing the identifiable and non-identifiable cases.

This experiment shows that one can implement the meta-algorithm (Algorithm 3) to solve the identifiability problem in the Gaussian case. For a generic pmDAG $\mathfrak{G}$, whenever the interventional distribution of $\boldsymbol{\mathcal{E}}$ given $\boldsymbol{\mathcal{T}} = \boldsymbol{t}$ is not identifiable, the value of $\mathrm{KL}\big(\Pi\big(\langle\mathbb{P},\Phi\rangle^*_{\boldsymbol{\mathcal{T}}=\mathbf{0}}\big)_{\boldsymbol{\mathcal{E}}} \,\|\, \Pi\big(\langle\mathbb{P}',\Phi'\rangle^*_{\boldsymbol{\mathcal{T}}=\mathbf{0}}\big)_{\boldsymbol{\mathcal{E}}}\big)$ is expected to be positive for each two solutions that the $\mathrm{SN}^2$ algorithm returns.

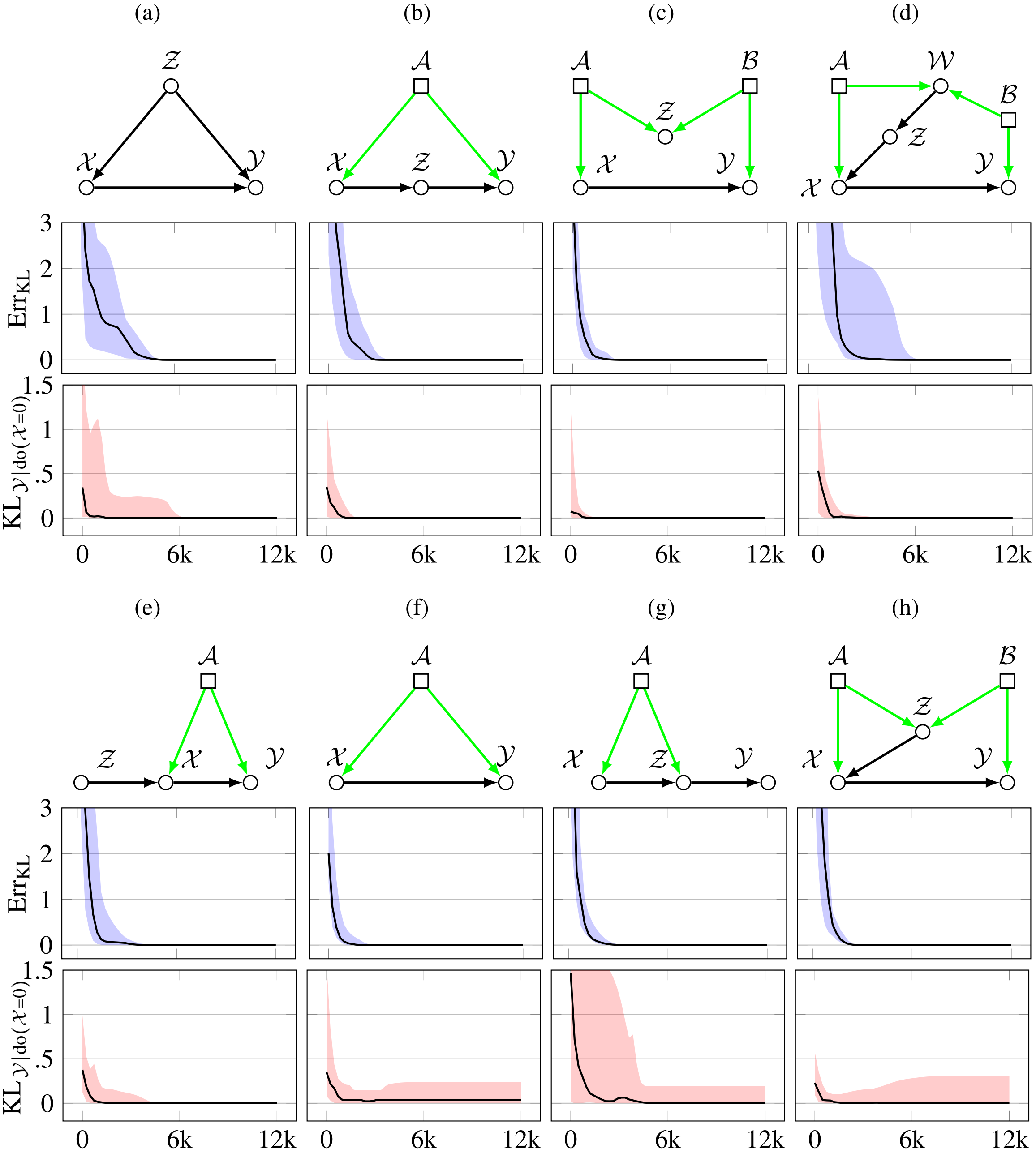

Figure 9: The identifiability problem for eight pmDAGs. (a), (b), (c), (d), (e): identifiable pmDAGs. (f), (g), (h): unidentifiable pmDAGs. On each row, top: the pmDAGs; middle: KL divergence of the induced marginal covariance matrix from the sample marginal covariance matrix; bottom: average effective distance w.r.t. $\mathcal{X}$ and $\mathcal{Y}$. Ten experiments have been done for each pmDAG over 12,000 iterations. The plot line is the median, while the colored area marks the first and ninth deciles. (a) back-door; (b) front-door; (c) M; (d) napkin; (e) IV; (f) bow; (g) extended bow; (h) bad M. pmDAGs are equivalent to those experimented on in Xia et al. (2021). The difference is the Gaussianity assumption in our experiments.

## 9 Concluding Notes

We showed that the existing graphical structures are not rich enough to capture the margins of Gaussian Bayesian networks. To address this problem, we generalized marginalized DAGs and proposed pre-marginalized DAGs (pmDAGs), which do not suffer from this problem. We restricted attention to the Gaussian assumption, and showed that these networks can be considered as the backbone of both deterministic structural systems (representing SCMs) and indeterministic structural systems (representing [causal] Bayesian networks). Furthermore, we showed that the optimal parameters for these structural systems can be obtained using methods that minimize the KL divergence of the sample and induced marginal distributions.

We showed that training a feed-forward neural network is dual to the problem of parameter-optimization of a pmDAG. Unlike the usual combination of neural networks and causal models, known collectively as neural causal models (NCMs), our algorithm works in the parameter space of the problem and considers the entire causal network as an object isomorphic to a feed-forward neural network. We proposed an algorithm, called the $\mathrm{SN}^2$ algorithm, that takes in a pmDAG and the covariance matrix of the observational margin of this pmDAG and returns an optimal set of parameters for the linear functions relating the variables in the pmDAG. The algorithm is implemented on CUDA and is compatible with PyTorch®.

Finally, we investigated the effect identifiability problem. We considered the Gaussian scenario and showed that, compatible with the general case, effect identifiability is not always possible. We proposed a condition for the identifiability of the distribution of a random vector in mutilated pmDAGs. We showed that this condition in combination with the $\mathrm{SN}^2$ algorithm hands in a meta-algorithm that can be used to test effect identifiability. This meta-algorithm is asymptotically complete. This straight-forward computational method for the identifiability problem is, however, not the only computational method. As part of our study on pmDAGs, it turned out that the uniqueness of the weights on the causal paths is a powerful sufficient condition for identifiability. We did not reflect this result in this work. However, exploring such computational methods further could expand this research direction.

The duality between causal graphs and neural networks leads us to interchanging solution for problems of these two domains. We showed that one of these problems, called the parameter optimization problem, can be solved by considering this duality. In terms of performance, our method might be superior to the existing NCMs as it transfers the whole causal DAG to a single neural network with no non-linearity. This is especially true as we can significantly optimize our neural network solution and provide optimized algorithms like the $\mathrm{SN}^2$ algorithm. On the other hand, our solution requires explicit representation of the marginal observed probability space. This might not always be possible when the number of dimensions is staggeringly high, for example, in the realms where causality is investigated in the image domain.

In order to show the duality between the two domains we introduced the notion of synchronization and the synchronized factorization property. Although not part of this article, we assert that the solutions that we presented for the Gaussian case are easily generalizable to the discrete distributions (with uniformly distributed latent space). (It is noteworthy that unlike the Gaussian settings, discrete distributions are compatible with mDAGs.) Therefore, this article acts as also a prelude for the more general scenarios. Of course, such a generalization is a potential subject for our future work.

## Acknowledgments and Disclosure of Funding

We would like to express our gratitude to Professor Philip Dawid (Emeritus Professor at Cambridge University) for his moral support during the composition of this manuscript. We appreciate Dr. Kayvan Sadeghi (University College London) for his constructive discussions, which improved this manuscript effectively. The author extends sincere thanks to the anonymous reviewers for their meticulous analysis and insightful feedback. The author assumes complete accountability for any potential mistakes and shortcomings in this article.

The author received no compensation related to the development of this manuscript. The author declares that he has no conflict of interest.

## Appendix A. Summary of Notation

| Notation | Description | First appearance |
|---|---|---|
| $\iota_{\mathcal{U}}(\mathfrak{G})$ | augmentation of pmDAG $\mathfrak{G}$ w.r.t. the random variable $\mathcal{U}$ | Def. 6 (Page 7) |
| $\text{aux}_{\mathfrak{G},\mathfrak{G}'}(\mathcal{V})$ | the extra parent(s) of $\mathcal{V}$ in lvDAG $\mathfrak{G}'$ compared to lvDAG $\mathfrak{G}$ | Page 7 |
| $\langle \mathbb{P}, \Phi \rangle$ | deterministic structural system | Def. 7 (Page 8) |
| $\mathfrak{D}(\mathfrak{G})$ | deterministic structural system model (DSM) of pmDAG $\mathfrak{G}$ | Def. 7 (Page 8) |
| $\{\kappa_{\mathcal{A}}\}_{\mathcal{A} \in \boldsymbol{\mathcal{A}}}$ | indeterministic structural system | Def. 8 (Page 9) |
| $\mathfrak{d}(\mathfrak{G})$ | indeterministic structural system model (ISM) of pmDAG $\mathfrak{G}$ | Def. 8 (Page 9) |
| $\mathfrak{Q}(\mathfrak{G})$ | deterministic probabilistic model (DPM) of pmDAG $\mathfrak{G}$ | Def. 11 (Page 9) |
| $\mathfrak{q}(\mathfrak{G})$ | indeterministic probabilistic model (IPM) of pmDAG $\mathfrak{G}$ | Def. 11 (Page 9) |
| $\Pi(\langle \mathbb{P}, \Phi \rangle)$ | induced probability measure of the deterministic structural system $\langle \mathbb{P}, \Phi \rangle$ | Page 10 |
| $\Pi(\{\kappa_{\mathcal{A}}\}_{\mathcal{A} \in \boldsymbol{\mathcal{A}}})$ | induced probability measure of the indeterministic structural system $\{\kappa_{\mathcal{A}}\}_{\mathcal{A} \in \boldsymbol{\mathcal{A}}}$ | Page 10 |
| $\text{I}_{\mathfrak{G}}(\langle \mathbb{P}, \Phi \rangle)$ | indetermination of the deterministic structural system $\langle \mathbb{P}, \Phi \rangle$ | Def. 12 (Page 10) |
| $\mathfrak{Q}^{\text{G}(0)}(\mathfrak{G})$ | Gaussian DPM of pmDAG $\mathfrak{G}$ | Page 10 |
| $\mathfrak{D}^{\text{G}(0)}(\mathfrak{G})$ | Gaussian DSM of pmDAG $\mathfrak{G}$ | Page 10 |
| $\mathfrak{D}^{\text{lG}(0)}(\mathfrak{G})$ | linear Gaussian DSM of pmDAG $\mathfrak{G}$ | Page 11 |
| $\mathfrak{q}^{\text{G}(0)}(\mathfrak{G})$ | Gaussian IPM of pmDAG $\mathfrak{G}$ | Page 11 |
| $\mathfrak{d}^{\text{G}(0)}(\mathfrak{G})$ | Gaussian ISM of pmDAG $\mathfrak{G}$ | Page 11 |
| $\text{T}_{\mathcal{L}}(\mathfrak{G})$ | deterministic exogenization of lvDAG $\mathfrak{G}$ w.r.t. variable $\mathcal{L}$ | Def. 13 (Page 12) |
| $\tau_{\mathcal{L}}(\mathfrak{G})$ | indeterministic exogenization of lvDAG $\mathfrak{G}$ w.r.t. variable $\mathcal{L}$ | Def. 13 (Page 12) |
| $\delta_{\mathcal{L}}(\mathfrak{G})$ | coalescence of lvDAG $\mathfrak{G}$ w.r.t. variable $\mathcal{L}$ | Def. 14 (Page 12) |
| $\mathfrak{D}^{\text{G}(0,1)}(\mathfrak{G})$ | Gaussian DSM of pmDAG $\mathfrak{G}$ with standardized latent variables | Page 17 |
| $\Xi(\mathfrak{G})$ | synchronization of pmDAG $\mathfrak{G}$ | Def. 22 (Page 18) |
| $\Xi_{\mathfrak{G}}(\{\kappa_{\mathcal{A}}\}_{\mathcal{A} \in \boldsymbol{\mathcal{A}}})$ | indeterministic structural system of synchronization $\Xi(\mathfrak{G})$ corresponding to $\{\kappa_{\mathcal{A}}\}_{\mathcal{A} \in \boldsymbol{\mathcal{A}}}$ | Def. 25 (Page 20) |
| $\Psi_{\mathfrak{G}}(\langle \mathbb{P}, \Phi \rangle)$ | projection of $\langle \mathbb{P}, \Phi \rangle$ onto lvDAG $\mathfrak{G}$ | Lem. 45 (Page 41) |

## Appendix B. Further Notes on pmDAGs with Gaussian Distributions

### B.1 General Properties of Gaussian pmDAGs

In this sub-section, we provide some basic statements regarding the Gaussian pmDAGs. These statements were claimed trough-out the article but were usually too trivial to consume the main body of the work.

Theorem 44 proves that for a structural system to induce a jointly Gaussian distribution, the joint probability distribution of the root nodes must be Gaussian and the functions associated to the non-root nodes must be linear transformations. Lemma 43 is a preliminary for that theorem.

**Lemma 43** *Let $\mathcal{X}$ and $\mathcal{Y}$ be two random variables and $\mathbb{P}_{\mathcal{X}}$ a probability distribution for $\mathcal{X}$. Also, let $f_1, f_2 : \Omega_{\mathcal{X}} \to \Omega_{\mathcal{Y}}$ be two measurable functions. If $\mathbb{P}_{\mathcal{X},\mathcal{Y}}(\mathcal{X} \in X \wedge f_1(X) \in Y) = \mathbb{P}_{\mathcal{X},\mathcal{Y}}(\mathcal{X} \in X \wedge f_2(X) \in Y)$ for all $X \in \sigma\Omega_{\mathcal{X}}, Y \in \sigma\Omega_{\mathcal{Y}}$, then $f_1 = f_2$ are equal up to $\mathbb{P}_{\mathcal{X}}$-measure zero.*

**Proof** $\mathbb{P}_{\mathcal{X},\mathcal{Y}}(\mathcal{X} \in X \wedge f_1(X) \in Y) = \mathbb{P}_{\mathcal{X},\mathcal{Y}}(\mathcal{X} \in X \wedge f_2(X) \in Y)$ means that we have $\mathbb{P}_{\mathcal{X}}\big(X \cap f_1^{-1}(Y)\big) = \mathbb{P}_{\mathcal{X}}\big(X \cap f_2^{-1}(Y)\big)$. We assume $\mathbb{P}_{\mathcal{X}}\big(f_1^{-1}(Y) \setminus f_2^{-1}(Y)\big) > 0$. Now, we choose $X = f_1^{-1}(Y) \setminus f_2^{-1}(Y)$. This implies $\mathbb{P}_{\mathcal{X}}(X) = \mathbb{P}_{\mathcal{X}}(\varnothing) = 0$, which violates the assumption. Therefore, $\mathbb{P}_{\mathcal{X}}\big(f_1^{-1}(Y) \setminus f_2^{-1}(Y)\big) = 0$, and $f_1 = f_2$ are equal up to $\mathbb{P}_{\mathcal{X}}$-measure zero. ■

**Theorem 44** *For any lvDAG $\mathfrak{G} = \langle \boldsymbol{\mathcal{A}} \equiv \boldsymbol{\mathcal{V}} \dot{\cup} \boldsymbol{\mathcal{L}}, \hookrightarrow \rangle$, we have $\mathfrak{D}^{\mathrm{lG}(0)}(\mathfrak{G}) = \mathfrak{D}^{\mathrm{G}(0)}(\mathfrak{G})$.*

**Proof** $\mathfrak{D}^{\mathrm{lG}(0)}(\mathfrak{G}) \subseteq \mathfrak{D}^{\mathrm{G}(0)}(\mathfrak{G})$ is trivial because the linear transformation of a jointly Gaussian distribution is jointly Gaussian. We only need to prove $\mathfrak{D}^{\mathrm{lG}(0)}(\mathfrak{G}) \supseteq \mathfrak{D}^{\mathrm{G}(0)}(\mathfrak{G})$: Let $\langle \mathbb{P}, \Phi \rangle \in \mathfrak{D}^{\mathrm{G}(0)}(\mathfrak{G})$. It is obvious that $\mathbb{P} \in \mathfrak{P}^{\mathrm{G}(0)}_{\mathrm{root}(\mathfrak{G})}$ because every margin of the induced Gaussian distribution is Gaussian and $\mathbb{P}$ is one of those margins. According to Lemma 43 in Appendix C, the functions $\Phi$ are equal to linear transformations up to $\mathbb{P}_{\mathrm{root}(\mathfrak{G})}$-measure zero. With the Gaussian density assumption, we have that $\Phi$ are linear transformations. ■

In the following, we prove Equation 4. We first consider the problem in the more general settings and introduce the operation called *projection*. Then, we specialize the result to the Gaussian case, and proceed to proving Equation 4.

**Lemma 45** *Let $\mathfrak{G} = \langle \boldsymbol{\mathcal{A}} \equiv \boldsymbol{\mathcal{V}} \dot{\cup} \boldsymbol{\mathcal{L}}, \hookrightarrow \rangle$ be an lvDAG and $\mathfrak{G}' = \iota_{\boldsymbol{\mathcal{A}}}(\mathfrak{G})$ its augmentation w.r.t. $\boldsymbol{\mathcal{A}}$. We represent $\mathrm{aux}_{\mathfrak{G},\mathfrak{G}'}(\boldsymbol{\mathcal{A}})$ using $\boldsymbol{\mathcal{E}}_{\boldsymbol{\mathcal{A}}}$. We define the function $\Psi_{\mathfrak{G}} : \mathfrak{D}(\mathfrak{G}') \to \mathfrak{d}(\mathfrak{G})$ as the one that takes in a structural system $\langle \mathbb{P}, \Phi \rangle$ of $\mathfrak{G}'$ and returns an indeterministic structural system $\{\kappa_{\mathcal{A}}\}_{\mathcal{A} \in \boldsymbol{\mathcal{A}}}$ of $\mathfrak{G}$ defined as follows:*

$$\kappa_{\mathcal{A}}(X | \boldsymbol{y}) := \mathbb{P}_{\mathcal{E}_{\mathcal{A}}}\Big(\Big\{e \in \Omega_{\mathcal{E}_{\mathcal{A}}} \Big| (\boldsymbol{y}, e) \in \Phi_{\mathcal{A}}^{-1}(X)\Big\}\Big)$$

*for every $X \in \sigma\Omega_{\mathcal{A}}, \boldsymbol{y} \in \Omega_{\mathrm{pa}_{\mathfrak{G}}(\mathcal{A})}$ for all $\mathcal{A} \in \boldsymbol{\mathcal{A}}$. These are trivially well-defined Markov kernels. We call $\Psi_{\mathfrak{G}}(\langle \mathbb{P}, \Phi \rangle)$ the* projection *of $\langle \mathbb{P}, \Phi \rangle$ onto $\mathfrak{G}$. Then, $\langle \mathbb{P}, \Phi \rangle$ and its projection induce the same probability distribution over $\boldsymbol{\mathcal{A}}$. That is,*

$$\Pi(\langle \mathbb{P}, \Phi \rangle)_{\boldsymbol{\mathcal{A}}} = \Pi(\Psi_{\mathfrak{G}}(\langle \mathbb{P}, \Phi \rangle)).$$

**Proof** We assume that $\mathfrak{G}' = \langle \boldsymbol{\mathcal{A}}' \equiv \boldsymbol{\mathcal{V}} \dot{\cup} \boldsymbol{\mathcal{L}}', \hookrightarrow \rangle$. We denote the probability distribution induced by the structural system $\langle \mathbb{P}, \Phi \rangle$ using $\mathbb{P}_{\boldsymbol{\mathcal{A}}'}$. We denote the indetermination of $\langle \mathbb{P}, \Phi \rangle$ using $\{\kappa_{\mathcal{A}}\}_{\mathcal{A} \in \boldsymbol{\mathcal{A}}'}$, the projection of $\langle \mathbb{P}, \Phi \rangle$ onto $\mathfrak{G}$ using $\{\kappa'_{\mathcal{A}}\}_{\mathcal{A} \in \boldsymbol{\mathcal{A}}}$, and the induced probability distribution of the projection using $\mathbb{P}'_{\boldsymbol{\mathcal{A}}}$. The goal is to show $\mathbb{P}_{\boldsymbol{\mathcal{A}}} = \mathbb{P}'_{\boldsymbol{\mathcal{A}}}$. Fix $\boldsymbol{x} \in \Omega_{\boldsymbol{\mathcal{A}}}$. We have:

$$
\begin{aligned}
\mathbb{P}_{\boldsymbol{\mathcal{A}}}\Big(\prod_{\mathcal{A}\in\boldsymbol{\mathcal{A}}} \mathrm{d}x_{\mathcal{A}}\Big) &= \mathbb{P}_{\boldsymbol{\mathcal{A}}'}\Big(\prod_{\mathcal{A}\in\boldsymbol{\mathcal{A}}} \mathrm{d}x_{\mathcal{A}} \times \Omega_{\mathcal{E}_{\mathcal{A}}}\Big) && \text{(by definition)}\\
&= \underset{\Omega_{\mathcal{E}_{\mathcal{A}}}|\mathcal{A}\in\boldsymbol{\mathcal{A}}}{\int\cdots\int} \mathbb{P}_{\boldsymbol{\mathcal{A}}'}\Big(\prod_{\mathcal{A}\in\boldsymbol{\mathcal{A}}} \mathrm{d}x_{\mathcal{A}} \times \mathrm{d}e_{\mathcal{E}_{\mathcal{A}}}\Big)\\
&= \underset{\Omega_{\mathcal{E}_{\mathcal{A}}}|\mathcal{A}\in\boldsymbol{\mathcal{A}}}{\int\cdots\int} \prod_{\mathcal{A}\in\boldsymbol{\mathcal{A}}} \kappa_{\mathcal{A}}\big(\mathrm{d}x_{\mathcal{A}}\,|\,x_{\mathrm{pa}_{\mathfrak{G}}(\mathcal{A})}, e_{\mathcal{E}_{\mathcal{A}}}\big)\,\kappa_{\mathcal{E}_{\mathcal{A}}}(\mathrm{d}e_{\mathcal{E}_{\mathcal{A}}}\,|\,\{0\}) && \text{(Equation 2)}\\
&= \prod_{\mathcal{A}\in\boldsymbol{\mathcal{A}}} \int_{\Omega_{\mathcal{E}_{\mathcal{A}}}} \kappa_{\mathcal{A}}\big(\mathrm{d}x_{\mathcal{A}}\,|\,x_{\mathrm{pa}_{\mathfrak{G}}(\mathcal{A})}, e_{\mathcal{E}_{\mathcal{A}}}\big)\,\kappa_{\mathcal{E}_{\mathcal{A}}}(\mathrm{d}e_{\mathcal{E}_{\mathcal{A}}}\,|\,\{0\})\\
&= \prod_{\mathcal{A}\in\boldsymbol{\mathcal{A}}} \int_{\Omega_{\mathcal{E}_{\mathcal{A}}}} \mathbb{1}_{\mathrm{d}x_{\mathcal{A}}}\big(\Phi_{\mathcal{A}}\big(x_{\mathrm{pa}_{\mathfrak{G}}(\mathcal{A})}, e_{\mathcal{E}_{\mathcal{A}}}\big)\big)\,\mathbb{P}_{\mathcal{E}_{\mathcal{A}}}(\mathrm{d}e_{\mathcal{E}_{\mathcal{A}}}) && \text{(Equation 3)}\\
&= \prod_{\mathcal{A}\in\boldsymbol{\mathcal{A}}} \mathbb{P}_{\mathcal{E}_{\mathcal{A}}}\big(\big\{e \in \Omega_{\mathcal{E}_{\mathcal{A}}}\,\big|\,\big(x_{\mathrm{pa}_{\mathfrak{G}}(\mathcal{A})}, e\big) \in \Phi_{\mathcal{A}}^{-1}(\mathrm{d}x_{\mathcal{A}})\big\}\big)\\
&= \prod_{\mathcal{A}\in\boldsymbol{\mathcal{A}}} \kappa'\big(\mathrm{d}x_{\mathcal{A}}\,|\,x_{\mathrm{pa}_{\mathfrak{G}}(\mathcal{A})}\big) && \text{(by definition)}\\
&= \mathbb{P}'_{\boldsymbol{\mathcal{A}}}\Big(\prod_{\mathcal{A}\in\boldsymbol{\mathcal{A}}} \mathrm{d}x_{\mathcal{A}}\Big) && \blacksquare
\end{aligned}
$$

The projection function that we defined above can be specialized for the Gaussian pmDAGs. Let $\mathfrak{G} = \langle \boldsymbol{\mathcal{A}} \equiv \boldsymbol{\mathcal{V}} \,\dot\cup\, \boldsymbol{\mathcal{L}}, \hookrightarrow \rangle$ be an lvDAG and $\mathfrak{G}' = \iota_{\boldsymbol{\mathcal{A}}}(\mathfrak{G})$ its augmentation w.r.t. $\boldsymbol{\mathcal{A}}$. Let $\varphi$ denote the standard Gaussian cumulative distribution function (CDF). The projection of $\langle \mathbb{P}, \Phi \rangle \in \mathfrak{D}^{\mathrm{G}(0)}(\mathfrak{G}')$ onto $\mathfrak{G}$ is $\{\kappa_{\mathcal{A}}\}_{\mathcal{A}\in\boldsymbol{\mathcal{A}}} \in \mathfrak{d}^{\mathrm{G}(0)}(\mathfrak{G})$ iff.

$$
\kappa_{\mathcal{A}}((-\infty, x]\,|\,\boldsymbol{y}) := \varphi^*(x; \Phi_{\mathcal{A}}(\boldsymbol{y}, 0), \Phi_{\mathcal{A}}(\boldsymbol{0}, \bar{\sigma}_{\mathcal{E}_{\mathcal{A}}})), \tag{33}
$$

for every $X \in \Omega_{\mathcal{A}}$, $\boldsymbol{y} \in \Omega_{\mathrm{pa}_{\mathfrak{G}}(\mathcal{A})}$ for all $\mathcal{A} \in \boldsymbol{\mathcal{A}}$, and where $\mathcal{E}_{\mathcal{A}} \equiv \mathrm{aux}_{\mathfrak{G},\mathfrak{G}'}(\mathcal{A})$ and

$$
\varphi^*(x; \mu, \sigma) := \begin{cases} \varphi\left(\frac{x-\mu}{\sigma}\right) & \sigma \neq 0, \\ \mathbb{1}_{(-\infty, x]}(\mu) & \text{otherwise.} \end{cases} \tag{34}
$$

**Lemma 46** *Let $\mathfrak{G} = \langle \boldsymbol{\mathcal{A}} \equiv \boldsymbol{\mathcal{V}} \,\dot\cup\, \boldsymbol{\mathcal{L}}, \hookrightarrow \rangle$ be an lvDAG. We define:*

$$
\mathfrak{d}^{\mathrm{lG}(0)}(\mathfrak{G}) = \{\{\kappa_{\mathcal{A}}\}_{\mathcal{A}\in\boldsymbol{\mathcal{A}}} \in \mathfrak{d}(\mathfrak{G})\,|\,\kappa_{\mathcal{A}}((-\infty, x]\,|\,\boldsymbol{y}) = \varphi^*(x; \Phi_{\mathcal{A}}(\boldsymbol{y}, 0), \Phi_{\mathcal{A}}(\boldsymbol{0}, 1)) \text{ where } \Phi \text{ are linear}\},
$$

*where $\varphi^*$ is defined in Equation 34. Then, we have $\mathfrak{d}^{\mathrm{G}(0)}(\mathfrak{G}) = \mathfrak{d}^{\mathrm{lG}(0)}(\mathfrak{G})$.*

**Proof** We use the fact that the induced probability $\mathbb{P}_{\boldsymbol{\mathcal{A}}}$ is uniquely decomposed to regular conditional probabilities $\mathbb{P}_{\mathcal{A}|\mathrm{pa}_{\mathfrak{G}}(\mathcal{A})}$ up to $\mathbb{P}_{\boldsymbol{\mathcal{A}}}$-measure zero and those probabilities uniquely determine $\mathbb{P}_{\boldsymbol{\mathcal{A}}}$ (Gelman and Speed, 1993). These conditional probabilities of jointly Gaussian distributions are defined as:

$$
\mathcal{A}\,|\,\mathrm{pa}_{\mathfrak{G}}(\mathcal{A}) = \boldsymbol{y} \sim \mathrm{Normal}\big(\mu = \Phi_{\mathcal{A}}(\boldsymbol{y}, 0), \sigma^2 = \Phi_{\mathcal{A}}^2(\boldsymbol{0}, 1)\big).
$$

We let the Markov kernels $\kappa_{\mathcal{A}}$ be equal to these conditional probabilities. As before, given the Gaussian density, uniqueness holds. ■

**Lemma 47** *Let $\mathfrak{G} = \langle \boldsymbol{\mathcal{A}} \equiv \boldsymbol{\mathcal{V}} \dot\cup \boldsymbol{\mathcal{L}}, \hookrightarrow \rangle$ be an lvDAG and $\mathfrak{G}' = \iota_{\boldsymbol{\mathcal{A}}}(\mathfrak{G})$ its augmentation w.r.t. $\boldsymbol{\mathcal{A}}$. Then, $\Psi_{\mathfrak{G}}\big(\mathfrak{D}^{\mathrm{G}(0)}(\mathfrak{G}')\big) = \mathfrak{d}^{\mathrm{G}(0)}(\mathfrak{G})$.*

**Proof** We let $\mathfrak{G}' = \langle \boldsymbol{\mathcal{A}}' \equiv \boldsymbol{\mathcal{V}} \dot\cup \boldsymbol{\mathcal{L}}', \hookrightarrow' \rangle$. $\subseteq$:

$$\begin{aligned}
& & \langle \mathbb{P}, \Phi \rangle \in \mathfrak{D}^{\mathrm{G}(0)}\big(\mathfrak{G}'\big) & \\
\Longrightarrow & & \Pi(\langle \mathbb{P}, \Phi \rangle) \in \mathfrak{P}^{\mathrm{G}(0)}_{\boldsymbol{\mathcal{A}}'} & \qquad \text{(by definition)} \\
\Longrightarrow & & \Pi(\langle \mathbb{P}, \Phi \rangle)_{\boldsymbol{\mathcal{A}}} \in \mathfrak{P}^{\mathrm{G}(0)}_{\boldsymbol{\mathcal{A}}} & \\
\Longrightarrow & & \Pi(\Psi_{\mathfrak{G}}(\langle \mathbb{P}, \Phi \rangle)) \in \mathfrak{P}^{\mathrm{G}(0)}_{\boldsymbol{\mathcal{A}}} & \qquad \text{(Lemma 45)} \\
\Longrightarrow & & \Psi_{\mathfrak{G}}(\langle \mathbb{P}, \Phi \rangle) \in \mathfrak{d}^{\mathrm{G}(0)}(\mathfrak{G}) & \qquad \text{(by definition)}
\end{aligned}$$

$\supseteq$:

$$\begin{aligned}
& & \{\Phi_{\mathcal{A}}\}_{\mathcal{A}\in\boldsymbol{\mathcal{A}}} \in \mathfrak{d}^{\mathrm{G}(0)}(\mathfrak{G}) & \\
\Longrightarrow & & \{\Phi_{\mathcal{A}}\}_{\mathcal{A}\in\boldsymbol{\mathcal{A}}} \in \mathfrak{d}^{\mathrm{IG}(0)}(\mathfrak{G}) & \qquad \text{(Lemma 46)}
\end{aligned}$$

But, based on Equation 34, there is a $\langle \mathbb{P}, \Phi \rangle \in \mathfrak{D}^{\mathrm{G}(0)}(\mathfrak{G}')$ such that $\{\Phi_{\mathcal{A}}\}_{\mathcal{A}\in\boldsymbol{\mathcal{A}}} \in \Psi_{\mathfrak{G}}(\langle \mathbb{P}, \Phi \rangle)$. In other words, $\{\Phi_{\mathcal{A}}\}_{\mathcal{A}\in\boldsymbol{\mathcal{A}}} \in \Psi_{\mathfrak{G}}\big(\mathfrak{D}^{\mathrm{G}(0)}(\mathfrak{G}')\big)$. ■

**Theorem 48** *Let $\mathfrak{G} = \langle \boldsymbol{\mathcal{A}} \equiv \boldsymbol{\mathcal{V}} \dot\cup \boldsymbol{\mathcal{L}}, \hookrightarrow \rangle$ be an lvDAG and $\mathfrak{G}' = \langle \boldsymbol{\mathcal{A}}' \equiv \boldsymbol{\mathcal{V}} \dot\cup \boldsymbol{\mathcal{L}}', \hookrightarrow' \rangle$ its augmentation w.r.t. $\boldsymbol{\mathcal{A}}$. Then, the Gaussian IPM of $\mathfrak{G}$ and the Gaussian DPM of $\mathfrak{G}'$ over $\boldsymbol{\mathcal{A}}$ are the same. That is,*

$$\mathfrak{q}^{\mathrm{G}(0)}(\mathfrak{G}) = \mathfrak{Q}^{\mathrm{G}(0)}_{\boldsymbol{\mathcal{A}}}\big(\mathfrak{G}'\big).$$

**Proof** $\supseteq$: Let $\langle \mathbb{P}, \Phi \rangle$ be a structural system of $\mathfrak{G}'$ that induces $\mathbb{P}_{\boldsymbol{\mathcal{A}}'}$. According to Lemma 47, there is an indeterministic structural system $\{\kappa_{\mathcal{A}}\}_{\mathcal{A}\in\boldsymbol{\mathcal{A}}} = \Psi_{\mathfrak{G}}(\langle \mathbb{P}, \Phi \rangle)$ that induces $\mathbb{P}_{\boldsymbol{\mathcal{A}}}$. Therefore, $\mathbb{P}_{\boldsymbol{\mathcal{A}}}$ is in $\mathfrak{q}^{\mathrm{G}(0)}(\mathfrak{G})$.

$\subseteq$: It is proven similarly. If $\{\kappa_{\mathcal{A}}\}_{\mathcal{A}\in\boldsymbol{\mathcal{A}}}$ is an indeterministic structural system of $\mathfrak{G}$, there is a $\langle \mathbb{P}, \Phi \rangle \in \Psi^{-1}_{\mathfrak{G}}(\{\kappa_{\mathcal{A}}\}_{\mathcal{A}\in\boldsymbol{\mathcal{A}}})$ that induces $\mathbb{P}_{\boldsymbol{\mathcal{A}}'}$. ■

### B.2 Standardization of Root Nodes is without Loss of Generality

In this section, we explain why restricting to the subset of Gaussian DSM, $\mathfrak{D}^{\mathrm{G}(0,1)}(\mathfrak{G})$, is without loss of generality. We remember from §2.3 that every structural system $\langle \mathbb{P}, \Phi \rangle \in \mathfrak{D}^{\mathrm{G}(0)}(\mathfrak{G})$ of a pmDAG $\mathfrak{G}$ is specified by the variance $\bar{\sigma}^2_{\mathcal{A}}$ of each variable $\mathcal{A} \in \mathrm{root}(\mathfrak{G})$ and a set of vectors $\bar{\boldsymbol{w}} = \{\bar{w}_{\mathcal{V}}\}_{\mathcal{V}\in\mathrm{nroot}(\mathfrak{G})}$ for the variables $\mathcal{V} \in \mathrm{nroot}(\mathfrak{G})$. Assume that the pmDAG $\mathfrak{G} = \langle \boldsymbol{\mathcal{A}} \equiv \boldsymbol{\mathcal{V}} \dot\cup \boldsymbol{\mathcal{L}}, \hookrightarrow \rangle$ and its structural system $\langle \mathbb{P}, \Phi \rangle$ are given. We consider the following procedure:

(a) Construct the lvDAG $\mathfrak{G}' = \langle \boldsymbol{\mathcal{A}}' \equiv \boldsymbol{\mathcal{V}} \dot\cup (\boldsymbol{\mathcal{L}} \dot\cup \boldsymbol{\mathcal{S}}), \hookrightarrow' \rangle$ such that $\mathfrak{G}' = \iota_{\mathrm{root}(\mathfrak{G})}(\mathfrak{G})$.

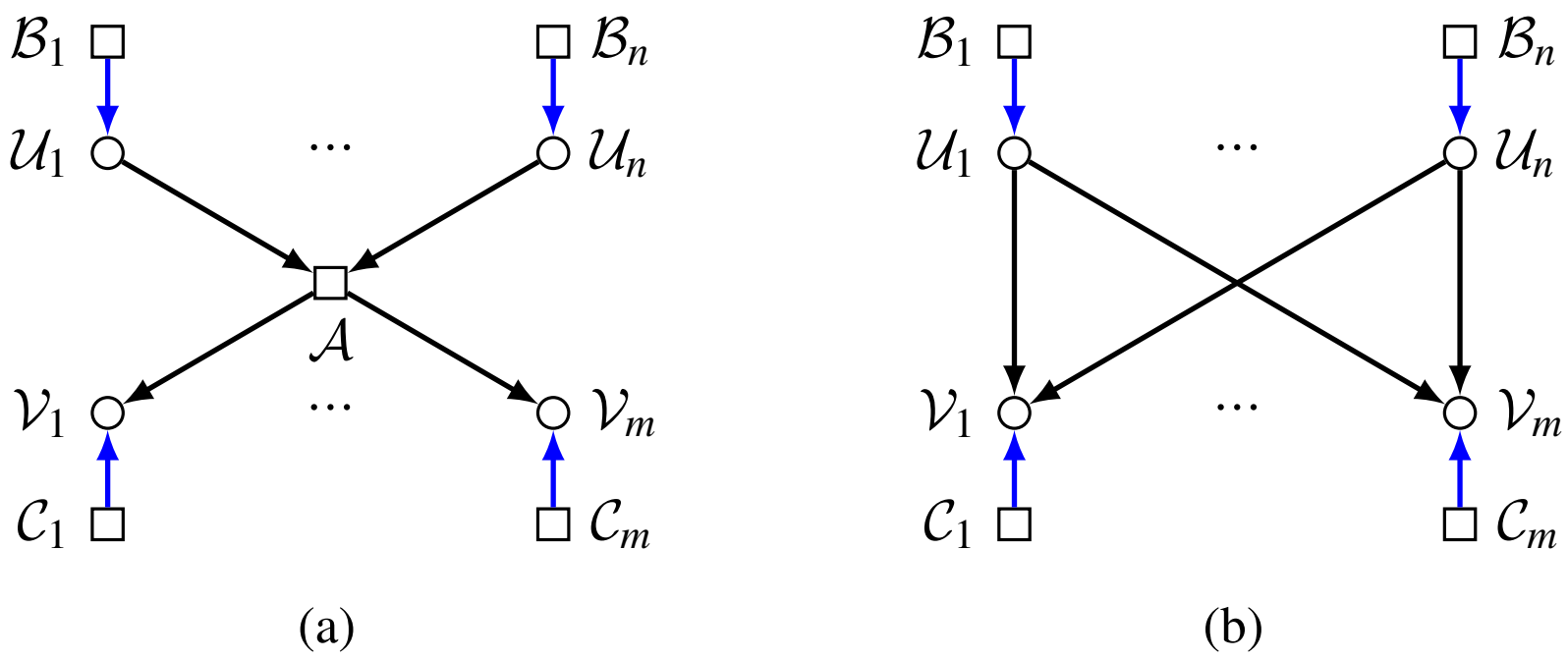

Figure 10: The lvDAGs that we us in the Proof of Theorem 15. (a) an lvDAG $\mathfrak{G}$ before deterministic exogenization, (b) the same lvDAG after deterministic exogenization w.r.t. $\mathcal{A}$, i.e. $\mathrm{T}_{\mathcal{A}}(\mathfrak{G})$.

(b) Construct the pmDAG $\mathfrak{G}'' = \langle \boldsymbol{\mathcal{A}}'' \equiv \boldsymbol{\mathcal{V}} \dot{\cup} \boldsymbol{\mathcal{S}}, \hookrightarrow'' \rangle$ such that $\mathfrak{G}'' = \mathrm{T}_{\mathrm{root}(\mathfrak{G})}(\mathfrak{G}')$.

For any $\mathcal{A} \in \mathrm{root}(\mathfrak{G})$, we let $\mathcal{P}_{\mathcal{A}} \equiv \mathrm{pa}_{\mathfrak{G}'}(\mathcal{A})$. If for all $\mathcal{A} \in \mathrm{root}(\mathfrak{G})$ we have $\bar{\sigma}'^2_{\mathcal{P}_{\mathcal{A}}} = 1$ and $\bar{w}'_{\mathcal{P}_{\mathcal{A}}\mathcal{A}} = \bar{\sigma}_{\mathcal{A}}$, and for $\mathcal{J} \in \mathrm{nroot}(\mathfrak{G})$ and $\mathcal{I} \in \mathrm{pa}_{\mathfrak{G}}(\mathcal{J})$ we let $\bar{w}'_{\mathcal{I}\mathcal{J}} = \bar{w}_{\mathcal{I}\mathcal{J}}$, we will get a structural system $\langle \mathbb{P}', \Phi' \rangle$ of $\mathfrak{G}'$ such that $\mathbb{P}'_{\mathrm{root}(\mathfrak{G})} = \mathbb{P}_{\mathrm{root}(\mathfrak{G})}$ and $\Phi'_{\mathrm{nroot}(\mathfrak{G})} = \Phi_{\mathrm{nroot}(\mathfrak{G})}$. Indeed, both lvDAGs induce the same probability distribution over $\boldsymbol{\mathcal{A}}$. The pmDAG $\mathfrak{G}''$ has a nice property. It has a structural system $\langle \mathbb{P}'', \Phi'' \rangle \in \mathfrak{D}^{\mathrm{G}(0)}(\mathfrak{G}')$ that induces the same probability distribution as $\langle \mathbb{P}', \Phi' \rangle$ (over $\boldsymbol{\mathcal{V}}$), and it is isomorphic to $\mathfrak{G}$ (it has the same graphical structure as $\mathfrak{G}$). This constitutes the basis for the claim that without loss of generality we can assume the variances of the root nodes are always equal to 1. We formalize this wording in the following theorem, which can be acknowledged without a formal proof.

**Theorem 49** *Let* $\mathfrak{G} = \langle \boldsymbol{\mathcal{A}} \equiv \boldsymbol{\mathcal{V}} \dot{\cup} \boldsymbol{\mathcal{L}}, \hookrightarrow \rangle$ *be a pmDAG and* $\langle \mathbb{P}, \Phi \rangle, \langle \mathbb{P}', \Phi' \rangle \in \mathfrak{D}^{\mathrm{G}(0)}(\mathfrak{G})$ *be two structural systems of* $\mathfrak{G}$ *such that* $\bar{\sigma}'^2_{\mathcal{A}} = 1$ *for each variable* $\mathcal{A} \in \mathrm{root}(\mathfrak{G})$ *and*

$$\bar{w}'_{\mathcal{I}\mathcal{J}} = \begin{cases} \bar{w}_{\mathcal{I}\mathcal{J}}\ \bar{\sigma}_{\mathcal{J}} & \mathcal{I} \in \mathrm{root}(\mathfrak{G}), \\ \bar{w}_{\mathcal{I}\mathcal{J}} & \mathcal{I} \in \mathrm{nroot}(\mathfrak{G}), \end{cases}$$

*for all* $\mathcal{I} \in \boldsymbol{\mathcal{A}}$, $\mathcal{J} \in \mathrm{ch}_{\mathfrak{G}}(\mathcal{I})$ *, then the following holds:*

$$\left\langle \mathbb{P}', \Phi' \right\rangle \in \mathrm{MLE}_{\mathfrak{G},V}\left(\mathfrak{D}^{\mathrm{G}(0,1)}(\mathfrak{G})\right) \qquad \Longleftrightarrow \qquad \left\langle \mathbb{P}, \Phi \right\rangle \in \mathrm{MLE}_{\mathfrak{G},V}\left(\mathfrak{D}^{\mathrm{G}(0)}(\mathfrak{G})\right).$$

## Appendix C. Lemmata and Theorems

**Proof of Theorem 15** We define the *deterministic exogenization of a structural system* $\langle \mathbb{P}, \Phi \rangle \in \mathfrak{D}(\mathfrak{G})$ *w.r.t. the variable* $\mathcal{L} \in \boldsymbol{\mathcal{L}} \cap \mathrm{nroot}(\mathfrak{G})$ as the function $\mathrm{T}_{\mathfrak{G},\mathcal{L}} : \mathfrak{D}(\mathfrak{G}) \to \mathfrak{D}(\mathrm{T}_{\mathcal{L}}(\mathfrak{G}))$ that returns the structural system $\langle \mathbb{P}', \Phi' \rangle \in \mathfrak{D}(\mathrm{T}_{\mathcal{L}}(\mathfrak{G}))$ defined as follows:

$$\mathbb{P}' := \mathbb{P}, \qquad \Phi'_{\mathcal{A}} := \begin{cases} \Phi_{\mathcal{L}} \circ \left( \Phi_{\mathcal{A}} \times \mathrm{id}_{\mathrm{pa}_{\mathfrak{G}}(\mathcal{A}) \smallsetminus \{\mathcal{L}\}} \right) & \mathcal{L} \in \mathrm{pa}_{\mathfrak{G}}(\mathcal{A}), \\ \Phi_{\mathcal{A}} & \text{otherwise}, \end{cases} \qquad \forall\ \mathcal{A} \in \boldsymbol{\mathcal{A}} \smallsetminus \{\mathcal{L}\},$$

where "id" is the identity function. This function simply removes the measurable function corresponding to the exogenized node and redefines the functions corresponding to its children.

As noted, we use weights $\bar{\boldsymbol{w}}_{\mathcal{A}}$ to represent the functions.Now, let us consider the lvDAG in Figure 10 (a). The variable $\mathcal{U}_1, \ldots, \mathcal{U}_n$ can be arbitrarily visible or latent. The same is true for the variables $\mathcal{V}_1, \ldots, \mathcal{V}_m$. Each $\mathcal{U}_i$ ($1 \leq i \leq n$) only points to $\mathcal{A}$ and $\mathcal{A}$ points to each $\mathcal{V}_i$ ($1 \leq i \leq n$). $\mathcal{V}_i$ does not point to any other variable. In this lvDAG, the structural function for each variable $\mathcal{V}_i$ is computed using $\Phi_{\mathcal{V}_i} := \bar{\boldsymbol{w}}_{\mathcal{V}_i\mathcal{A}}\mathcal{A} + \bar{\boldsymbol{w}}_{\mathcal{V}_i\mathcal{C}_i}\mathcal{C}_i$.

After exogenization, for two variables $\mathcal{V}_i$ and $\mathcal{V}_j$ ($1 \leq i, j \leq n$), the structural functions will be:

$$\Phi'_{\mathcal{V}_i}(\mathrm{pa}_{\mathfrak{G}'}(\mathcal{V}_i)) := \bar{\boldsymbol{w}}_{\mathcal{V}_i\mathcal{A}} \sum_{k=1}^{m} \bar{\boldsymbol{w}}_{\mathcal{A}\mathcal{U}_k} + \bar{\boldsymbol{w}}_{\mathcal{V}_i\mathcal{C}_i}\mathcal{C}_i$$

$$\Phi'_{\mathcal{V}_j}(\mathrm{pa}_{\mathfrak{G}'}(\mathcal{V}_i)) := \bar{\boldsymbol{w}}_{\mathcal{V}_j\mathcal{A}} \sum_{k=1}^{m} \bar{\boldsymbol{w}}_{\mathcal{A}\mathcal{U}_k} + \bar{\boldsymbol{w}}_{\mathcal{V}_i\mathcal{C}_j}\mathcal{C}_j$$

It is seen that for these variables, we have a restricting relation:

$$\mathbb{E}_{\mathcal{C}_i}[\mathcal{V}_i] = c\mathbb{E}_{\mathcal{C}_j}[\mathcal{V}_j],$$

where $c$ is a constant. As seen, this puts a constraint on the functions $\Phi'_{\mathcal{V}_i}$, whereas we do not expect such a constrain on the lvDAG aster exogenization (Figure 10 (b)). ■

**Proof of Corollary 16** Let $\boldsymbol{\mathcal{X}} \equiv \boldsymbol{\mathcal{V}} \mathbin{\dot\cup} \boldsymbol{\mathcal{L}}$ be a vector of visible random variables $\boldsymbol{\mathcal{V}}$ and latent random variables $\boldsymbol{\mathcal{L}}$. Let $\mathfrak{G} = \langle \boldsymbol{\mathcal{A}} \equiv \boldsymbol{\mathcal{V}} \mathbin{\dot\cup} (\boldsymbol{\mathcal{L}} \mathbin{\dot\cup} \{\mathcal{L}\}), \hookrightarrow \rangle$ and $\mathfrak{G}' = \langle \boldsymbol{\mathcal{A}}' \equiv \boldsymbol{\mathcal{V}} \mathbin{\dot\cup} (\boldsymbol{\mathcal{L}} \mathbin{\dot\cup} \{\mathcal{L}'\}), \hookrightarrow' \rangle$, such that $\mathfrak{G}' = \tau_{\mathcal{L}}(\mathfrak{G})$. Now we augment $\mathfrak{G}$ and $\mathfrak{G}'$. Let $\mathfrak{G}_2 = \iota_{\boldsymbol{\mathcal{A}}}(\mathfrak{G}) = \langle \boldsymbol{\mathcal{A}}_2 \equiv \boldsymbol{\mathcal{V}} \mathbin{\dot\cup} ((\boldsymbol{\mathcal{L}} \mathbin{\dot\cup} \{\mathcal{L}\}) \mathbin{\dot\cup} \boldsymbol{\mathcal{S}}), \hookrightarrow_2 \rangle$ and $\mathfrak{G}'_2 = \iota_{\boldsymbol{\mathcal{A}}'}(\mathfrak{G}) = \langle \boldsymbol{\mathcal{A}}'_2 \equiv \boldsymbol{\mathcal{V}} \mathbin{\dot\cup} ((\boldsymbol{\mathcal{L}} \mathbin{\dot\cup} \{\mathcal{L}'\}) \mathbin{\dot\cup} \boldsymbol{\mathcal{S}}), \hookrightarrow'_2 \rangle$ be the augmentation of $\mathfrak{G}$ and $\mathfrak{G}'$ (such that the same $\mathcal{S}$'s in $\boldsymbol{\mathcal{S}}$ point to the same variables $\mathcal{X}$ in $\boldsymbol{\mathcal{X}}$). Now, we deterministically exogenize $\mathcal{L}$ and $\mathcal{L}'$ from these two lvDAGs. We have $\mathfrak{G}_3 = \mathrm{T}_{\mathcal{L}}(\mathfrak{G}_2) = \langle \boldsymbol{\mathcal{A}}_3 \equiv \boldsymbol{\mathcal{V}} \mathbin{\dot\cup} (\boldsymbol{\mathcal{L}} \mathbin{\dot\cup} \boldsymbol{\mathcal{S}}), \hookrightarrow_3 \rangle$ and $\mathfrak{G}'_3 = \mathrm{T}_{\mathcal{L}'}(\mathfrak{G}'_2) = \langle \boldsymbol{\mathcal{A}}'_3 \equiv \boldsymbol{\mathcal{V}} \mathbin{\dot\cup} (\boldsymbol{\mathcal{L}} \mathbin{\dot\cup} \boldsymbol{\mathcal{S}}), \hookrightarrow'_3 \rangle$. But $\mathfrak{G}_3$ and $\mathfrak{G}'_3$ are the same lvDAG. Following this result, we can show that:

$$\begin{aligned}
\mathfrak{q}^{\mathrm{G}(0)}_{\boldsymbol{\mathcal{X}}}(\mathfrak{G}) &= \mathfrak{Q}^{\mathrm{G}(0)}_{\boldsymbol{\mathcal{X}}}(\mathfrak{G}_2) && \text{(Equation 4)}\\
&= \mathfrak{Q}^{\mathrm{G}(0)}_{\boldsymbol{\mathcal{X}}}(\mathfrak{G}_3)\\
&= \mathfrak{Q}^{\mathrm{G}(0)}_{\boldsymbol{\mathcal{X}}}(\mathfrak{G}'_3)\\
&= \mathfrak{Q}^{\mathrm{G}(0)}_{\boldsymbol{\mathcal{X}}}(\mathfrak{G}'_2)\\
&= \mathfrak{q}^{\mathrm{G}(0)}_{\boldsymbol{\mathcal{X}}}(\mathfrak{G}') && \text{(Equation 4)} \blacksquare
\end{aligned}$$

**Lemma 50** *For two lvDAGs $\mathfrak{G}_1 = \langle \boldsymbol{\mathcal{A}}_1 \equiv \boldsymbol{\mathcal{V}}_1 \mathbin{\dot\cup} \boldsymbol{\mathcal{L}}_1, \hookrightarrow_1 \rangle$ and $\mathfrak{G}_2 = \langle \boldsymbol{\mathcal{A}}_2 \equiv \boldsymbol{\mathcal{V}}_2 \mathbin{\dot\cup} \boldsymbol{\mathcal{L}}_2, \hookrightarrow_2 \rangle$ be two lvDAGs, if (a) $\boldsymbol{\mathcal{V}}_1 \equiv \boldsymbol{\mathcal{V}}_2$, (b) $\boldsymbol{\mathcal{L}}_1 \subseteq \boldsymbol{\mathcal{L}}_2$, and (c) for all $\mathcal{I}, \mathcal{J} \in \boldsymbol{\mathcal{A}}_1$, if $\mathcal{I} \hookrightarrow_1 \mathcal{J}$ then $\mathcal{I} \hookrightarrow_2 \mathcal{J}$, then we say that $\mathfrak{G}_1$ is a* sub-DAG *of $\mathfrak{G}_2$ . We denote this by writing $\mathfrak{G}_1 \subseteq \mathfrak{G}_2$. For two lvDAGs $\mathfrak{G}_1$ and $\mathfrak{G}_2$ such that $\mathfrak{G}_2 \subseteq \mathfrak{G}_1$, we have $\mathfrak{Q}^{\mathrm{G}(0)}_{\boldsymbol{\mathcal{V}}}(\mathfrak{G}_2) \subseteq \mathfrak{Q}^{\mathrm{G}(0)}_{\boldsymbol{\mathcal{V}}}(\mathfrak{G}_1)$.*

**Proof** We note that any function $\Phi$ in the structural system $\langle \mathbb{P}, \Phi \rangle$ is a linear transformation (Theorem 44). Removing an edge is equivalent to setting the corresponding linear coefficient to zero. Removing a latent node is equivalent to removing all the incoming an outgoing edges (and then removing that node). Therefore, the DPM of $\mathfrak{G}_2$ is obtained by restricting to the sub-DSM of $\mathfrak{G}_1$ where some linear coefficients are equal to zero. ■

**Proof of Theorem 17** Let $\mathfrak{G} = \langle \boldsymbol{\mathcal{A}} \equiv \boldsymbol{\mathcal{V}} \dot{\cup} \boldsymbol{\mathcal{L}}, \hookrightarrow \rangle$ be an mDAG that is a correlation scenario. Also, let $\mathfrak{G}' = \langle \boldsymbol{\mathcal{A}}' \equiv \boldsymbol{\mathcal{V}} \dot{\cup} \boldsymbol{\mathcal{L}}', \hookrightarrow' \rangle$ such that $\mathfrak{G}' = \mathrm{T}_{\mathrm{root}(\mathfrak{G})}(\iota_{\boldsymbol{\mathcal{A}}}(\mathfrak{G}))$. We consider three collectively exhaustive cases for $\mathfrak{G}$ and prove that for each case $\mathfrak{Q}^{\mathrm{G}(0)}_{\boldsymbol{\mathcal{V}}}(\mathfrak{G}') \subset \mathfrak{P}^{\mathrm{G}(0)}_{\boldsymbol{\mathcal{V}}}$.

(i) Let $\boldsymbol{\mathcal{L}} \equiv \{\mathcal{P}\}$. That is, one latent node points to all visible nodes. This results in $\boldsymbol{\mathcal{L}}' \equiv \boldsymbol{\mathcal{S}} \dot{\cup} \{\mathcal{P}'\}$ such that $\boldsymbol{\mathcal{S}}$ is indexed by $\boldsymbol{\mathcal{V}}$, each $\mathcal{S}_{\mathcal{V}} \in \boldsymbol{\mathcal{S}}$ points only to $\mathcal{V} \in \boldsymbol{\mathcal{V}}$, and $\mathrm{ch}_{\mathfrak{G}'}(\mathcal{P}') \equiv \boldsymbol{\mathcal{V}}$. Let $\varepsilon$ be an arbitrary positive value. If for distinct $\mathcal{X}, \mathcal{Y}, \mathcal{Z} \in \boldsymbol{\mathcal{V}}$ we have $|\mathrm{cov}(\mathcal{X}, \mathcal{Y})| \geq \varepsilon$ and $|\mathrm{cov}(\mathcal{X}, \mathcal{Z})| \geq \varepsilon$, this implies that $|\mathrm{cov}(\mathcal{Y}, \mathcal{Z})| \geq \varepsilon^2 \mathrm{var}^{-1}(\mathcal{X})$ ($\mathrm{var}(\mathcal{X}) > 0$, from Cauchy-Schwarz inequality). This forms a space strictly smaller than the cone of $n$-dimensional positive semi-definite matrices. Therefore, $\mathfrak{Q}^{\mathrm{G}(0)}_{\boldsymbol{\mathcal{V}}}(\mathfrak{G}') \subset \mathfrak{P}^{\mathrm{G}(0)}_{\boldsymbol{\mathcal{V}}}$.

(ii) Let $\boldsymbol{\mathcal{L}} \equiv \boldsymbol{\mathcal{P}}$ such that $\boldsymbol{\mathcal{P}}$ is indexed by $\boldsymbol{\mathcal{V}}$ and each $\mathcal{P}_{\mathcal{V}} \in \boldsymbol{\mathcal{P}}$ points only to all $\mathcal{V}' \in \boldsymbol{\mathcal{V}} \setminus \{\mathcal{V}\}$. This means that $\boldsymbol{\mathcal{L}}' \equiv \boldsymbol{\mathcal{S}} \dot{\cup} \boldsymbol{\mathcal{P}}'$ such that $\boldsymbol{\mathcal{S}}$ and $\boldsymbol{\mathcal{P}}'$ are indexed by $\boldsymbol{\mathcal{V}}$, each $\mathcal{S}_{\mathcal{V}} \in \boldsymbol{\mathcal{S}}$ points only to $\mathcal{V} \in \boldsymbol{\mathcal{V}}$ and each $\mathcal{P}'_{\mathcal{V}} \in \boldsymbol{\mathcal{P}}'$ points only to all $\mathcal{V}' \in \boldsymbol{\mathcal{V}} \setminus \{\mathcal{V}\}$. Moreover, let $\varepsilon$ be an arbitrary positive value. We arbitrarily pick two nodes $\mathcal{X}, \mathcal{Y} \in \boldsymbol{\mathcal{V}}$. For any $\mathcal{Z} \in \boldsymbol{\mathcal{V}} \setminus \{\mathcal{X}, \mathcal{Y}\}$ we assume that (a) $\mathrm{cov}^2(\mathcal{X}, \mathcal{Z}) + \varepsilon \geq \mathrm{var}(\mathcal{X})\, \mathrm{var}(\mathcal{Z})$ and (b) $\mathrm{cov}^2(\mathcal{Y}, \mathcal{Z}) + \varepsilon \geq \mathrm{var}(\mathcal{Y})\, \mathrm{var}(\mathcal{Z})$. We have:

$$\begin{aligned} \mathrm{cov}^2(\mathcal{X}, \mathcal{Z}) + \varepsilon \geq \mathrm{var}(\mathcal{X})\, \mathrm{var}(\mathcal{Z}) \implies \\ \varepsilon \geq \left( \sum_{\mathcal{A} \in \mathrm{pa}_{\mathfrak{G}'}(\mathcal{X})} \bar{w}^2_{\mathcal{A}\mathcal{X}} \bar{\sigma}^2_{\mathcal{A}} \right) \left( \sum_{\mathcal{A} \in \mathrm{pa}_{\mathfrak{G}'}(\mathcal{Z})} \bar{w}^2_{\mathcal{A}\mathcal{Z}} \bar{\sigma}^2_{\mathcal{A}} \right) - \left( \sum_{\mathcal{A} \in \mathrm{pa}_{\mathfrak{G}}(\mathcal{X}) \cap \mathrm{pa}_{\mathfrak{G}}(\mathcal{Z})} \bar{w}^2_{\mathcal{A}\mathcal{X}} \bar{\sigma}^2_{\mathcal{A}} \bar{w}^2_{\mathcal{A}\mathcal{Z}} \right)^2 \implies \\ \varepsilon \geq \mathrm{var}(\mathcal{Z}) \sum_{\mathcal{A} \in \mathrm{pa}_{\mathfrak{G}'}(\mathcal{X}) \setminus \mathrm{pa}_{\mathfrak{G}'}(\mathcal{Z})} \bar{w}^2_{\mathcal{A}\mathcal{X}} \bar{\sigma}^2_{\mathcal{A}}. \end{aligned}$$

This implies that (a) $\bar{w}^2_{\mathcal{P}'_{\mathcal{Z}}\mathcal{X}} \bar{\sigma}^2_{\mathcal{P}'_{\mathcal{Z}}} \leq \frac{\varepsilon}{\mathrm{var}(\mathcal{Z})}$ and, similarly, (b) $\bar{w}^2_{\mathcal{P}'_{\mathcal{Z}}\mathcal{Y}} \bar{\sigma}^2_{\mathcal{P}'_{\mathcal{Z}}} \leq \frac{\varepsilon}{\mathrm{var}(\mathcal{Z})}$. This means that for the nodes $\mathcal{X}$ and $\mathcal{Y}$ themselves, we have:

$$\mathrm{cov}^2(\mathcal{X}, \mathcal{Y}) \leq \left( \sum_{\mathcal{Z} \in \boldsymbol{\mathcal{V}} \setminus \{\mathcal{X}, \mathcal{Y}\}} \frac{\varepsilon}{\mathrm{var}(\mathcal{Z})} \right)^2.$$

For very small values of $\varepsilon$, i.e. when $\mathcal{Z}$'s are highly correlated with both $\mathcal{X}$ and $\mathcal{Y}$, we have that $\mathcal{X}$ and $\mathcal{Y}$ are almost independent. This excludes the case that all variables $\mathcal{V} \in \boldsymbol{\mathcal{V}}$ are highly correlated. Therefore, again, $\mathfrak{Q}^{\mathrm{G}(0)}_{\boldsymbol{\mathcal{V}}}(\mathfrak{G}') \subset \mathfrak{P}^{\mathrm{G}(0)}_{\boldsymbol{\mathcal{V}}}$.[11]

(iii) We consider any other mDAG $\mathfrak{G}$ over $\boldsymbol{\mathcal{V}}$. Then, $\mathfrak{G}'$ will be a sub-DAG of $\mathfrak{G}'$ in (ii). According to Lemma 50, we have $\mathfrak{Q}^{\mathrm{G}(0)}_{\boldsymbol{\mathcal{V}}}(\mathfrak{G}') \subset \mathfrak{P}^{\mathrm{G}(0)}_{\boldsymbol{\mathcal{V}}}$. ■

**Lemma 51** *Let $\mathfrak{C}_n$ be the cone of $n$-dimensional positive definite matrices. Let $\Sigma$ be a non-stochastic variable with $\Omega_\Sigma \subseteq \mathfrak{C}_n$. Let $\hat{\Sigma}_{n \times n} \in \Omega_\Sigma$ be an $n$-dimensional matrix. Then:*

$$\arg\min_{\Sigma} \left\{ \mathrm{KL}\big(\Sigma \,\|\, \hat{\Sigma}\big) \right\} = \arg\min_{\Sigma} \left\{ \mathrm{KL}\big(\hat{\Sigma} \,\|\, \Sigma\big) \right\}.$$

11. For a similar proof for this case, see Evans (2016, Lemma A.2.).

**Proof** Let $\Sigma^* \in \arg\min_\Sigma \left\{ \mathrm{KL}\left(\Sigma \,\|\, \hat{\Sigma}\right) \right\}$. Then,

$$\frac{\partial}{\partial \Sigma} \mathrm{KL}\left(\Sigma^* \,\|\, \hat{\Sigma}\right) = \mathbf{0} \iff -{\Sigma^*}^{-1}\, S\, {\Sigma^*}^{-1} + {\Sigma^*}^{-1} = \mathbf{0} \iff \hat{\Sigma} = \Sigma^*.$$

From Gibbs' inequality $\mathrm{KL}\left(\Sigma \,\|\, \hat{\Sigma}\right) \geq 0$, and since $\mathrm{KL}\left(\hat{\Sigma} \,\|\, \hat{\Sigma}\right) = 0$, $\Sigma^* = \hat{\Sigma}$ is the global minimum of $\mathrm{KL}\left(\Sigma \,\|\, \hat{\Sigma}\right)$. That is,

$$\arg\min_{\Sigma} \left\{ \mathrm{KL}\left(\Sigma \,\|\, \hat{\Sigma}\right) \right\} = \left\{ \hat{\Sigma} \right\}. \tag{35}$$

The same can be proven for $\mathrm{KL}\left(\hat{\Sigma} \,\|\, \Sigma\right)$:

$$\arg\min_{\Sigma} \left\{ \mathrm{KL}\left(\hat{\Sigma} \,\|\, \Sigma\right) \right\} = \left\{ \hat{\Sigma} \right\}. \tag{36}$$

From Equations 35 and 36, $\arg\min_\Sigma \left\{ \mathrm{KL}\left(\Sigma \,\|\, \hat{\Sigma}\right) \right\} = \arg\min_\Sigma \left\{ \mathrm{KL}\left(\hat{\Sigma} \,\|\, \Sigma\right) \right\}$. ∎

**Proof of Theorem 20** We let $\Sigma(\langle \mathbb{P}, \Phi \rangle) := \mathrm{cov}\left(\mathcal{V}; \Pi(\langle \mathbb{P}, \Phi \rangle)_{\mathcal{V}}\right)$. Moreover, let $m = \mathrm{card}(V)$ and $n = \mathrm{card}(\mathcal{V})$. The log-likelihood estimation will be:

$$\begin{aligned} \ell\ell_{\mathfrak{G},V}(\langle \mathbb{P}, \Phi \rangle) &:= \ln \ell_{\mathfrak{G},V}(\langle \mathbb{P}, \Phi \rangle) \\ &= -\frac{m}{2} \left\{ \ln |\Sigma(\langle \mathbb{P}, \Phi \rangle)| + \mathrm{tr}\left( \Sigma(\langle \mathbb{P}, \Phi \rangle)^{-1}\, \hat{\Sigma}(V) \right) + n \ln (2\pi) \right\}, \end{aligned}$$

and the KL divergence is computed using:

$$\mathrm{KL}\left(\mathbb{P}^V_{\mathcal{V}} \,\|\, \Pi(\langle \mathbb{P}, \Phi \rangle)_{\mathcal{V}}\right) = \frac{1}{2} \left\{ \ln |\Sigma(\langle \mathbb{P}, \Phi \rangle)| + \mathrm{tr}\left( \Sigma(\langle \mathbb{P}, \Phi \rangle)^{-1}\, \hat{\Sigma}(V) \right) - \ln \left| \hat{\Sigma}(V) \right| - n \right\}.$$

We see that the log-likelihood and KL divergence have a linear relationship, i.e.

$$\ell\ell_{\mathfrak{G},V}(\langle \mathbb{P}, \Phi \rangle) = -\alpha\, \mathrm{KL}\left( \Sigma(\langle \mathbb{P}, \Phi \rangle) \,\|\, \hat{\Sigma}(V) \right) + \beta,$$

with $\alpha = m$ and $\beta = -{}^{m}\!/_{2} - \ln\left|\hat{\Sigma}(V)\right| - mn - {}^{mn}\!/_{2} \ln(2\pi)$. The monotonicity of the KL divergence w.r.t. the log-likelihood and the monotonicity of the log-likelihood w.r.t. the likelihood function proves that we can minimize the KL divergence to get the MLE. We only need to prove that the KL divergence is symmetric in its minimum point. This is proven in Lemma 51. Therefore,

$$\mathrm{MLE}_{\mathfrak{G},V}(\mathfrak{I}) = \arg\min_{\langle \mathbb{P}, \Phi \rangle \in \mathfrak{I}} \left\{ \mathrm{KL}\left( \Pi(\langle \mathbb{P}, \Phi \rangle)_{\mathcal{V}} \,\|\, \mathbb{P}^V_{\mathcal{V}} \right) \right\}.$$ ∎

**Lemma 52** *Let $\mathfrak{G}$ be a pmDAG and $\Xi(\mathfrak{G}) = \langle \mathfrak{G}, \mathfrak{A}, \leq \rangle$ its synchronization. For each $0 < l < \|\Xi(\mathfrak{G})\|$, if $\mathcal{A} \in \mathfrak{A}_l$, then $\mathrm{pa}_{\Xi(\mathfrak{G})|l}(\mathcal{A}) \subseteq \mathfrak{A}_{l-1}$.*

**Proof** Either $\mathrm{app}_{\leq}(\mathcal{A}) = l$ or not. If $\mathrm{app}_{\leq}(\mathcal{A}) = l$, it means that layer $l$ is the first layer that $\mathcal{A}$ is met. Therefore, it was a root node in the remainder of $\mathfrak{G}'$ in Definition 22. As per 22 (c), all of its parents must be in $\mathfrak{A}_{l-1}$. If $\mathrm{app}_{\leq}(\mathcal{A}) \neq l$, $\mathcal{A}$ has already been visited. If its a visible variable, it must also be in $\mathfrak{A}_{l-1}$ as per 22 (c). Otherwise, it has a non-visited child in the remainder of $\mathfrak{G}'$. Since it has a non-visited child in iteration $l$, it must have had a non-visited child in iteration $l-1$; therefore $\mathcal{A} \in \mathfrak{A}_{l-1}$. We have $\mathcal{A} \in \mathfrak{A}_{l-1}$ and $\mathrm{pa}_{\Xi(\mathfrak{G})|l}(\mathcal{A}) = \{\mathcal{A}\}$. Therefore, the consequent holds in all cases. ■

**Proof of SFP** We denote the regular conditional probability using $\mathbb{P}_{\cdot|\cdot}$.

$$
\begin{aligned}
\mathbb{P}_{\boldsymbol{\mathcal{U}}}(\boldsymbol{X}) &= \int_{\Omega_{\mathrm{pa}_{\Xi(\mathfrak{G})|l}(\boldsymbol{\mathcal{U}})}} \mathbb{P}_{\boldsymbol{\mathcal{U}}|\mathrm{pa}_{\Xi(\mathfrak{G})|l}(\boldsymbol{\mathcal{U}})}(\boldsymbol{X} \mid \boldsymbol{y})\, \mathrm{d}\mathbb{P}_{\mathrm{pa}_{\Xi(\mathfrak{G})|l}(\boldsymbol{\mathcal{U}})}(\boldsymbol{y}) \\
&= \int_{\Omega_{\mathrm{pa}_{\Xi(\mathfrak{G})|l}(\boldsymbol{\mathcal{U}})}} \prod_{\mathcal{U} \in \boldsymbol{\mathcal{U}}} \mathbb{P}_{\mathcal{U}|\mathrm{pa}_{\Xi(\mathfrak{G})|l}(\mathcal{U})}(X_{\mathcal{U}} \mid \boldsymbol{y})\, \mathrm{d}\mathbb{P}_{\mathrm{pa}_{\Xi(\mathfrak{G})|l}(\boldsymbol{\mathcal{U}})}(\boldsymbol{y}) \\
&= \int_{\Omega_{\mathrm{pa}_{\Xi(\mathfrak{G})|l}(\boldsymbol{\mathcal{U}})}} \prod_{\mathcal{U} \in \boldsymbol{\mathcal{U}}} \mathbb{P}_{\mathcal{U}|\mathrm{pa}_{\Xi(\mathfrak{G})|l}(\mathcal{U})}\Big(X_{\mathcal{U}} \,\Big|\, \mathrm{proj}_{\mathrm{pa}_{\Xi(\mathfrak{G})|l}(\mathcal{U})}(\boldsymbol{y})\Big)\, \mathrm{d}\mathbb{P}_{\mathrm{pa}_{\Xi(\mathfrak{G})|l}(\boldsymbol{\mathcal{U}})}(\boldsymbol{y}) \\
&= \int_{\Omega_{\mathrm{pa}_{\Xi(\mathfrak{G})|l}(\boldsymbol{\mathcal{U}})}} \prod_{\mathcal{U} \in \boldsymbol{\mathcal{U}}} \kappa_{\mathcal{U}|l}\Big(X_{\mathcal{U}} \,\Big|\, \mathrm{proj}_{\mathrm{pa}_{\Xi(\mathfrak{G})|l}(\mathcal{U})}(\boldsymbol{y})\Big)\, \mathrm{d}\mathbb{P}_{\mathrm{pa}_{\Xi(\mathfrak{G})|l}(\boldsymbol{\mathcal{U}})}(\boldsymbol{y}) \\
&= \mathbb{E}_{\mathrm{pa}_{\Xi(\mathfrak{G})|l}(\boldsymbol{\mathcal{U}})}\Bigg[\prod_{\mathcal{U} \in \boldsymbol{\mathcal{U}}} \kappa_{\mathcal{U}|l}(X_{\mathcal{U}} \mid \cdot)\Bigg].
\end{aligned}
$$

■

**Lemma 53** *Let $\Xi(\mathfrak{G}) = \langle \mathfrak{G}, \mathfrak{A}, \leq \rangle$ be the synchronization of $\mathfrak{G}$. For $0 < l_1 \leq l_2 \leq l_3 < \|\Xi(\mathfrak{G})\|$, $\mathcal{P} \in \mathfrak{A}_{l_1}$ and $\mathcal{P} \in \mathfrak{A}_{l_3}$ implies $\mathcal{P} \in \mathfrak{A}_{l_2}$.*

**Proof** From Definition 22 (2) (c), since $\boldsymbol{\mathcal{P}}$ accumulates nodes from $\boldsymbol{\mathcal{A}}'$ in each iteration, once $\mathcal{V} \in \boldsymbol{\mathcal{V}}$ appears in $\mathfrak{A}_{l_1}$ it will remain present in any $\mathfrak{A}_l$, $l \geq l_1$. For the same reason and given that nodes are removed from $\boldsymbol{\mathcal{C}}$, $\mathcal{L} \in \boldsymbol{\mathcal{L}}$ remains present $\mathfrak{A}_{l_i}$ until none of its children remains in $\boldsymbol{\mathcal{C}}$. ■

## Appendix D. Numerical Stability of the SN$^2$ Algorithm

We consider an optimization algorithm that is based on gradient descent to be numerically stable if the gradient of the objective function w.r.t. the parameters tends to be a finite vector. In our case, we define the numerical stability over $I$ iterations of the SN$^2$ algorithm as the probability that $\nabla_{\boldsymbol{W}} \mathrm{Err}\big(\boldsymbol{W}; \hat{\Sigma}\big)$ remains finite within $I$ iterations of the algorithm, when the stochastic gradient descent (SGD; Equation 21) optimization is used. We denote the numerical stability over $I$ iterations for pmDAG $\mathfrak{G}$ and with learning rate $\eta$ using $\rho_{\mathfrak{G}}(\eta; I)$.

In our setup, we test the numerical stability on eight pmDAGs, which are the same DAGs present in the §8.2 of this article, namely: (a) back-door, (b) front-door, (c) M, (d) napkin, (e) IV, (f) bow, (g) extended bow, and (h) bad M. These pmDAGs have been depicted in Figure 11. We construct the pmDAG $\mathfrak{G}' = \mathrm{T}_{\mathrm{root}(\mathfrak{G})}(\iota_{\boldsymbol{\mathcal{A}}}(\mathfrak{G}))$ first. The goal is to test the numerical stability against different loss functions, namely the KL divergence, Bhattacharya, and squared Hellinger loss functions. Moreover, we would like to compare the stability when using either of the covariance or accumulation methods.

We estimate numerical stability by running the SN$^2$ algorithm $N$ times and dividing by $N$ the number of times the gradient has remained finite within the $I$ iterations. We denote this estimation using $\hat{\rho}_{\mathfrak{G}'}(\eta;I)$. We estimate the numerical stability over $I = 100$ iterations for the pmDAGs by setting $N = 200$. To see how $\rho_{\mathfrak{G}'}(\eta;I)$ reacts to the learning rate, we estimated it for six different learning rates $\eta \in \left\{10^{-4}, 10^{-3}, 10^{-2}, 10^{-1}, 10^{0}, 10^{1}\right\}$.

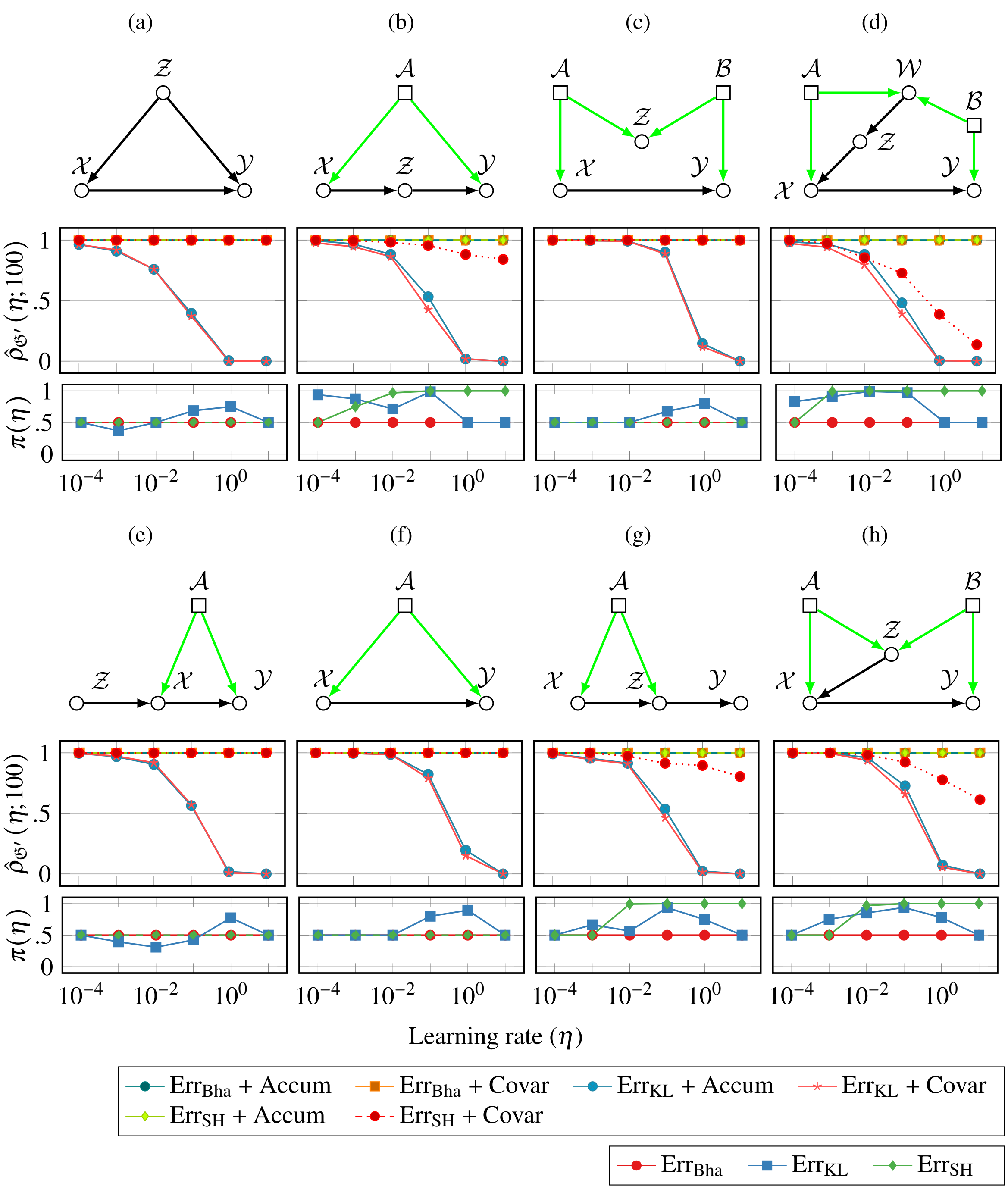

Figure 11: On each row, top: the pmDAGs; middle: the estimated success rate of the $\mathrm{SN}^2$ algorithm within 100 iteration given the learning rate $\eta$ using three loss functions and two methods; bottom: the posterior probability that the accumulation method has a higher success rate than the covariance method. 200 experiments have been done for each pmDAG and learning rate and the results have been averaged out.. (a) back-door; (b) front-door; (c) M; (d) napkin; (e) IV; (f) bow; (g) extended bow; (h) bad M. pmDAGs are equivalent to those experimented on in Xia et al. (2021).

As seen in Figure 11 (Middle), the numerical stability is robust when using the Bhattacharya loss function. Conversely, when we use the KL divergence loss function, the numerical stability of $\mathrm{SN}^2$ becomes very sensitive to the learning rate $\eta$: If we set $\eta = 10$ it is never the case that $\mathrm{SN}^2$ remains numerically stable within 100 iterations. It is, therefore, important that when using the KL divergence loss, the value of $\eta$ is chosen small enough so that the algorithm remains stable. The squared Hellinger distance falls in-between; its stability tends to drop more mildly than the KL divergence when $\eta$ gets bigger. The reason behind the instability of the KL divergence and squared Hellinger loss functions requires further investigation.

Secondly, we measure the probability that the accumulation method is more stable than the covariance method. We denote the numerical stability using the accumulation method by $\rho^{\mathrm{ACC}}_{\mathfrak{G}'}(\eta;I)$ and the one using the covariance method by $\rho^{\mathrm{COV}}_{\mathfrak{G}'}(\eta;I)$. Then, we compute the posterior probability:

$$\pi(\eta) := \mathbb{P}\left(\rho^{\mathrm{ACC}}_{\mathfrak{G}'}(\eta;I) > \rho^{\mathrm{COV}}_{\mathfrak{G}'}(\eta;I) \,\middle|\, \hat{\rho}^{\mathrm{ACC}}_{\mathfrak{G}'}(\eta;I), \hat{\rho}^{\mathrm{COV}}_{\mathfrak{G}'}(\eta;I)\right),$$

where $\rho^{\cdot}_{\mathfrak{G}'}(\eta;I) \sim \mathrm{Beta}(\alpha = 1, \beta = 1)$ are the non-informative priors, and the likelihood variables are $N\hat{\rho}^{\cdot}_{\mathfrak{G}'}(\eta;I) \,|\, \rho^{\cdot}_{\mathfrak{G}'}(\eta;I) \sim \mathrm{Binomial}\left(N = 200, \rho^{\cdot}_{\mathfrak{G}'}(\eta;I)\right)$. Note that $\hat{\rho}^{\mathrm{ACC}}_{\mathfrak{G}'}(\eta;I)$ and $\hat{\rho}^{\mathrm{COV}}_{\mathfrak{G}'}(\eta;I)$ are sufficient statistics and completely represent the samples.

We report $\pi(\eta)$ in Figure 11 (Bottom). We observe that, for both the KL divergence and Bhattacharyya distance, when the numerical stability is near 0 or near 1, there is no difference in terms of which method works better. Therefore, for the Bhattacharyya loss function, $\pi(\eta)$ is always equal to $1/2$. For the KL divergence loss function, the accumulation method tends to be [slightly] more robust ($\pi > 1/2$). Although this difference is negligible, one might prefer the accumulation to the covariance method in order to get better numerical stability. The superiority of the accumulation method, however, is more prominent in the squared Hellinger distance. It remains immune to the change of learning rate when using the accumulation method, but tends to lose numerical stability as $\eta$ increases when the covariance method is used. Therefore, the accumulation method should always be preferred when using this loss function.